\pdfoutput=1

\documentclass[10pt,twocolumn,letterpaper]{article}

\usepackage[pagenumbers]{cvpr} %

\usepackage{amsmath}       %
\usepackage{amssymb}       %
\usepackage{algorithm}     %
\usepackage{tikz}     %
\usepackage{graphicx}
\usetikzlibrary{calc}
\usepackage{fontawesome5}

\newlength{\figW}
\newlength{\vocabw}\newlength{\vocabh}\newlength{\vocabpitch}
\newlength{\trainw}\newlength{\trainh}\newlength{\trainpitch}
\newlength{\capdrop}\newlength{\capH}\newlength{\capHtmp}\newlength{\capHlast}
\newlength{\rowgap}\newlength{\capWmax}\newlength{\stackdepth}
\newlength{\vocabx}\newlength{\trainx}\newlength{\divx}
\newlength{\titley}\newlength{\rowy}
\newsavebox{\capbox}

\newcommand{\figfont}{\scriptsize}      %

\newcommand{\evsep}{\kern0.12em\textbullet\kern0.12em}
\newcommand{\evarr}{\kern0.15em\ensuremath{\rightarrow}\kern0.15em}
\newcommand{\evtok}[1]{\mbox{#1}}

\newcommand{\clipline}[1]{\scalebox{\capscale}{\fontsize{10}{12}\selectfont #1}}

\newcommand{\measurecapW}[1]{%
  \settowidth{\capHtmp}{\fontsize{10}{12}\selectfont #1}%
  \ifdim\capHtmp>\capWmax\setlength{\capWmax}{\capHtmp}\fi}

\newcommand{\clipseqA}{\evtok{6 obj}\evarr\evtok{move}\evsep\evtok{recolor}\evsep\evtok{-rem}\evsep\evtok{swap}\evsep\evtok{-rem}\evsep\evtok{move}\evarr\evtok{4 obj}}
\newcommand{\clipseqB}{\evtok{9 obj}\evarr\evtok{swap}\evsep\evtok{move}\evsep\evtok{recolor}\evsep\evtok{+add}\evsep\evtok{+add}\evsep\evtok{move}\evsep\evtok{+add}\evarr\evtok{12 obj}}
\newcommand{\clipseqC}{\evtok{6 obj}\evarr\evtok{swap}\evsep\evtok{recolor}\evsep\evtok{-rem}\evsep\evtok{+add}\evsep\evtok{move}\evsep\evtok{move}\evsep\evtok{recolor}\evarr\evtok{6 obj}}
\newcommand{\clipseqD}{\evtok{7 obj}\evarr\evtok{swap}\evsep\evtok{recolor}\evsep\evtok{-rem}\evsep\evtok{+add}\evsep\evtok{-rem}\evsep\evtok{move}\evsep\evtok{move}\evarr\evtok{6 obj}}

\newcommand{\measureclipseq}[1]{%
  \sbox{\capbox}{\parbox{\trainw}{\centering\capfont #1}}%
  \setlength{\capHtmp}{\ht\capbox}\addtolength{\capHtmp}{\dp\capbox}%
  \ifdim\capHtmp>\capH\setlength{\capH}{\capHtmp}\fi}
  
\usepackage{algpseudocode} %

\usepackage{array}       

\usepackage{multirow}     

\usepackage{threeparttable} 
\usepackage[table]{xcolor}
\usepackage{subcaption}
\usepackage{pifont}
\usepackage{booktabs}

\usepackage{pgfplots}
\usepgfplotslibrary{groupplots}
\pgfplotsset{compat=1.18}

\definecolor{grouprow}{RGB}{252, 217, 217}
\definecolor{ourrow}{RGB}{247, 238, 244}   %
\definecolor{rank1}{RGB}{214, 176, 197}    %
\definecolor{rank2}{RGB}{233, 213, 224}    %
\newcommand{\rk}[1]{\cellcolor{rank1}#1}        %
\newcommand{\rs}[1]{\cellcolor{rank2}#1}        %
\definecolor{cvprblue}{rgb}{0.21,0.49,0.74}
\usepackage[pagebackref,breaklinks,colorlinks,allcolors=cvprblue]{hyperref}

\makeatletter
\renewcommand\paragraph{\@startsection{paragraph}{4}{\z@}%
  {1ex \@plus .2ex \@minus .2ex}%
  {-1em}%
  {\normalfont\normalsize\bfseries}}
\makeatother

\usepackage[capitalize,nameinlink]{cleveref}

\AddToHook{cmd/appendix/before}{%
  \crefalias{section}{appendix}%
}

\renewcommand{\mid}{\,|\,}

\usepackage{booktabs,tabularx,array,makecell,xcolor,tikz}
\definecolor{headblue}{HTML}{071D4A}
\definecolor{lightblue}{HTML}{F2F6FC}
\definecolor{lightgreen}{HTML}{F1F8EE}
\definecolor{rulegray}{HTML}{CCD2DC}

\usepackage{fontawesome5}
\newcommand{\ok}{%
  \tikz[baseline=-.55ex]\node[
    fill=green!70!black,text=white,
    minimum height=1.3em,
    minimum width=1.3em,
    rounded corners=1.2pt,
    inner sep=1pt,font=\bfseries,scale=.9] {\faCheck};%
}
\newcommand{\bad}{%
  \tikz[baseline=-.5ex,scale=.8]{
  \draw[line width=2pt,red!90!black] (-.5em,-.5em)--(.5em,.5em);
  \draw[line width=2pt,red!90!black] (.5em,-.5em)--(-.5em,.5em);
  }
}

\newcolumntype{L}{>{\centering\arraybackslash\bfseries}p{2.6cm}}
\newcolumntype{C}[1]{>{\centering\arraybackslash}m{#1}}

\makeatletter
\newcommand{\rowstretch}[1]{%
  \global\setbox\@arstrutbox\hbox{%
    \vrule
      \@height #1\ht\strutbox
      \@depth  #1\dp\strutbox
      \@width  0pt}}
\makeatother

\def\confName{CVPR}
\def\confYear{2027}

\title{
    \begin{minipage}{0.12\textwidth} %
        \raggedleft
        \includegraphics[height=1.5cm]{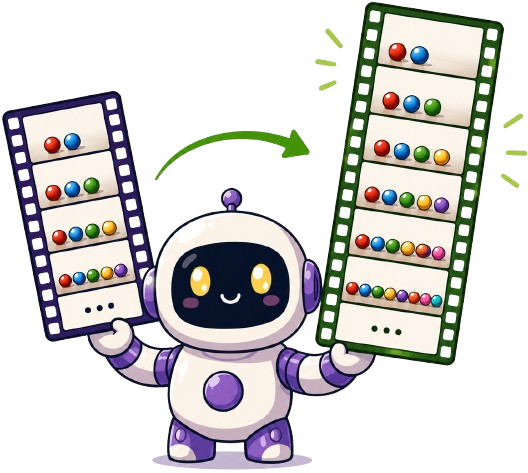} %
    \end{minipage}%
    \hspace{0.005cm} %
    \begin{minipage}{0.8\textwidth} %
        \centering
        \textbf{Temporal Self-Distillation: Learning Visual State Tracking \\ in Videos Without Supervision}
    \end{minipage}
}

\author{
\parbox{\textwidth}{
\vspace*{-6pt}
\centering
Shravan Venkatraman$^{1,2*}$ \quad
Wenshuai Zhao$^{2}$ \quad
Mohammad Hassan Vali$^{2}$ \quad
Arno Solin$^{2}$ \\[2pt]
$^{1}$Mohamed bin Zayed University of Artificial Intelligence, UAE \\
$^{2}$ELLIS Institute Finland \& Department of Computer Science, Aalto University, Finland \\
{\small\texttt{shravan.venkatraman@mbzuai.ac.ae,\{wenshuai.zhao,mohammad.vali,arno.solin\}@aalto.fi}} \\[4pt]
{\small
\faGlobe~\href{https://aaltoml.github.io/s3t/}{Project Page}\quad
\faGithub~\href{https://github.com/AaltoML/s3t}{Code}\quad
\raisebox{-0.2em}{\includegraphics[height=1.1em]{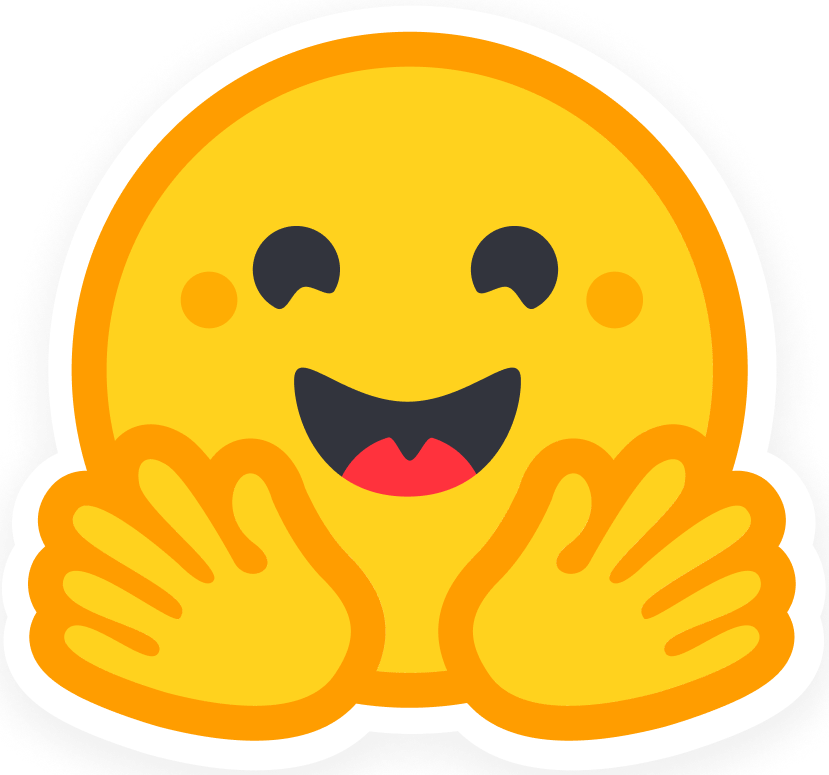}}~\href{https://huggingface.co/shravvvv/s3t}{Models}
}
}
\vspace*{-14pt}
}

\begin{document}

\renewcommand{\ref}[1]{USE CREF INSTEAD OF REF.}

\crefname{figure}{Fig.}{Figs.}
\crefname{algorithm}{Alg.}{Algs.}
\Crefname{section}{Section}{Sections}
\crefname{appendix}{App.}{Apps.}

\twocolumn[{%
    \renewcommand\twocolumn[1][]{#1}%
    \maketitle
    \vspace*{-6pt}%
    \centering
        \captionsetup{type=figure}
\begingroup
\graphicspath{{figs/teaser_assets/}}%
\def\teasertextscale{1}%
\definecolor{priorbg}{HTML}{F2F2F2}%
\definecolor{edgegray}{HTML}{595959}%
\definecolor{oursbg}{HTML}{D9F2D0}%
\definecolor{oursedge}{HTML}{47D45A}%
\definecolor{tabgray}{HTML}{E8E8E8}%
\definecolor{boxgray}{HTML}{D9D9D9}%
\definecolor{studentbg}{HTML}{DCEAF7}%
\definecolor{flawred}{HTML}{C00000}%
\definecolor{goodgreen}{HTML}{3B7D23}%
\definecolor{labelgray}{HTML}{404040}%
\definecolor{studentink}{HTML}{104862}%
\definecolor{distillblue}{HTML}{4E95D9}%
\definecolor{teachergreen}{HTML}{00B050}%
\definecolor{statepurple}{HTML}{7030A0}%
\definecolor{studentviolet}{HTML}{4602FD}%
\definecolor{guidegray}{HTML}{D1D1D1}%
\definecolor{framegray}{HTML}{808080}%
\definecolor{statebg}{HTML}{F8F6FD}%
\definecolor{stateedge}{HTML}{78206E}%
\definecolor{cardbg}{HTML}{F7F5E0}%
\definecolor{cardedge}{HTML}{FFC000}%
\definecolor{cropbox}{HTML}{F2AA84}%
\definecolor{cropline}{HTML}{C04F15}%
\definecolor{removered}{HTML}{FF0000}%
\definecolor{movegray}{HTML}{747474}%
\definecolor{dotyellow}{HTML}{FFFF00}%
\definecolor{dotblue}{HTML}{96DCF8}%
\definecolor{flowA}{HTML}{A6CAEC}\definecolor{flowB}{HTML}{74C4E9}\definecolor{flowC}{HTML}{4E95D9}%
\definecolor{flowD}{HTML}{84E291}\definecolor{flowE}{HTML}{B4E5A2}%
\pgfdeclareradialshading{teaserToStudent}{\pgfpointorigin}{%
  color(0bp)=(flowA); color(7.75bp)=(flowB); color(12.25bp)=(flowC);
  color(25bp)=(studentviolet); color(50bp)=(studentviolet)}%
\pgfdeclareradialshading{teaserToTeacher}{\pgfpointorigin}{%
  color(0bp)=(flowD); color(7.75bp)=(flowE); color(14.25bp)=(flowD);
  color(24.75bp)=(teachergreen); color(50bp)=(teachergreen)}%
\def\fs#1{\dimen0=\teasertextscale\dimexpr#1bp\relax\fontsize{\the\dimen0}{1.2\dimen0}\selectfont}%
\def\img#1#2#3#4#5{\node[inner sep=0pt, anchor=north west] at (#1,#2)
  {\includegraphics[width=#3bp, height=#4bp]{#5}};}%
\def\imgc#1#2#3#4#5#6{\node[inner sep=0pt, #1] at (#2,#3)
  {\includegraphics[width=#4bp, height=#5bp]{#6}};}%
\def\cropimg#1#2#3#4#5#6#7#8#9{%
  \begin{scope}
    \clip (#1,#2) rectangle ++(#3,#4);
    \pgfmathsetmacro\cropW{#3/(1-#5-#7)}%
    \pgfmathsetmacro\cropH{#4/(1-#6-#8)}%
    \node[inner sep=0pt, anchor=north west] at ({#1-#5*\cropW},{#2-#6*\cropH})
      {\includegraphics[width=\cropW bp, height=\cropH bp]{#9}};
  \end{scope}}%
\def\icon#1#2#3#4#5{\begin{scope}[shift={(#1,#2)}]\pic[x=#3bp/96, y=-#4bp/96] {#5};\end{scope}}%
\def\glyph#1#2#3#4{\begin{scope}[shift={(#1,#2)}]\pic[x=#3bp/1000, y=#3bp/1000] {#4};\end{scope}}%
\tikzset{
  txt/.style        = {inner sep=0pt, outer sep=0pt, anchor=base, align=center},
  frame sep/.style  = {draw=black, line width=1.5bp},        %
  link/.style       = {draw=edgegray, line width=0.75bp, dash pattern=on 2.25bp off 0.75bp},
  guide/.style      = {draw=guidegray, line width=0.75bp, dash pattern=on 2.25bp off 0.75bp},
  zoom guide/.style = {draw=cropline, line width=0.75bp, dash pattern=on 3bp off 2.25bp},
  row/.style        = {fill=priorbg, draw=edgegray, line width=0.5bp},   %
  question/.style   = {fill=boxgray, draw=edgegray, line width=0.75bp},  %
  verdict/.style    = {fill=boxgray, fill opacity=0.55, draw=edgegray, line width=0.5bp},
  answer/.style     = {fill=white, draw=black, line width=0.5bp},        %
  card/.style       = {fill=cardbg, draw=cardedge, line width=0.75bp},   %
  pics/document/.style={code={\fill[black]
      (23,82) -- (23,14) -- (48,14) -- (48,35) -- (73,35) -- (73,82) -- (23,82) -- cycle (54,16.5) -- (66.5,29) -- (54,29) -- (54,16.5) -- cycle (54,8) -- (17,8) -- (17,88) -- (79,88) -- (79,30) -- (54,8) -- cycle
      (29,45) rectangle ++(38,4)
      (29,37) rectangle ++(13,4)
      (29,53) rectangle ++(38,4)
      (29,61) rectangle ++(38,4)
      (29,69) rectangle ++(38,4);}},
  pics/gavel/.style={code={\fill[black]
      (44,80) -- (42,80) -- (42,76) -- (14,76) -- (14,80) -- (12,80) .. controls (9.8,80) and (8,81.8) .. (8,84) .. controls (8,86.2) and (9.8,88) .. (12,88) -- (44,88) .. controls (46.2,88) and (48,86.2) .. (48,84) .. controls (48,81.8) and (46.2,80) .. (44,80) -- cycle
      (43.9,60.1) .. controls (45.1,58.9) and (45.1,57) .. (43.9,55.9) -- (28.1,40.1) .. controls (26.9,38.9) and (25,38.9) .. (23.9,40.1) -- (20,44) -- (40,64) -- (43.9,60.1) -- cycle
      (67.9,31.9) .. controls (69.1,33.1) and (71,33.1) .. (72.1,31.9) -- (76,28) -- (56,8) -- (52.1,11.9) .. controls (50.9,13.1) and (50.9,15) .. (52.1,16.1) -- (67.9,31.9) -- cycle
      (87,67) -- (60,40) -- (66,34) -- (50,18) -- (30,38) -- (46,54) -- (54,46) -- (81,73) .. controls (82.7,74.7) and (85.3,74.7) .. (87,73) .. controls (88.7,71.3) and (88.7,68.7) .. (87,67) -- cycle;}},
  pics/magnifier/.style={code={\fill[black]
      (85.5,75.6) -- (73,63.1) .. controls (71.3,61.4) and (69,60.8) .. (66.8,61.2) -- (62.4,56.8) .. controls (66.3,51.8) and (68.6,45.4) .. (68.6,38.6) .. controls (68.6,22.1) and (55.1,8.6) .. (38.6,8.6) .. controls (22.1,8.6) and (8.6,22.1) .. (8.6,38.6) .. controls (8.6,55.1) and (22.1,68.6) .. (38.6,68.6) .. controls (45.4,68.6) and (51.7,66.3) .. (56.8,62.4) -- (61.2,66.8) .. controls (60.8,69) and (61.4,71.3) .. (63.1,73) -- (75.6,85.5) .. controls (77,86.9) and (78.8,87.6) .. (80.6,87.6) .. controls (82.4,87.6) and (84.2,86.9) .. (85.6,85.5) .. controls (88.2,82.7) and (88.2,78.3) .. (85.5,75.6) -- cycle (38.5,62.5) .. controls (25.3,62.5) and (14.5,51.7) .. (14.5,38.5) .. controls (14.5,25.3) and (25.3,14.5) .. (38.5,14.5) .. controls (51.7,14.5) and (62.5,25.3) .. (62.5,38.5) .. controls (62.5,51.7) and (51.7,62.5) .. (38.5,62.5) -- cycle;}},
  pics/hourglass/.style={code={\fill[black]
      (53.6,53.3) .. controls (57,56) and (60.2,60.7) .. (62.6,66) -- (33.4,66) .. controls (35.9,60.7) and (39,55.9) .. (42.4,53.3) .. controls (44,52) and (45,50.1) .. (45,48) .. controls (45,45.9) and (44,44) .. (42.4,42.7) .. controls (35.6,37.4) and (29.7,23.8) .. (28.7,14) -- (67.4,14) .. controls (66.3,23.8) and (60.5,37.4) .. (53.7,42.7) .. controls (52,44) and (51,45.9) .. (51,48) .. controls (51,50.1) and (52,52) .. (53.6,53.3) -- cycle (73.3,14) -- (76,14) -- (76,8) -- (20,8) -- (20,14) -- (22.6,14) .. controls (23.7,25.1) and (30.1,41.3) .. (39.4,48) .. controls (30.1,54.7) and (23.6,70.9) .. (22.6,82) -- (20,82) -- (20,88) -- (76,88) -- (76,82) -- (73.3,82) .. controls (72.3,70.9) and (65.8,54.7) .. (56.5,48) .. controls (65.8,41.3) and (72.3,25.1) .. (73.3,14) -- cycle;}},
  pics/plus/.style={code={\fill[teachergreen] (249,104) -- (249,313) -- (41,313) -- (41,363) -- (249,363) -- (249,572) -- (301,572) -- (301,363) -- (509,363) -- (509,313) -- (301,313) -- (301,104) -- cycle;}},
  pics/minus/.style={code={\fill[removered] (77,248) -- (77,331) -- (386,331) -- (386,248) -- cycle;}},
  pics/movearrow/.style={code={\fill[framegray]
      (759.28,427.73) .. controls (802.57,380.53) and (840.01,344.65) .. (871.58,320.07) .. controls (903.16,295.49) and (932.94,277.18) .. (960.94,265.14) -- (960.94,243.16) .. controls (928.71,227.54) and (897.46,207.6) .. (867.19,183.35) .. controls (836.91,159.1) and (800.78,123.54) .. (758.79,76.66) -- (721.19,76.66) .. controls (751.79,142.09) and (783.85,192.38) .. (817.38,227.54) -- (38.57,227.54) -- (38.57,276.86) -- (817.38,276.86) .. controls (792.64,308.43) and (775.39,331.87) .. (765.62,347.17) .. controls (755.86,362.47) and (741.37,389.32) .. (722.17,427.73) -- cycle;}},
}%
\resizebox{\linewidth}{!}{%
\begin{tikzpicture}[x=1bp, y=-1bp]
\useasboundingbox (-107.05,64.85) rectangle (1008.95,453.65);

\path[fill=priorbg, fill opacity=0.51, draw=edgegray, line width=1bp, rounded corners=12.25bp] (-105.96,65.41) rectangle ++(473.38,387.65);  %
\node[txt, font=\fs{14}] at (127.51,83.18) {Prior Methods};

\path[row, rounded corners=7.83bp] (-96.93,91.88) rectangle ++(455.19,84.06);
\fill[tabgray] [rounded corners=3.1bp] (-97.25,91.57) rectangle ++(33.22,85.11) [sharp corners] (-86.26,91.35) rectangle ++(33.22,85.11);
\node[txt, rotate=90, font=\fs{14}] at (-63.05,132.44) {Labels /\\Judge};

\path[row, rounded corners=7.83bp] (-96.89,180.88) rectangle ++(455.19,84.06);
\fill[tabgray] [rounded corners=3.1bp] (-97.55,180.3) rectangle ++(33.22,85.11) [sharp corners] (-86.33,180.2) rectangle ++(33.22,85.11);
\node[txt, rotate=90, font=\fs{14}] at (-62.75,223.97) {Self-\\Evolution};

\path[row, rounded corners=7.83bp] (-96.26,271.07) rectangle ++(455.19,84.06);
\fill[tabgray] [rounded corners=3.1bp] (-96.59,270.42) rectangle ++(33.22,85.11) [sharp corners] (-85.6,270.56) rectangle ++(33.22,85.11);
\node[txt, rotate=90, font=\fs{14}] at (-62.35,312.32) {Spatial-\\Privilege};

\path[row, rounded corners=7.83bp] (-96.93,361.56) rectangle ++(455.19,84.06);
\fill[tabgray] [rounded corners=3.1bp] (-97.06,361.23) rectangle ++(33.22,85.11) [sharp corners] (-86.07,361.01) rectangle ++(33.22,85.11);
\node[txt, rotate=90, font=\fs{14}] at (-61.95,403.4) {Temporal\\SSL};

\cropimg{-36.59}{102.03}{160.32}{39.68}{0}{0}{0.6002}{0}{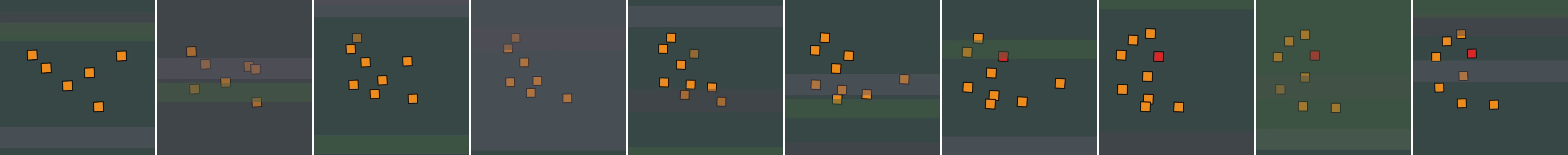}
\cropimg{-36.59}{96.87}{161.54}{8.16}{0}{0.2388}{0.54574}{0.69781}{filmstrip}
\cropimg{-36.73}{141.44}{161.54}{8.16}{0}{0.2388}{0.54574}{0.69781}{filmstrip}
\foreach \x/\ya/\yb in {2.9/107.25/141.71, -35.82/107.25/141.71, 43.58/107.25/141.71, 83.87/107.25/141.71, 123.74/107.25/141.71}
  \draw[frame sep] (\x,\ya) -- (\x,\yb);
\path[question, rounded corners=2.78bp] (-38.2,154.12) rectangle ++(164.58,16.66);  %
\node[txt, font=\fs{10}] at (44.67,165.85) {How many objects are there?};
\img{155.94}{109.79}{37.82}{45.01}{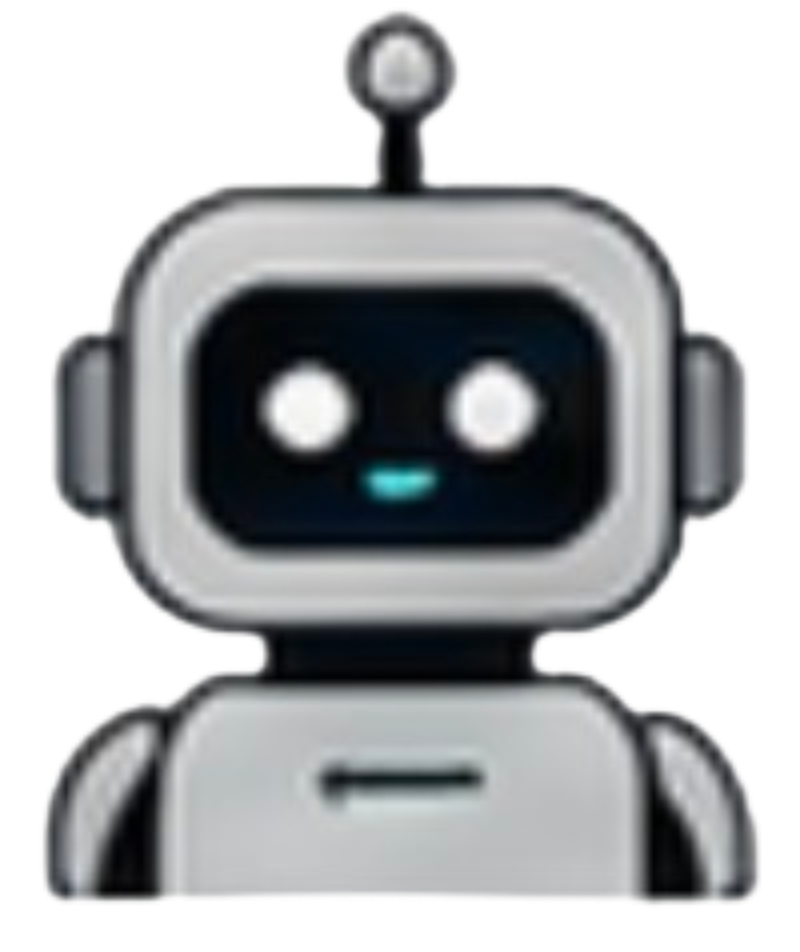}  %
\imgc{rotate=-84.57, yscale=-1}{142.67}{111.04}{19.4}{18.32}{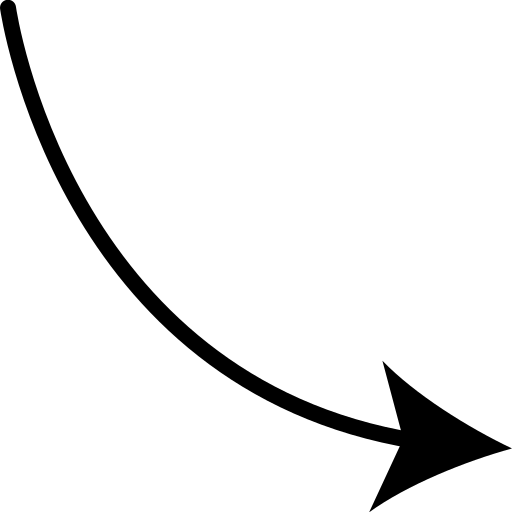}  %
\imgc{rotate=-292.66}{143.34}{158.27}{19.3}{18.72}{arrow-curved}  %
\icon{224.92}{96.45}{34.8}{34.8}{document}  %
\icon{226.99}{135.17}{36.96}{36.96}{gavel}  %
\imgc{rotate=-180, xscale=-1}{208.44}{111.03}{20.27}{17.21}{arrow-curved}  %
\imgc{rotate=-180, xscale=-1, yscale=-1}{209.12}{157.34}{20.27}{15.4}{arrow-curved}  %
\path[verdict, rounded corners=4.88bp] (268.02,99.41) rectangle ++(81.65,29.27);
\node[txt, font=\fs{10}, text=flawred] at (308.92,123.05) {Ground-Truth\\Labels};
\path[verdict, rounded corners=4.94bp] (268.45,137.5) rectangle ++(81.65,29.62);
\node[txt, font=\fs{10}, text=flawred] at (308.9,161.55) {External\\Judge};

\cropimg{-35.72}{201.65}{160.32}{39.68}{0}{0}{0.6002}{0}{prior-frames}
\cropimg{-35.72}{196.49}{161.54}{8.16}{0}{0.2388}{0.54574}{0.69781}{filmstrip}
\cropimg{-35.86}{241.06}{161.54}{8.16}{0}{0.2388}{0.54574}{0.69781}{filmstrip}
\foreach \x/\ya/\yb in {3.77/206.87/241.33, -34.96/206.87/241.33, 44.44/206.87/241.33, 84.74/206.87/241.33, 124.61/206.87/241.33}
  \draw[frame sep] (\x,\ya) -- (\x,\yb);
\img{129.69}{205.64}{19.74}{19.74}{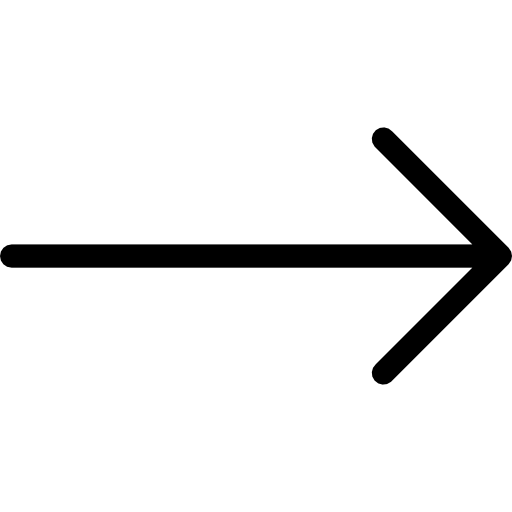}  %
\img{151.76}{187.43}{33.31}{52.54}{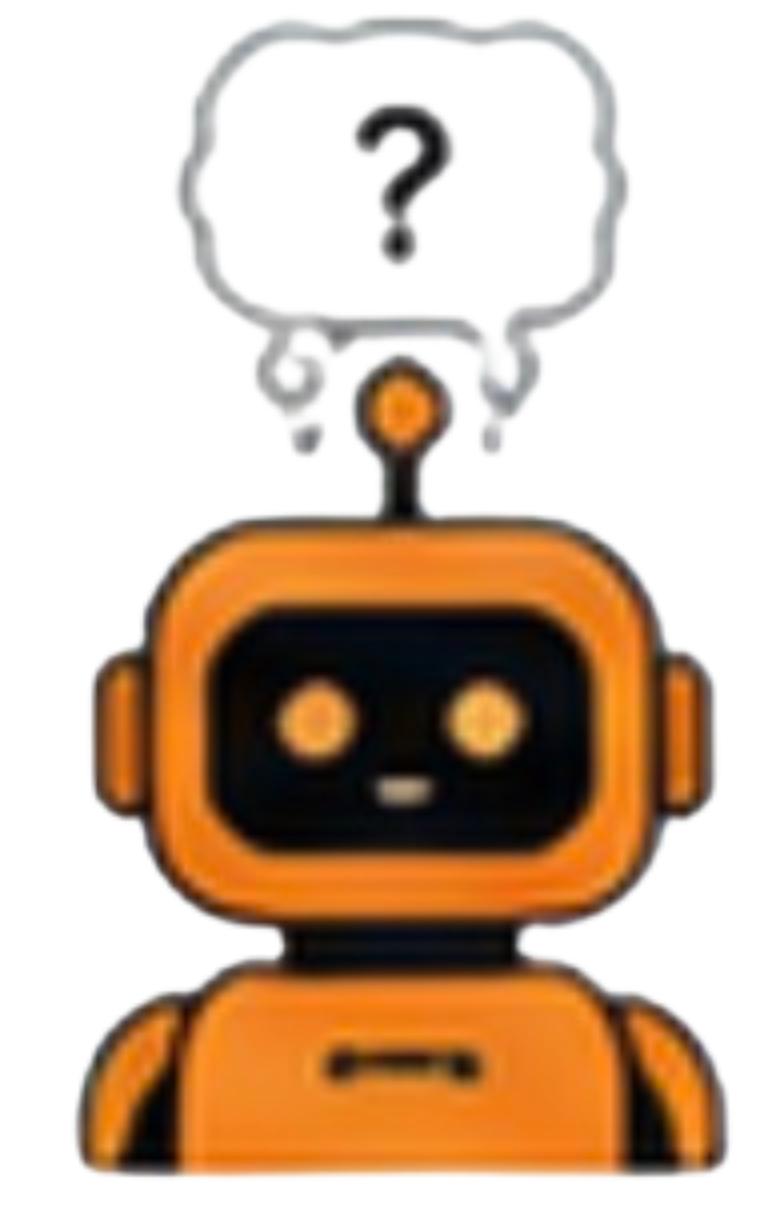}  %
\img{195.15}{201.43}{28.81}{28.81}{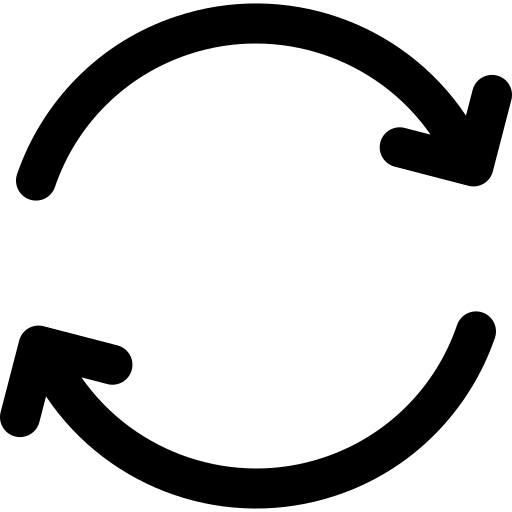}  %
\img{231.61}{187.11}{33.7}{51.41}{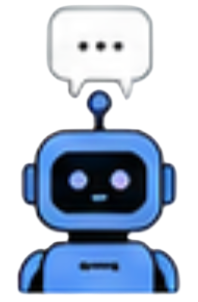}  %
\path[answer, rounded corners=2.78bp] (172.47,243.78) rectangle ++(18.41,16.66);
\node[txt, font=\fs{10}] at (181.31,255.55) {5};
\path[answer, rounded corners=2.78bp] (197.59,243.56) rectangle ++(18.41,16.66);
\node[txt, font=\fs{10}] at (206.09,255.35) {7};
\path[answer, rounded corners=2.78bp] (222.7,243.33) rectangle ++(18.41,16.66);
\node[txt, font=\fs{10}] at (231.62,255.05) {4};
\path[answer, rounded corners=2.78bp] (246.86,243.1) rectangle ++(18.41,16.66);
\node[txt, font=\fs{10}] at (255.7,254.85) {5};
\node[txt, font=\fs{8}, text=labelgray] at (149.62,259.05) {Predicted\\Answer};
\path[verdict, rounded corners=6.68bp] (271.65,201.49) rectangle ++(81.65,40.05);
\node[txt, font=\fs{9}, text=flawred] at (312.38,235.25) {Optimizes\\Consistency over\\Correctness};

\cropimg{-35.43}{282.19}{160.32}{39.68}{0}{0}{0.6002}{0}{prior-frames}
\cropimg{-35.43}{277.03}{161.54}{8.16}{0}{0.2388}{0.54574}{0.69781}{filmstrip}
\cropimg{-35.58}{321.6}{161.54}{8.16}{0}{0.2388}{0.54574}{0.69781}{filmstrip}
\foreach \x/\ya/\yb in {4.05/287.41/321.88, -34.67/287.41/321.88, 44.73/287.41/321.88, 85.02/287.41/321.88, 124.89/287.41/321.88}
  \draw[frame sep] (\x,\ya) -- (\x,\yb);
\path[draw=cropbox, line width=1bp, dash pattern=on 4bp off 3bp] (44.58,285.42) rectangle ++(40.44,37.48);  %
\img{162.76}{279.43}{48.51}{31.14}{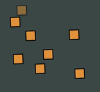}  %
\path[zoom guide] (68.75,291.84) -- (163.43,279.99);  %
\path[zoom guide] (68.75,310.07) -- (162.94,310.07);  %
\icon{200.88}{314.25}{17.1}{17.1}{magnifier}  %
\path[question, rounded corners=2.78bp] (-37.05,334.28) rectangle ++(164.58,16.66);  %
\node[txt, font=\fs{10}] at (46.52,346.05) {How many objects are there?};
\node[txt, font=\fs{8}, text=labelgray] at (171.72,330.85) {Teacher View\\(Crop/Zoom)};
\imgc{rotate=-84.57, yscale=-1}{228.76}{297.03}{19.4}{20.05}{arrow-curved}  %
\img{228.1}{308.77}{31}{36.89}{robot-gray}  %
\path[verdict, rounded corners=4.88bp] (272.62,278.86) rectangle ++(81.65,29.27);
\node[txt, font=\fs{10}, text=flawred] at (313.5,303.35) {Only Spatial\\Knowledge};
\path[verdict, rounded corners=4.94bp] (273.05,316.95) rectangle ++(81.65,29.62);
\node[txt, font=\fs{10}, text=flawred] at (313.51,341.75) {No Temporal\\Information};

\cropimg{-36.06}{380.44}{87.34}{23.41}{0}{0}{0.70221}{0}{prior-frames}
\cropimg{-36.06}{377.4}{87.72}{5.87}{0}{0.2388}{0.66273}{0.6839}{filmstrip}
\cropimg{-36.06}{403.25}{87.72}{5.87}{0}{0.2388}{0.66273}{0.6839}{filmstrip}
\foreach \x/\ya/\yb in {-7.18/382.35/403.85, -35.51/383.52/403.85, 22.57/382.35/403.85, 50.89/381.98/405.74}
  \draw[frame sep] (\x,\ya) -- (\x,\yb);
\cropimg{58.31}{380.24}{87.34}{23.41}{0}{0}{0.70221}{0}{prior-frames}
\cropimg{58.31}{377.19}{87.72}{5.87}{0}{0.2388}{0.66273}{0.6839}{filmstrip}
\cropimg{58.31}{403.05}{87.72}{5.87}{0}{0.2388}{0.66273}{0.6839}{filmstrip}
\foreach \x/\ya/\yb in {87.19/382.14/403.65, 58.87/383.32/403.65, 116.94/382.14/403.65, 145.27/381.78/405.53}
  \draw[frame sep] (\x,\ya) -- (\x,\yb);
\cropimg{152.68}{380.03}{87.34}{23.41}{0}{0}{0.70221}{0}{prior-frames}
\cropimg{152.68}{376.99}{87.72}{5.87}{0}{0.2388}{0.66273}{0.6839}{filmstrip}
\cropimg{152.68}{402.84}{87.72}{5.87}{0}{0.2388}{0.66273}{0.6839}{filmstrip}
\foreach \x/\ya/\yb in {181.56/381.94/403.44, 153.24/383.11/403.44, 211.31/381.94/403.44, 239.64/381.57/405.33}
  \draw[frame sep] (\x,\ya) -- (\x,\yb);
\node[txt, font=\fs{9}, text=labelgray] at (9.21,374.05) {Shuffle};
\node[txt, font=\fs{9}, text=labelgray] at (101.75,373.75) {Reverse};
\node[txt, font=\fs{9}, text=labelgray] at (193.84,373.55) {Fast/Slow};
\img{-3.32}{413.87}{25.38}{25.38}{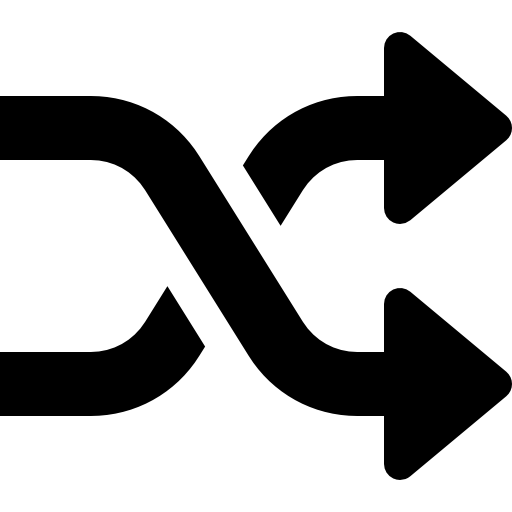}  %
\img{83.43}{417.26}{36.95}{18.26}{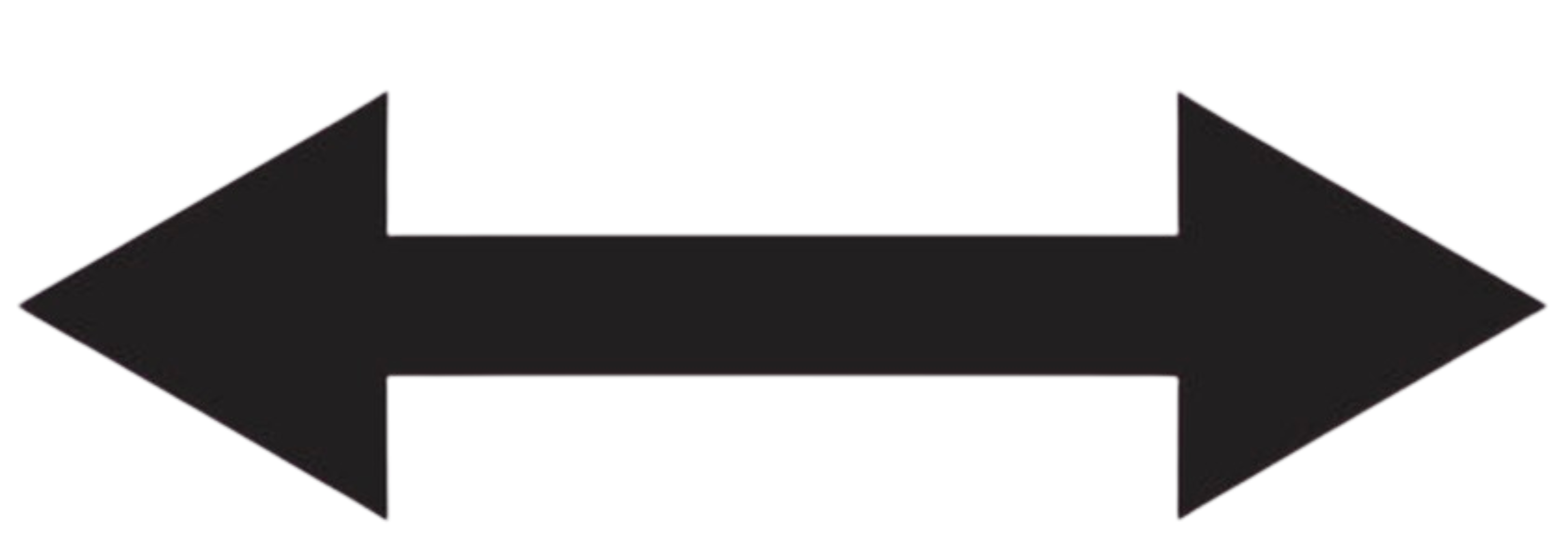}  %
\icon{185.02}{414.39}{24.25}{24.25}{hourglass}  %
\path[verdict, rounded corners=4.94bp] (261.9,370.86) rectangle ++(92.38,29.62);
\node[txt, font=\fs{9.5}, text=goodgreen] at (307.62,394.55) {Learns Temporal\\Transformations};
\path[verdict, rounded corners=4.94bp] (261.41,409.68) rectangle ++(92.46,29.62);
\node[txt, font=\fs{10}, text=flawred] at (308.7,434.25) {Can’t Maintain\\Running State};

\path[fill=oursbg, fill opacity=0.5, draw=oursedge, line width=1bp, rounded corners=12.25bp] (371.81,65.41) rectangle ++(635.73,387.65);  %
\path[fill=studentbg, rounded corners=8.48bp] (463.32,239.43) rectangle ++(145.59,107.64);  %
\path[fill=oursbg, rounded corners=8.48bp] (760.1,239.43) rectangle ++(145.59,107.64);  %
\node[txt, font=\fs{14}] at (691.02,82.96) {S\raisebox{4.2bp}{\fs{9.33}3}T: Self-Supervised Self-Distillation over Time (Ours)};

\path[fill=dotyellow, draw=framegray, line width=0.5bp] (509.19,108.53) ellipse [x radius=5.1bp, y radius=5.19bp];  %
\path[fill=dotblue, draw=framegray, line width=0.5bp] (529.07,108.53) ellipse [x radius=5.1bp, y radius=5.19bp];  %
\path[draw=black, line width=0.5bp] (514.29,108.53) -- (517.97,108.53);
\fill[black] (517.97,111.53) -- (523.97,108.53) -- (517.97,105.53) -- cycle;  %
\path[draw=teachergreen, line width=1.5bp] (598.06,108.12) ellipse [x radius=6.91bp, y radius=6.79bp];  %
\glyph{593.08}{113.94}{18}{plus}
\path[draw=removered, line width=1.5bp] (670.84,107.77) ellipse [x radius=6.91bp, y radius=6.79bp];  %
\glyph{666.88}{112.85}{18}{minus}
\path[draw=movegray, line width=1.5bp] (868.73,107.77) ellipse [x radius=6.91bp, y radius=6.79bp];  %
\glyph{863.36}{110.23}{10.5}{movearrow}
\node[txt, font=\fs{9}] at (522.31,96.95) {Recolor};
\node[txt, font=\fs{9}] at (599.52,97.15) {Add};
\node[txt, font=\fs{9}] at (675.61,96.45) {Remove};
\node[txt, font=\fs{9}] at (871.63,96.95) {Move};
\path[guide] (519.41,115.93) -- (519.41,126.46);
\path[guide] (598.25,117.64) -- (598.25,126.46);
\path[guide] (670.84,117.77) -- (670.84,126.6);
\path[guide] (868.73,117.56) -- (868.73,126.39);

\path[fill=framegray] (462.15,127.06) rectangle ++(456.9,66.2);  %
\cropimg{439.8}{125.99}{58.11}{58.78}{0.50447}{0.75467}{0.40409}{0.09987}{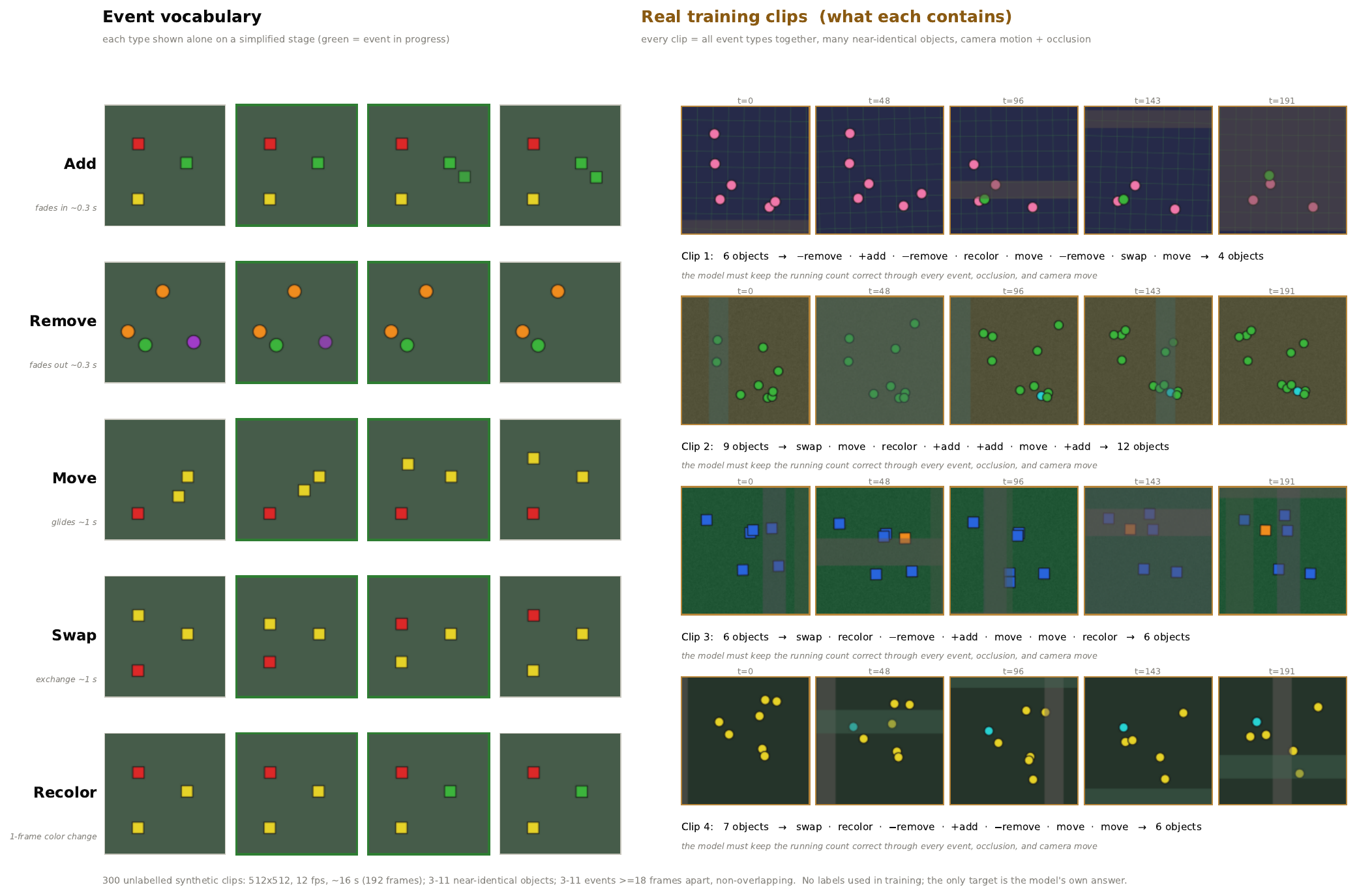}
\cropimg{540.85}{127.03}{57.86}{59.07}{0.60047}{0.75467}{0.30499}{0.09987}{clip-frames}
\cropimg{599.08}{126.06}{58.11}{59.04}{0.70042}{0.75467}{0.20504}{0.09987}{clip-frames}
\cropimg{798.13}{126.94}{58.11}{59.04}{0.79909}{0.75467}{0.10637}{0.09987}{clip-frames}
\cropimg{940.25}{126.13}{58.11}{59.04}{0.8977}{0.75467}{0.00776}{0.09987}{clip-frames}
\cropimg{380.99}{184}{355.62}{10.38}{0}{0.2388}{0}{0.6806}{filmstrip}
\cropimg{881.98}{126.94}{58.11}{59.04}{0.8977}{0.75467}{0.00776}{0.09987}{clip-frames}
\cropimg{380.99}{125.66}{58.11}{61.03}{0.50447}{0.75467}{0.40409}{0.09987}{clip-frames}
\cropimg{740.13}{126.08}{58.11}{59.04}{0.79909}{0.75467}{0.10637}{0.09987}{clip-frames}
\cropimg{681.94}{126.54}{58.11}{59.04}{0.79909}{0.75467}{0.10637}{0.09987}{clip-frames}
\cropimg{642.91}{117.26}{355.62}{10.38}{0}{0.2388}{0}{0.6806}{filmstrip}
\cropimg{642.91}{184}{355.62}{10.38}{0}{0.2388}{0}{0.6806}{filmstrip}
\cropimg{380.99}{117.39}{355.62}{10.38}{0}{0.2388}{0}{0.6806}{filmstrip}
\foreach \x/\ya/\xb/\yb in {382.24/127.64/382.24/185.45, 439.52/127.06/439.52/184.87, 496.22/126.34/496.22/184.16, 541.28/127.06/541.28/184.87, 598.9/127.06/598.9/184.87, 657.2/126.89/657.2/184.7, 682.46/127.77/682.46/184.7, 740.21/127.07/740.21/184, 798.21/127.65/798.21/184.58, 856.21/128.23/856.21/185.16, 882.22/127.65/882.22/184.58, 940.52/127.73/940.52/184.66, 997.41/128.23/996.9/184}
  \draw[frame sep] (\x,\ya) -- (\xb,\yb);

\path[guide] (367.07,193.55) -- (367.07,210.36);  %
\path[draw=studentviolet, line width=1.5bp] (380.99,204.44) -- (998.37,204.44);  %
\foreach \x/\y in {380.99/199.68, 456.2/199.68, 528.32/199.68, 603.53/199.68, 697.17/199.68, 788.99/199.68, 887.99/199.68, 989.09/199.68}
  \fill[studentviolet, rounded corners=1.55bp] (\x,\y) rectangle ++(9.28,9.52);
\path[draw=teachergreen, line width=1.5bp] (380.99,224.93) -- (998.37,224.93);  %
\foreach \x/\y in {380.99/220.17, 418.64/220.17, 458.97/220.17, 496.68/220.17, 534.17/219.84, 573.51/219.84, 612.85/219.84, 652.18/219.84, 691.52/219.84, 730.86/219.84, 770.19/219.84, 802.52/220.17, 840.21/219.84, 877.53/220.17, 916.5/220.17, 951.6/219.84, 989.09/220.17}
  \fill[teachergreen, rounded corners=1.55bp] (\x,\y) rectangle ++(9.28,9.52);
\foreach \xa/\ya/\xb/\yb in {385.63/210.36/385.63/220.17, 460.84/209.2/463.61/220.17, 532.96/209.2/538.81/219.84, 608.17/209.2/617.49/219.84, 701.81/209.2/696.16/219.84, 793.63/209.2/774.83/219.84, 892.63/209.2/921.14/220.17, 993.73/209.2/993.73/220.17}
  \draw[link] (\xa,\ya) -- (\xb,\yb);

\img{471.09}{267.52}{107.46}{73.62}{robot-student}
\node[txt, font=\fs{10.5}, text=studentink] at (540.11,255.35) {Student (Sparse View)};
\img{554.28}{302.36}{50.69}{38.99}{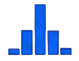}  %
\node[txt, font=\fs{10.5}, text=goodgreen] at (836.77,255.35) {Teacher (Dense View)};
\img{763.37}{269.07}{105.36}{65.81}{robot-teacher}
\img{861.25}{303.45}{41.47}{38.28}{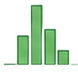}  %
\img{582.79}{249.93}{194.76}{56.51}{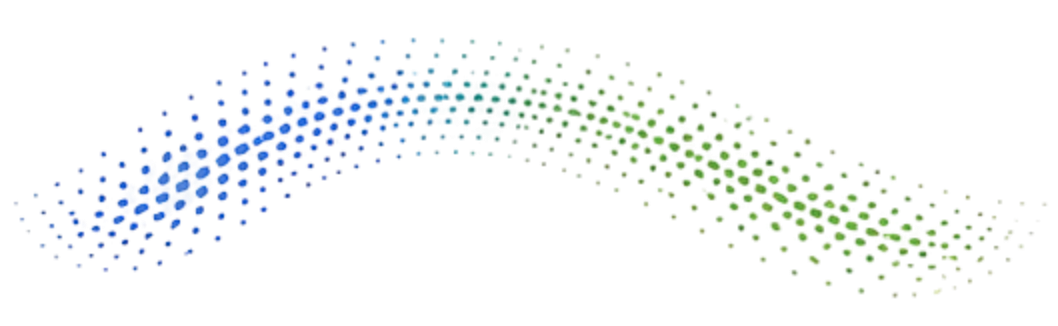}  %
\node[txt, font=\fs{10.5}] at (678.28,308.12) {\textcolor{distillblue}{Self-}\textcolor{teachergreen}{Distillation}};
\imgc{rotate=-90}{682.12}{335.49}{22.55}{22.55}{chevrons}  %

\path[fill=statebg, draw=stateedge, line width=0.75bp, rounded corners=8.88bp] (381.31,350.45) rectangle ++(617.05,95.36);
\node[txt, font=\fs{10.5}, text=statepurple] at (697.87,363.95) {Learns to Maintain Running Scene State};
\path[card, rounded corners=6.86bp] (393.12,373.82) rectangle ++(81.79,67.78);
\cropimg{399.02}{389.53}{69.23}{47.68}{0.50447}{0.75467}{0.40409}{0.09987}{clip-frames}
\node[txt, font=\fs{8}] at (434.02,383.85) {7 yellow objects};
\path[card, rounded corners=6.86bp] (532.69,373.82) rectangle ++(81.79,67.78);
\cropimg{538.81}{389.08}{70.09}{49.29}{0.60806}{0.76086}{0.30498}{0.09987}{clip-frames}
\node[txt, font=\fs{8}] at (574.41,383.85) {6 yellow 1 blue};
\path[card, rounded corners=6.86bp] (672.26,373.82) rectangle ++(81.79,67.78);
\cropimg{677.75}{390.39}{69.63}{46.17}{0.70042}{0.75882}{0.20504}{0.09987}{clip-frames}
\node[txt, font=\fs{8}] at (713.77,384.45) {object moved};
\path[card, rounded corners=6.86bp] (811.82,373.82) rectangle ++(81.79,67.78);
\cropimg{818.11}{389.71}{68.11}{45.71}{0.89932}{0.75862}{0.00834}{0.10948}{clip-frames}
\node[txt, font=\fs{8}] at (853.2,384.25) {1 yellow removed};
\img{487.92}{390.61}{31.28}{31.28}{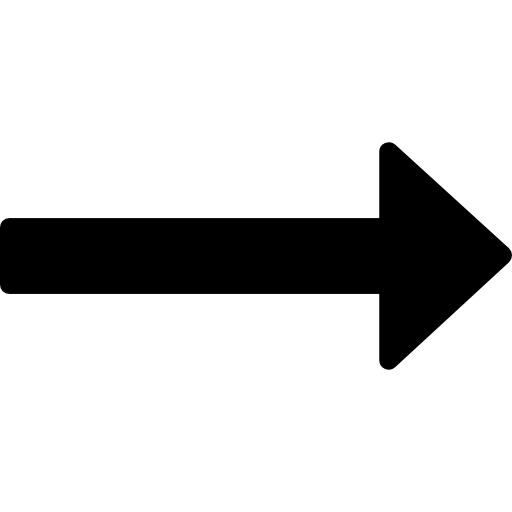}
\img{628.84}{390.03}{31.28}{31.28}{arrow-thick}
\img{769.77}{389.44}{31.28}{31.28}{arrow-thick}
\path[draw=black, line width=1.5bp, dash pattern=on 1.5bp off 1.5bp] (962.63,402.89) -- (917.56,402.89);  %

\begin{scope}
  \clip (406.24,204.44) |- (449.82,295.5) -- (449.82,300) -- (463.32,293.25) -- (449.82,286.5) -- (449.82,291) -| (410.74,204.44) -- cycle;
  \shade[shading=teaserToStudent] (351.79,181.33) rectangle (602.26,431.8);
\end{scope}
\begin{scope}
  \clip (971.82,224.93) |- (919.19,295.5) -- (919.19,300) -- (905.69,293.25) -- (919.19,286.5) -- (919.19,291) -| (967.32,224.93) -- cycle;
  \shade[shading=teaserToTeacher] (777.94,190.78) rectangle (1007.89,436.7);
\end{scope}
\end{tikzpicture}}%
\endgroup
\\[-6pt]
        \caption{\textbf{S$^3$T uses temporal sampling density as self-supervision for visual state tracking.} \emph{Prior methods (left)} rely on labels or external judges, optimize consistency without verifying correctness, use only spatial privileged views, or learn temporal transformations rather than persistent scene state. \emph{S$^3$T (right)} instead self-distills from a denser temporal view into a sparse-view student with shared frozen weights, enabling the model to track evolving object counts and states over time without labels, an external judge, or a separate teacher.}
        \label{fig:teaser}
        \vspace*{12pt}
}]

{\renewcommand{\thefootnote}{*}\footnotetext{Work done during internship at Aalto University.}}

\begin{abstract}
We introduce S$^3$T (Self-Supervised Self-Distillation over Time), which, to the best of our knowledge, is the first fully self-contained self-distillation framework for improving continuous video state tracking. Our method is motivated by the hypothesis that denser temporal observations of the same video enable more accurate estimation of its running state, providing a valuable supervisory signal for long-horizon state tracking in Video-LLMs. Specifically, the temporally dense view serves as the teacher, while a sparse-view student with the same model parameters learns to match the teacher's next-token distribution.
This allows the model to learn directly from its own predictions, without labels, a separate teacher model, or external rewards. Our experiments show that S$^3$T improves VSTAT accuracy by up to $2.61$ points across backbones, while prior self-evolving methods show little improvement. Despite training only on $20$-second synthetic clips, S$^3$T generalizes to videos up to $100$ seconds and transfers to real videos.
\end{abstract}

\begin{figure*}[t]
\centering

\newsavebox{\myfullbox}
\sbox{\myfullbox}{%
\begin{minipage}{\textwidth}
\small
\setlength{\tabcolsep}{0pt}
\renewcommand{\arraystretch}{1}
\arrayrulecolor{rulegray}
\renewcommand{\makecell}[1]{\parbox[c]{2.8cm}{\centering#1}}
\newcommand{\pad}[1]{\tikz[inner sep=0]\node[text height=1.15cm]{#1};}

\newcommand{\tabvstretch}{1.12}

\newcommand{\thumbmag}{1.0}
\newlength{\thumbwd}\setlength{\thumbwd}{2.6cm}
\newlength{\thumbht}
\newcommand{\fitthumb}[1]{%
  \setbox0=\hbox{\includegraphics[width=\thumbwd]{#1}}%
  \ifdim\ht0<\thumbht \global\thumbht=\ht0 \fi}
\newcommand{\thumb}[1]{%
  \makebox[\thumbwd][c]{\includegraphics[height=\thumbht]{#1}}}

\newcommand{\videocell}[3][0pt]{%
  \parbox[c]{4.29cm}{\centering
    \vspace*{1.0mm}%
    \textbf{#2}\par\vspace{1pt}%
    \vspace*{#1}%
    \includegraphics[width=3.5cm]{#3}\par\vspace*{0.6mm}}}

\newcommand{\questioncell}[1]{%
  \parbox[c]{4.19cm}{%
    \vspace*{0.7mm}%
    \tiny\raggedright\emph{#1}\par
    \vspace*{0.7mm}}}
\newcommand{\topmodelcell}[1]{%
  \parbox[c]{2.10cm}{\centering#1}}

\newsavebox{\myboxC}
\sbox{\myboxC}{%
\resizebox{\textwidth}{!}{%
\scalebox{1}[\tabvstretch]{%
\scriptsize
\begin{tabular}{|>{\centering\arraybackslash}m{2.20cm}|*{4}{>{\centering\arraybackslash}m{4.39cm}|}}
\hline
\textbf{FULL VIDEO}
&
\videocell{Cup position}{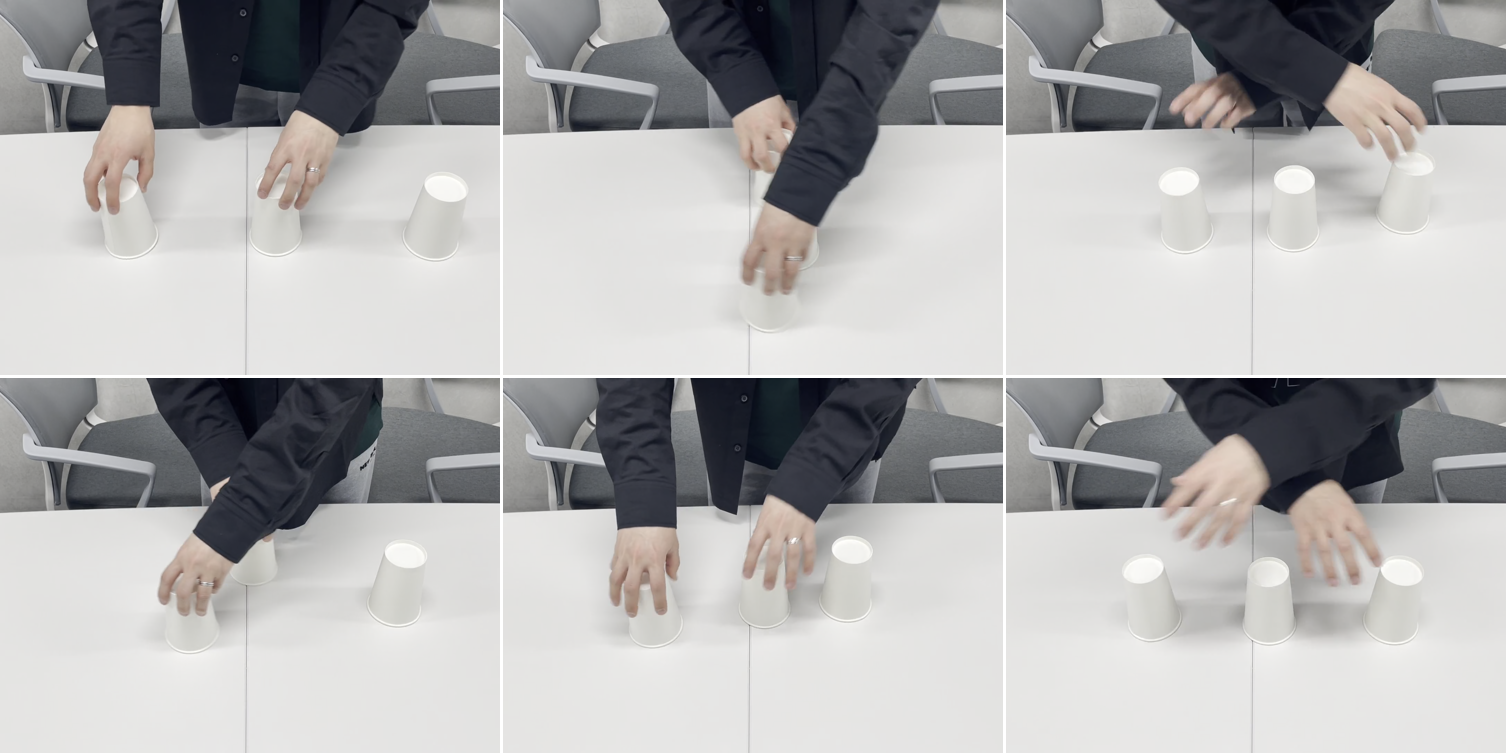}
&
\videocell{Cup stacking}{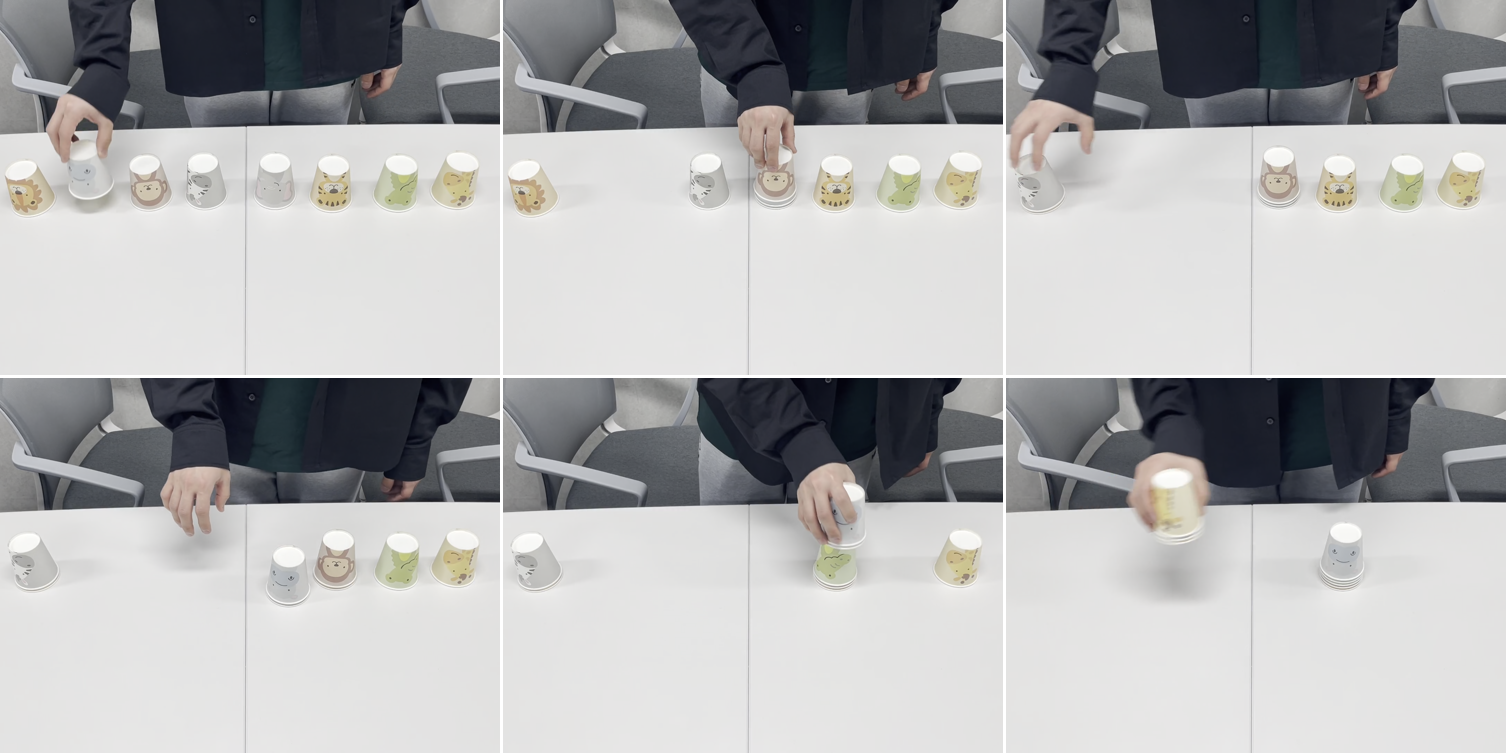}
&
\videocell[0.3mm]{Funnel ball}{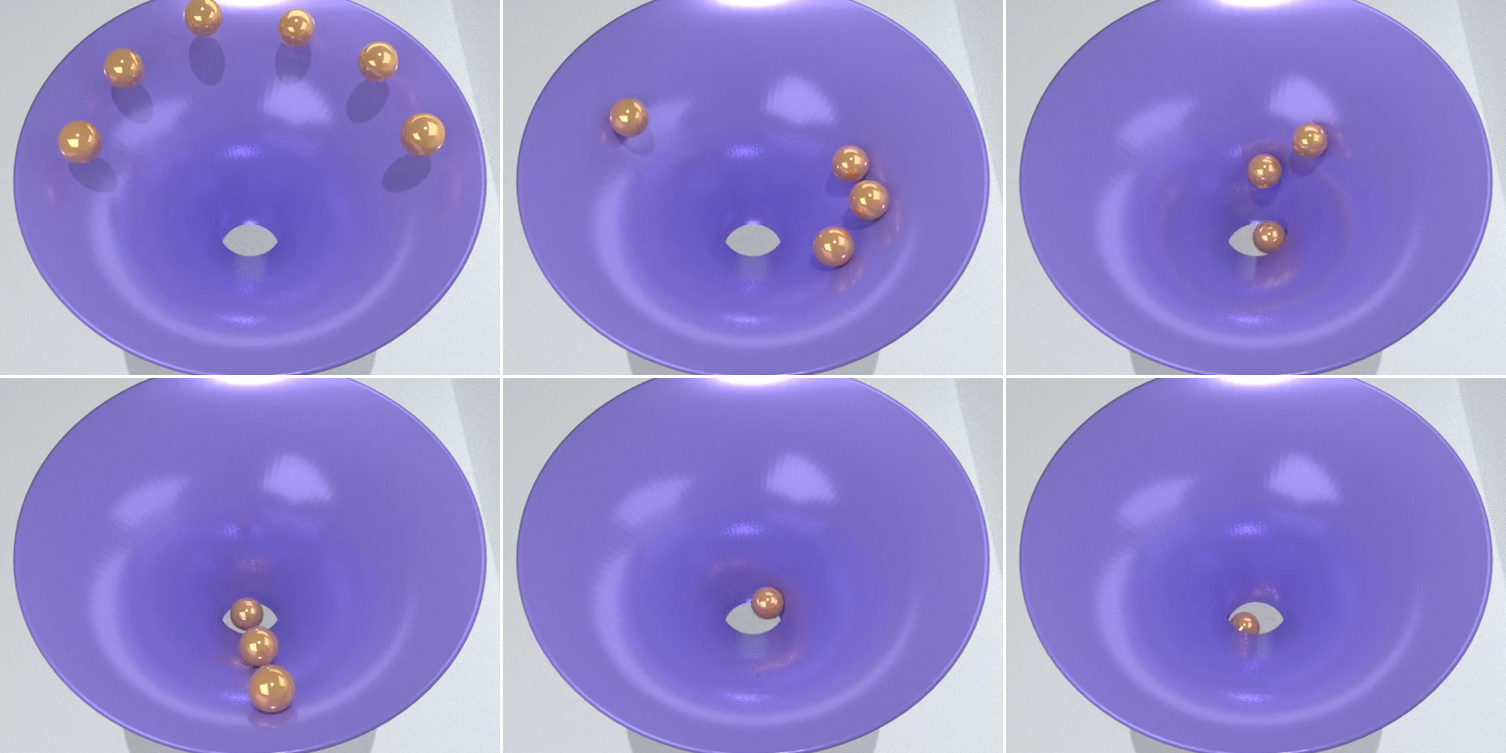}
&
\videocell[0.3mm]{Tilted box}{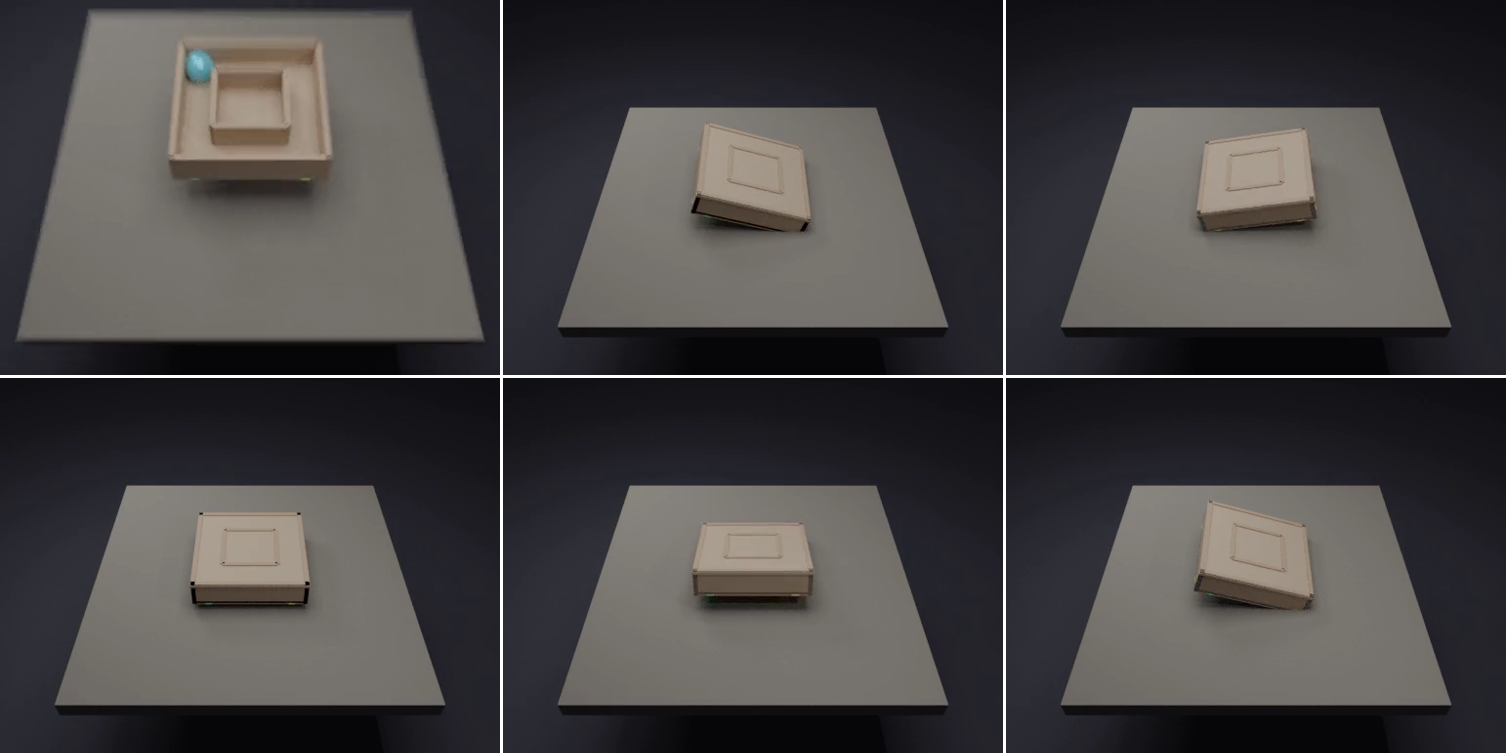}
\\ \hline

\textcolor{headblue}{\textbf{QUESTION}}
&
\questioncell{At the end of the video, which position is the cup containing the smaller cup in? Use the viewer's perspective: (A)~Right, (B)~Left, (C)~Center.}
&
\questioncell{What animal is drawn on the cup that is seventh from the bottom? (A)~Monkey, (B)~Lion, (C)~Tiger, (D)~Zebra.}
&
\questioncell{Balls are indexed from left to right in the last frame before any release. Which ball took the longest to fall through the hole after its own release? (A)~ball\_3, (B)~ball\_1, (C)~ball\_2, (D)~ball\_6.}
&
\questioncell{A ball starts at corner 1, the top-left. After the box is closed and tilted six times, which corner contains the ball? Corner mapping: 1~top-left, 2~top-right, 3~bottom-right, 4~bottom-left.}
\\ \hline
\noalign{\rowstretch{1.4}}
\rowcolor{lightgreen}
\textcolor{green!40!black}{\textit{S$^3$T} (ours)}
& \ok\;A & \ok\;D & \ok\;B & \ok\;2
\\ \hline

\textcolor{headblue}{\topmodelcell{LLaVA-OV-2-8B\\[-1pt]\tiny(base)}}
& \bad\;C & \bad\;C & \bad\;A & \bad\;4
\\ \hline

\textcolor{headblue}{\topmodelcell{Qwen3-VL-8B}}
& \bad\;C & \bad\;C & \bad\;A & \bad\;4
\\ \hline

\textcolor{headblue}{\topmodelcell{Molmo2-8B}}
& \bad\;C & \bad\;C & \bad\;A & \bad\;3
\\ \hline

\textcolor{headblue}{\topmodelcell{InternVL3.5-8B}}
& \bad\;C & \bad\;A & \bad\;D & \bad\;4
\\ \hline

\textcolor{headblue}{\topmodelcell{Video-Zero-8B\\[-1pt]\tiny(self-evolving)}}
& \bad\;C & \bad\;C & \bad\;A & \bad\;4
\\ \hline
\end{tabular}}}}

\begin{tikzpicture}[baseline=0]
  \begin{scope}
    \clip[rounded corners=1mm]
      (1pt,1pt-\dp\myboxC) rectangle (\wd\myboxC+1pt,\ht\myboxC-1pt);
    \node[inner sep=0,anchor=base west] at (0,0) {\usebox{\myboxC}};
  \end{scope}

  \draw[rulegray,rounded corners=1mm]
    (0,-\dp\myboxC) rectangle (\wd\myboxC,\ht\myboxC);
\end{tikzpicture}

\par\vspace{0.5em}

\setlength{\thumbht}{50cm}%
\fitthumb{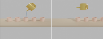}%
\fitthumb{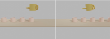}%
\fitthumb{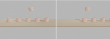}%
\fitthumb{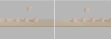}%
\fitthumb{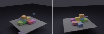}%
\fitthumb{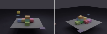}%
\fitthumb{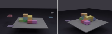}%
\fitthumb{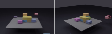}%
\setlength{\thumbht}{\thumbmag\thumbht}%
\setlength{\thumbwd}{\thumbmag\thumbwd}%

\begin{minipage}[b]{0.485\linewidth}
\centering
{\tiny\raggedright
\textbf{Q:} {\em There is first a water-pouring phase and then a separate espresso-pouring phase. A cup counts as successful \\ only if it receives both water in first phase and espresso in second phase. How many cups were successful?}\par}
\newsavebox{\myboxA}
\sbox{\myboxA}{%
\resizebox{\linewidth}{!}{%
\scalebox{1}[\tabvstretch]{%
\begin{tabular}{|c|c|c|c|c|}
\hline
\textbf{VIDEO PREFIX}
&
\pad{\thumb{figs/teaser_quals/espresso1}}
&
\pad{\thumb{figs/teaser_quals/espresso2}}
&
\pad{\thumb{figs/teaser_quals/espresso3}}
&
\pad{\thumb{figs/teaser_quals/espresso4}}
\\

&
\scriptsize\textbf{first 25\%} (0--5.0\,s)
&
\scriptsize\textbf{first 50\%} (0--10.0\,s)
&
\scriptsize\textbf{first 75\%} (0--15.0\,s)
&
\scriptsize\textbf{first 100\%} (0--20.0\,s)
\\ \hline
\noalign{\rowstretch{1.5}}
\rowcolor{lightblue}
\textcolor{headblue}{TIME (s)}
&
\makecell{0.0 -- 5.0\\[-1pt]\scriptsize(end at 5.0 s)}
&
\makecell{0.0 -- 10.0\\[-1pt]\scriptsize(end at 10.0 s)}
&
\makecell{0.0 -- 15.0\\[-1pt]\scriptsize(end at 15.0 s)}
&
\makecell{0.0 -- 20.0\\[-1pt]\scriptsize(end at 20.0 s)}
\\ \hline

\rowcolor{lightgreen}
\textcolor{green!40!black}{\textit{S$^3$T} (ours)}
& 0 & 1 & 1 & \ok\;1
\\ \hline

\textcolor{headblue}{\makecell{LLaVA-OV-2-8B\\[-1pt]\scriptsize(base)}}
& 0 & 5 & 5 & \bad\;5
\\ \hline

\textcolor{headblue}{Qwen3-VL-8B}
& 1 & 0 & 0 & \bad\;2
\\ \hline

\textcolor{headblue}{Molmo2-4B}
& 0 & 0 & 2 & \bad\;0
\\ \hline

\textcolor{headblue}{\makecell{Video-Zero-8B\\[-1pt]\scriptsize(self-evolving)}}
& 1 & 0 & 0 & \bad\;0
\\ \hline
\end{tabular}}}}

\begin{tikzpicture}[baseline=0]
  \begin{scope}
    \clip[rounded corners=1mm]
      (1pt,1pt-\dp\myboxA) rectangle (\wd\myboxA+1pt,\ht\myboxA-1pt);
    \node[inner sep=0,anchor=base west] at (0,0) {\usebox{\myboxA}};
  \end{scope}

  \draw[rulegray,rounded corners=1mm]
    (0,-\dp\myboxA) rectangle (\wd\myboxA,\ht\myboxA);
\end{tikzpicture}
\end{minipage}
\hfill
\begin{minipage}[b]{0.485\linewidth}
\centering
{\tiny\raggedright
\textbf{Q:} {\em The video first shows an initial central block structure. Then the camera keeps moving while a robot hand makes several add/remove attempts. After all actions finish, how many blocks remain in the central structure?}\par}
\raggedleft
\newsavebox{\myboxB}
\sbox{\myboxB}{%
\resizebox{\linewidth}{!}{%
\scalebox{1}[\tabvstretch]{%
\begin{tabular}{|c|c|c|c|c|}
\hline
\textbf{VIDEO PREFIX}
&
\pad{\thumb{figs/teaser_quals/blocks1}}
&
\pad{\thumb{figs/teaser_quals/blocks2}}
&
\pad{\thumb{figs/teaser_quals/blocks3}}
&
\pad{\thumb{figs/teaser_quals/blocks4}}
\\

&
\scriptsize\textbf{first 25\%} (0--5.0\,s)
&
\scriptsize\textbf{first 50\%} (0--10.0\,s)
&
\scriptsize\textbf{first 75\%} (0--15.0\,s)
&
\scriptsize\textbf{first 100\%} (0--20.0\,s)
\\ \hline
\noalign{\rowstretch{1.5}}
\rowcolor{lightblue}
\textcolor{headblue}{TIME (s)}
&
\makecell{0.0 -- 5.0\\[-1pt]\scriptsize(end at 5.0 s)}
&
\makecell{0.0 -- 10.0\\[-1pt]\scriptsize(end at 10.0 s)}
&
\makecell{0.0 -- 15.0\\[-1pt]\scriptsize(end at 15.0 s)}
&
\makecell{0.0 -- 20.0\\[-1pt]\scriptsize(end at 20.0 s)}
\\ \hline

\rowcolor{lightgreen}
\textcolor{green!40!black}{\textit{S$^3$T} (ours)}
& 3 & 5 & 8 & \ok\;10
\\ \hline

\textcolor{headblue}{\makecell{LLaVA-OV-2-8B\\[-1pt]\scriptsize(base)}}
& 6 & 6 & 7 & \bad\;9\phantom{0}
\\ \hline

\textcolor{headblue}{Qwen3-VL-8B}
& 5 & 4 & 5 & \bad\;12
\\ \hline

\textcolor{headblue}{Molmo2-4B}
& 4 & 8 & 10 & \bad\;9\phantom{0}
\\ \hline

\textcolor{headblue}{\makecell{Video-Zero-8B\\[-1pt]\scriptsize(self-evolving)}}
& 5 & 9 & 9 & \bad\;13
\\ \hline
\end{tabular}}}}

\begin{tikzpicture}[baseline=0]
  \begin{scope}
    \clip[rounded corners=1mm]
      (1pt,1pt-\dp\myboxB) rectangle (\wd\myboxB+1pt,\ht\myboxB-1pt);
    \node[inner sep=0,anchor=base west] at (0,0) {\usebox{\myboxB}};
  \end{scope}

  \draw[rulegray,rounded corners=1mm]
    (0,-\dp\myboxB) rectangle (\wd\myboxB,\ht\myboxB);
\end{tikzpicture}
\end{minipage}
\end{minipage}}%
\scalebox{1}[0.9]{\usebox{\myfullbox}}\par

 \vspace*{-2pt}%
\caption{\textbf{Current Video-LLMs struggle to maintain visual state over time, while S$^3$T consistently tracks the evolving scene.} Across diverse VSTAT examples (top), strong open-source and self-evolving models fail on questions requiring persistent state reasoning, despite observing the full video. Prefix analysis (bottom) further shows that their predictions fluctuate as more evidence arrives, whereas S$^3$T progressively accumulates the correct state and preserves it through the end of the sequence.}
\label{fig:qualitative}
\end{figure*}

\section{Introduction}
\label{sec:intro}
Recent progress in Video Large Language Models (Video-LLMs) has improved multimodal reasoning and enabled complex question answering over dynamic visual content~\cite{vidllm1,vidllm2,vidllm3,vidllm4}. However, visual state tracking remains a major limitation, as models must maintain accurate running counts and scene states while objects are added, removed, moved, or occluded over time~\cite{lim1,lim2}. VSTAT~\cite{vstat} is a benchmark designed to test this cumulative state reasoning, and it shows that human performance reaches $90.5\%$, while the strongest open-source models score only $34$--$36\%$. It also notes that current benchmarks focus mainly on action recognition and moment localization, leaving cumulative state tracking weakly represented in the training signal~\cite{bai2025qwen3vltechnicalreport,wang2025internvl3,action1}. These models often fail even on seemingly simple state changes and continue revising their predictions as more of the video is revealed, rather than maintaining a stable running state (see~\cref{fig:qualitative}). A likely reason is that existing training objectives, whether supervised fine-tuning for single-answer correctness~\cite{sup1,sup2,sup3,mahmood2024vurf} or contrastive alignment of video--text pairs~\cite{contr1,contr2,contr3,contr4}, do not require models to accumulate evidence across the full clip. 

Reinforcement learning with verifiable rewards improves video reasoning but depends on ground-truth labels or expensive judge models~\cite{rlvr1,rlvr2,rlvr3,rlvr4}. Self-distillation reduces this dependence by using the model's own outputs as supervision~\cite{sdpo,cvpd}. However, existing video self-distillation methods still require ground-truth labels, judge models, or annotated evidence~\cite{lin2026visd,wang2026videoopsdexploitingprivilegedvisual,wang2026lookmattersonpolicyselfdistillation,zhu2026mesdmultievidenceselfdistillationvideollm}. Recent self-evolving frameworks~\cite{evo1,evo2,evo3,evo4,kulkarni2025videosaviselfalignedvideolanguage} instead use questioner--solver co-evolution, proposer--solver coupling, or self-critique, but target general reasoning or temporal grounding, which prioritizes event localization over maintaining a running visual state.\looseness-1

We present S$^3$T (Self-Supervised Self-Distillation over Time), the first fully self-contained self-distillation framework for improving continuous state tracking in videos without external supervision. The central idea is to use temporal sampling density as privileged information. We hypothesize that denser sampling provides more evidence about the scene state (such as object counts through occlusion) than sparse sampling. S$^3$T uses this denser view as a teacher and distills its next-token distribution into the sparse-view student, using only the model's self-generated outputs. Our method requires no labels, separate teacher network, reward model, or tracker (see~\cref{fig:teaser}). On VSTAT, S$^3$T improves accuracy by $2.61$ points on LLaVA-OV-2-8B and $2.44$ on Qwen3-VL-8B, with gains across Count, Atomic, and Sequence tasks. Despite training only on $20$-second clips, it achieves larger gains on longer VSTAT videos, generalizing to durations of up to $100$ seconds. These improvements also transfer to real videos while preserving general video-understanding performance.

\paragraph{Contributions.} 
To summarize, our contributions are:
\begin{itemize}
\item We introduce temporal sampling density as privileged information for self-distillation in Video-LLMs, enabling supervision from unlabeled videos without external annotations or reward models.

\item We propose S$^3$T, the first fully self-contained self-distillation framework for Video-LLMs that improves cumulative-state reasoning while preserving general video-understanding performance.

\item We show that S$^3$T generalizes across Video-LLMs and that the capability learned from unlabeled synthetic clips transfers to real video and to videos five times longer than the training clips, without retraining on real data.
\end{itemize}

\section{Related work}
\label{sec:relatedwork}
\paragraph{Video understanding and state tracking.}

Video-LLMs have evolved from early cross-modal alignment methods~\cite{videobert,clip4clip,frozenintime,alayrac2022flamingo} to recent systems~\cite{llavavideo,videochatgpt,song2024moviechat,lin2024vila} that extend large language models to video through frame sampling and temporal aggregation. These models support open-ended dialogue~\cite{openended1,openended2} and temporal reasoning~\cite{temporal1,temporal2}, but are mainly trained for action recognition, moment localization, and video question answering~\cite{vqa1,vqa2}. Such tasks often involve finding frames containing the answer rather than maintaining a running scene state throughout the video. VSTAT~\cite{vstat} highlights this gap, with current open-source Video-LLMs achieving only $34$--$36\%$ on state tracking, compared to about $90.5\%$ for humans.
Temporal self-supervised objectives such as frame-order prediction~\cite{frameorder,frame2}, arrow-of-time classification~\cite{arrow,arrow2}, pace estimation~\cite{pace,pace2}, and masked video modeling~\cite{videomae,wu2025stm} learn temporal structure from chronology, motion, or missing frames. But, they do not learn to track running counts or evolving scene states across long sequences and occlusions.

\paragraph{Self-distillation and self-evolution.}

Visual representations can be learned without human annotations through self-generated supervision or privileged information available only during training~\cite{evolmm,vise,asg}. For images, CVPD~\cite{cvpd} and Vision-OPD~\cite{visionopd} use privileged spatial views, such as high-resolution crops or zooms, to improve fine-grained recognition. However, these methods exploit spatial redundancy rather than accumulating temporal evidence across frames. Recent self-evolving methods use model-generated supervision to improve Video-LLMs. Video-Zero~\cite{evo3}, EvoGround~\cite{evo2}, CurEvo~\cite{evo1}, and VideoSAVi~\cite{kulkarni2025videosaviselfalignedvideolanguage} focus on question difficulty, answer consistency, self-critiqued answers, evidence discovery, or temporal grounding rather than cumulative-state tracking. Their signals encourage finding relevant moments or producing consistent answers, but not integrating evidence across an entire video to maintain a running scene state. Recent video self-distillation methods, including VISD~\cite{lin2026visd}, Video-OPSD~\cite{wang2026videoopsdexploitingprivilegedvisual}, Clue-OPSD~\cite{wang2026lookmattersonpolicyselfdistillation}, and MeSD~\cite{zhu2026mesdmultievidenceselfdistillationvideollm}, are also related but rely on ground-truth answers, external judges, or annotated evidence such as keyframes and clue intervals. Thus, to the best of our knowledge, no prior method directly targets cumulative-state tracking while remaining fully self-contained.

\section{Methods}
\label{sec:method}

We present the proposed framework by first describing the procedurally generated unlabeled clips used to expose the model to visual state changes~(\cref{sec:method:data}), followed by the two temporal views and distillation objective~(\cref{sec:method:tsd}). S$^3$T samples each clip at two temporal densities, using the denser view to teach the sparser one. Both views share the same frozen base model and trainable LoRA adapter, using only the model's own predictions for supervision.

\subsection{Synthesizing training data}
\label{sec:method:data}
We train on a fixed set of $300$ unlabeled clips generated using our procedural video generator, \emph{StateGen}. Each clip is a $512{\times}512$ video at $12$ FPS and lasts $20$ seconds, giving $T=240$ frames. Scenes contain $6$--$15$ near-identical entities on a textured background and evolve through $7$--$11$ non-overlapping events sampled from a fixed set of add, remove, move, swap, and recolor operations. \emph{StateGen} randomly samples the number of entities and events, as well as event timings, to produce clips of varying difficulty (sampling details in \cref{app:data}). Consecutive events are separated by at least $13$ frames, while occlusion and global camera motion continue throughout the clip (\cref{fig:data}). Because entities can appear and disappear over time, recovering the correct scene state requires information from the full sequence. The clips contain no state annotations, question-answer pairs, or captions. The model receives only RGB frames, with all supervision coming from its own generated outputs (\cref{sec:method:tsd}). We use procedurally generated videos because temporal-density distillation requires clips with meaningful state changes over time, while manually curating and annotating such data at scale would be expensive. S$^3$T instead learns directly from unlabeled clips through self-generated supervision. Despite using short clips to keep training inexpensive, S$^3$T transfers cumulative-state tracking to real videos (\cref{sec:exp:transfer}) and generalizes to videos up to five times longer than those seen during training (\cref{sec:exp:horizon}).

\begin{figure}[t]
  \centering

  \setlength{\figW}{\columnwidth}%
  \setlength{\vocabw}{0.417\figW}%
  \setlength{\trainw}{0.522\figW}%
  \pgfmathsetlength{\vocabh}{\vocabw/4.236}%
  \pgfmathsetlength{\trainh}{\trainw/5.102}%
  \setlength{\capdrop}{0.05\trainh}%
  \setlength{\rowgap}{0.16\trainh}%
  \setlength{\capWmax}{0pt}%
  \measurecapW{\clipseqA}\measurecapW{\clipseqB}%
  \measurecapW{\clipseqC}\measurecapW{\clipseqD}%
  \pgfmathsetmacro{\capscale}{0.97*\trainw/\capWmax}%
  \typeout{AUTO-CAPSCALE = \capscale (trainw=\the\trainw, widest=\the\capWmax)}%
  \sbox{\capbox}{\clipline{\clipseqB}}%
  \setlength{\capH}{\ht\capbox}\addtolength{\capH}{\dp\capbox}%
  \setlength{\capHlast}{\capH}%
  \setlength{\trainpitch}{\trainh}%
  \addtolength{\trainpitch}{\capdrop}%
  \addtolength{\trainpitch}{\capH}%
  \addtolength{\trainpitch}{\rowgap}%
  \setlength{\stackdepth}{3\trainpitch}%
  \addtolength{\stackdepth}{\trainh}%
  \addtolength{\stackdepth}{\capdrop}%
  \addtolength{\stackdepth}{\capHlast}%
  \pgfmathsetlength{\vocabpitch}{(\stackdepth-\vocabh)/4}%
  \setlength{\vocabx}{0.030\figW}%
  \setlength{\divx}{0.462\figW}%
  \setlength{\trainx}{0.478\figW}%
  \setlength{\titley}{0.10\vocabh}%
  \begin{tikzpicture}[inner sep=0pt, outer sep=0pt]
  \tikzset{
    rowimg/.style    ={anchor=north west},
    rowlabel/.style  ={rotate=90, anchor=south, font=\figfont\strut},
    paneltitle/.style={anchor=south, font=\figfont},
    clipcap/.style   ={anchor=north, inner sep=0pt},
  }

  \foreach \row [count=\i from 0] in {add,remove,move,swap,recolor}{
    \pgfmathsetlength{\rowy}{-\i*\vocabpitch}
    \node[rowimg] (\row) at (\vocabx,\rowy)
      {\includegraphics[width=\vocabw]{figs/data_assets/\row.pdf}};
  }
  \foreach \row/\lbl in {add/Add, remove/Remove, move/Move, swap/Swap,
                         recolor/Recolor}{
    \node[rowlabel] at ($(\row.north west)!0.5!(\row.south west)+(-0.6mm,0)$)
      {\lbl};
  }

  \foreach \n [count=\j from 0] in {1,2,3,4}{
    \pgfmathsetlength{\rowy}{-\j*\trainpitch}
    \node[rowimg] (train\n) at (\trainx,\rowy)
      {\includegraphics[width=\trainw]{figs/data_assets/train\n.pdf}};
  }
  \node[clipcap] at ($(train1.south)-(0,\capdrop)$) {\clipline{\clipseqA}};
  \node[clipcap] at ($(train2.south)-(0,\capdrop)$) {\clipline{\clipseqB}};
  \node[clipcap] at ($(train3.south)-(0,\capdrop)$) {\clipline{\clipseqC}};
  \node[clipcap] at ($(train4.south)-(0,\capdrop)$) {\clipline{\clipseqD}};

  \node[paneltitle] at ([xshift=0.5\vocabw]\vocabx,\titley) {Event Vocabulary};
  \node[paneltitle] at ([xshift=0.5\trainw]\trainx,\titley) {Training Clips};

  \draw[dashed, dash pattern=on 3pt off 3pt, black!55]
    (\divx,0) -- (\divx,-\stackdepth);

  \end{tikzpicture}
   \caption{\textbf{Our training data generator, \emph{StateGen}.} \textit{Left:} The five event types: add, remove, move, swap, and recolor. \textit{Right:} Four example training clips shown at five uniformly spaced time points. Scenes contain near-identical objects undergoing these events under occlusion and global camera motion. Object counts below each strip are for visualization only and are not used in training.
  }
  \vspace{-5pt}
  \label{fig:data}
\end{figure}

\begin{figure*}[t]
  \centering
\begingroup
\graphicspath{{figs/architecture_assets/}}%
\def\figtextscale{1}%
\definecolor{adaptergray}{HTML}{808080}%
\definecolor{densegreen}{HTML}{275417}%
\definecolor{densetick}{HTML}{8ED973}%
\definecolor{divbg}{HTML}{FBE3D6}%
\definecolor{divedge}{HTML}{80350E}%
\definecolor{gradred}{HTML}{C00000}%
\definecolor{lorastudent}{HTML}{CAEEFB}%
\definecolor{lorateacher}{HTML}{C2F1C8}%
\definecolor{prefbg}{HTML}{D9D9D9}%
\definecolor{pstudentedge}{HTML}{156082}%
\definecolor{pteacheredge}{HTML}{196B24}%
\definecolor{questionbg}{HTML}{E9E1F7}%
\definecolor{refbg}{HTML}{F2F2F2}%
\definecolor{refgray}{HTML}{595959}%
\definecolor{sggray}{HTML}{A6A6A6}%
\definecolor{sparsetab}{HTML}{A6CAEC}%
\definecolor{sparsetick}{HTML}{4E95D9}%
\definecolor{studentbg}{HTML}{C1E5F5}%
\definecolor{studentink}{HTML}{104862}%
\definecolor{studentpurple}{HTML}{590EE3}%
\definecolor{teacherbg}{HTML}{D9F2D0}%
\definecolor{teacherink}{HTML}{3B7D23}%
\definecolor{viewsbg}{HTML}{DCEAF7}%
\definecolor{vlmbg}{HTML}{F9F3EC}%
\definecolor{vlmedge}{HTML}{BFBFBF}%
\def\fs#1{\dimen0=\figtextscale\dimexpr#1bp\relax\fontsize{\the\dimen0}{1.2\dimen0}\selectfont}%
\def\fsm#1{\dimen0=\figtextscale\dimexpr#1pt\relax\fontsize{\the\dimen0}{1.2\dimen0}\selectfont}%
\def\img#1#2#3#4#5{\node[inner sep=0pt, anchor=north west] at (#1,#2)
  {\includegraphics[width=#3bp, height=#4bp]{#5}};}%
\def\imgc#1#2#3#4#5#6{\node[inner sep=0pt, #1] at (#2,#3)
  {\includegraphics[width=#4bp, height=#5bp]{#6}};}%
\def\cropimg#1#2#3#4#5#6#7#8#9{%
  \begin{scope}
    \clip (#1,#2) rectangle ++(#3,#4);
    \pgfmathsetmacro\cropW{#3/(1-#5-#7)}%
    \pgfmathsetmacro\cropH{#4/(1-#6-#8)}%
    \node[inner sep=0pt, anchor=north west] at ({#1-#5*\cropW},{#2-#6*\cropH})
      {\includegraphics[width=\cropW bp, height=\cropH bp]{#9}};
  \end{scope}}%
\def\icon#1#2#3#4#5{\begin{scope}[shift={(#1,#2)}]\pic[x=#3bp/96, y=-#4bp/96] {#5};\end{scope}}%
\tikzset{
  txt/.style       = {inner sep=0pt, outer sep=0pt, anchor=base, align=center},
  frame sep/.style = {draw=black, line width=1.5bp},      %
  pics/nosign/.style={code={\fill[refgray]
      (48,10) .. controls (27.01,9.99) and (9.98,27.01) .. (9.98,48) .. controls (9.97,68.99) and (26.99,86.02) .. (47.98,86.02) .. controls (68.97,86.03) and (85.99,69.01) .. (86,48.02) .. controls (86,48.02) and (86,48.01) .. (86,48.01) .. controls (86.01,27.03) and (69.01,10.01) .. (48.03,10) .. controls (48.02,10) and (48.01,10) .. (48,10) -- cycle (48,84.02) .. controls (28.13,84.03) and (12.01,67.93) .. (12,48.06) .. controls (11.99,38.84) and (15.52,29.97) .. (21.87,23.28) -- (72.73,74.14) .. controls (66.06,80.49) and (57.21,84.03) .. (48,84.02) -- cycle (74.14,72.73) -- (23.28,21.87) .. controls (37.71,8.2) and (60.48,8.83) .. (74.14,23.26) .. controls (87.28,37.13) and (87.28,58.85) .. (74.14,72.73) -- cycle;}},                                       %
}%
\resizebox{\linewidth}{!}{%
}%
\endgroup
   \caption{\textbf{Overview of S$^3$T.} Each clip is uniformly sampled into a sparse $12$-frame view and a dense $24$-frame view. A frozen video-language model with a single trainable LoRA adapter is used in three roles: the \emph{Student} on the sparse view, the \emph{Teacher} on the dense view, and the \emph{Reference} on the sparse view with the adapter disabled. Student and Teacher use the same current adapter and differ only in frame density and gradient flow, with gradients passing through the Student while the Teacher is detached. All three score the same self-generated answer. The loss distills the Teacher into the Student while anchoring it to the base model, and only the single LoRA adapter is updated.
  }
  \label{fig:architecture}
\end{figure*}

\subsection{Temporal self-distillation}
\label{sec:method:tsd}
Let $f_\theta$ be the frozen base model with parameters $\theta$, and
$f_{\theta+\phi}$ the same model equipped with a single trainable LoRA
adapter $\phi$, applied to both the vision encoder and the language model. The same adapter is used for both student and teacher;
these differ only in temporal sampling density and whether gradients are
retained. We write $f_{\theta+\phi}(\cdot \mid x, k)$ for its next-token
distributions from $k$ frames of clip $x$.

\paragraph{Two temporal views.}
For a clip $x$ decoded into $T$ native frames, each view uniformly samples frames across $[0,T-1]$. The student uses $k_s=12$ frames and the teacher $k_t=24$, both covering the full clip without cropping or shorter windows. We hypothesize that denser sampling better captures changes over time, helping the teacher recover the cumulative scene state. We test this before training by scoring the frozen base model's answers under both views against the known end states of $300$ held-out \emph{StateGen} clips. With $24$ instead of $12$ frames, the model predicts more state tokens correctly and gives the exact count on more clips, so the teacher's view gives a more accurate estimate of the scene state (\cref{app:premise}). We treat the additional frames as privileged information (LUPI; \cite{vapnik2015learning, lopez2015unifying}), available only to the teacher during training. The student learns from the teacher's predictive distribution, while the resulting model is evaluated with the standard frame budget (\cref{sec:experiments}). Teacher and student are two passes through the same $f_{\theta+\phi}$ and single adapter $\phi$, with no separate teacher model or adapter. They differ only in frame density and gradient flow: the sparse student carries gradients, while the dense teacher is detached. The goal is to make the teacher more informative while keeping its readout recoverable from the sparse student view (see~\cref{sec:discussion}).

\paragraph{On-policy, self-generated target.}
We use a fixed probe question $q$, chosen in advance and shared by both views of each clip. The question is neither generated by the model nor derived from the clip; only the answer is self-generated. This keeps the question identical across views, allowing their answer distributions to be compared based on frame density alone. Visual state tracking requires knowing the current scene state and tracking how it changes over time~\cite{vstat}. We train two probes in separate runs to capture both. The state probe $q_1$ asks which objects are visible, their colors, and their count. The change probe $q_2$ alternates these full-state questions with shorter questions about what changed during the clip. Both use fixed templates filled with the object names of the training domain. Since we generate the clips ourselves, we know these names in advance without needing annotations. Merging the two runs outperforms either alone (\cref{sec:exp:ablation}). After training, we average the weight updates from the two LoRA adapters~\cite{souping}, scale the result by $\alpha$, and add it to the base weights (\emph{Probe Merging}, \cref{fig:architecture}). The merged model retains the base model's size and inference cost. The exact probes are provided in \cref{app:souping}.

\paragraph{Objective.}
The model is trained to match the teacher's next-token distribution while
remaining anchored to the base, and both terms use the same token-averaged Jensen-Shannon divergence (JSD;~\cite{lin1991divergence})
over the answer region,
\begin{equation}
\footnotesize
  \begin{aligned}
    D_{\mathrm{JSD}}\!\big(P \,\big\|\, Q\big)
    &\;=\;
    \frac{1}{T_a}\sum_{j=1}^{T_a}
    \Big[\tfrac12 D_{\mathrm{KL}}\!\big(p_j \,\big\|\, m_j\big)
       + \tfrac12 D_{\mathrm{KL}}\!\big(q_j \,\big\|\, m_j\big)\Big], \\
    m_j &\;=\; \tfrac12\big(p_j + q_j\big),
  \end{aligned}
  \label{eq:jsd}
\end{equation}
computed in bits ($\log_2$), so each position contributes at most one bit. We use JSD rather than Kullback–Leibler divergence (KLD;~\cite{kullback1951information}) because it is symmetric and bounded on $[0,1]$, which keeps the objective and its gradients well behaved when the moving teacher and student differ on some tokens. For a comparable teacher--student gap, forward and reverse KLD produce a roughly $3.8\times$ larger and noticeably spikier objective, together with a similar increase in the gradient norm. \cref{app:jsd_kl} shows this over the course of training.

The distillation term moves the student toward the teacher's reading (see~\cref{fig:architecture}),
\begin{equation}
  d_{\mathrm{pos}}
  \;=\;
  D_{\mathrm{JSD}}\!\big(\,\operatorname{sg}[P^{(k_t)}] \,\big\|\, P^{(k_s)}\,\big),
  \label{eq:dpos}
\end{equation}
where $\operatorname{sg}[\cdot]$ denotes stop-gradient. At training step $i$,
both distributions are produced using the same current adapter $\phi_i$: the
teacher distribution from the $k_t=24$ frame view is detached, while gradients
flow only through the student distribution from the $k_s=12$ frame view. After
updating $\phi_i$ to $\phi_{i+1}$, this updated adapter is used by both passes in
the next step. Thus, the teacher evolves together with the student without a
separate teacher model or adapter.

The anchor keeps the adapted student within a
bounded distance of the base on identical input,
\begin{equation}
  \begin{aligned}
    d_{\mathrm{ref}}
    &\;=\;
    D_{\mathrm{JSD}}\!\big(\,\operatorname{sg}[P^{\mathrm{ref}}] \,\big\|\, P^{(k_s)}\,\big), \\
    P^{\mathrm{ref}} &= f_{\theta}(\,\cdot \mid x, k_s, q, a\,),
  \end{aligned}
  \label{eq:dref}
\end{equation}
where $P^{\mathrm{ref}}$ is the base distribution on the student's own sparse
view, obtained by disabling $\phi$ to recover $f_\theta$. Because
$d_{\mathrm{ref}}$ is a JSD to the base rather than a KLD, it inherits the same
boundedness. The full objective is
\begin{equation}
  \boxed{\;
  \mathcal{L}
  \;=\;
  d_{\mathrm{pos}} \;+\; \beta\, d_{\mathrm{ref}}
  \;}
  \label{eq:loss}
\end{equation}
with $d_{\mathrm{pos}}$ at unit weight and $\beta$ weighting the anchor.
A fixed $\beta$ is difficult to tune. If too small, the adapter moves too far from the base model and loses calibration. If too large, the student cannot learn enough from the teacher. Thus, we update $\beta$ to keep $d_{\mathrm{ref}}$ near a target $\kappa$. It increases when $d_{\mathrm{ref}}$ exceeds $\kappa$ and decreases when it falls below $\kappa$. The multiplicative update clips the per-step change and $\beta$, preventing spikes in $d_{\mathrm{ref}}$ from pushing $\beta$ to its bound and keeping it within a fixed range. \cref{app:data} gives the update rule and parameter values. \cref{alg:sst} summarizes one S$^3$T training step.

\newsavebox{\vstatbox}
\newlength{\vstatwd}
\newcommand{\dashsep}{%
  \multicolumn{9}{c}{\makebox[0pt]{%
    \tikz{\draw[dashed, line width=0.4pt] (0,0) -- (\vstatwd,0);}}}}
\newcommand{\vstattable}{%
\begin{tabular}{l c ccc cccc}
\toprule
 & & \multicolumn{3}{c}{State Element} & \multicolumn{4}{c}{State Structure}\\
\cmidrule(lr){3-5}\cmidrule(lr){6-9}
Method & Avg & Count & Loc. & Attr. & Atom. & Seq. & Set & Dict\\
\midrule
\textit{Human Performance}~\cite{vstat} & \textit{90.5} & \textit{92.8} & \textit{89.9} & \textit{86.4} & \textit{93.7} & \textit{77.5} & \textit{90.0} & \textit{92.4}\\
\midrule
\rowcolor{grouprow}\multicolumn{9}{l}{\textit{Open-source Models}}\\
LLaVA-OV-2-8B~\cite{llavaov2} & 34.83\,{\scriptsize$\pm$1.09} & 27.7\,{\scriptsize$\pm$1.3} & 42.7\,{\scriptsize$\pm$2.5} & 41.1\,{\scriptsize$\pm$2.3} & 32.7\,{\scriptsize$\pm$1.5} & 38.4\,{\scriptsize$\pm$3.3} & 47.3\,{\scriptsize$\pm$2.7} & 27.6\,{\scriptsize$\pm$2.3}\\ %
LLaVA-OV-2-8B (codec)~\cite{llavaov2} & 34.87\,{\scriptsize$\pm$1.09} & 30.3\,{\scriptsize$\pm$1.4} & 40.7\,{\scriptsize$\pm$2.4} & 38.1\,{\scriptsize$\pm$2.2} & 34.2\,{\scriptsize$\pm$1.5} & 35.5\,{\scriptsize$\pm$3.3} & 44.1\,{\scriptsize$\pm$2.6} & 29.0\,{\scriptsize$\pm$2.3}\\ %
Molmo2-4B~\cite{molmo2} & 34.79\,{\scriptsize$\pm$1.08} & 32.6\,{\scriptsize$\pm$1.4} & 40.0\,{\scriptsize$\pm$2.4} & 33.8\,{\scriptsize$\pm$2.2} & 38.1\,{\scriptsize$\pm$1.5} & 34.7\,{\scriptsize$\pm$3.3} & 35.1\,{\scriptsize$\pm$2.5} & 27.5\,{\scriptsize$\pm$2.2}\\ %
Cambrian-S-7B~\cite{action1} & 34.70\,{\scriptsize$\pm$1.07} & 34.2\,{\scriptsize$\pm$1.4} & 34.0\,{\scriptsize$\pm$2.4} & 36.4\,{\scriptsize$\pm$2.1} & 35.1\,{\scriptsize$\pm$1.5} & 31.0\,{\scriptsize$\pm$3.2} & 39.8\,{\scriptsize$\pm$2.4} & 32.3\,{\scriptsize$\pm$2.3}\\ %
Molmo2-8B~\cite{molmo2} & 36.04\,{\scriptsize$\pm$1.09} & 33.0\,{\scriptsize$\pm$1.4} & 39.9\,{\scriptsize$\pm$2.4} & 38.2\,{\scriptsize$\pm$2.2} & 38.0\,{\scriptsize$\pm$1.5} & 39.7\,{\scriptsize$\pm$3.4} & 37.5\,{\scriptsize$\pm$2.5} & 28.3\,{\scriptsize$\pm$2.3}\\ %
Qwen3-VL-8B~\cite{bai2025qwen3vltechnicalreport} & 34.11\,{\scriptsize$\pm$1.08} & 32.8\,{\scriptsize$\pm$1.4} & 38.4\,{\scriptsize$\pm$2.4} & 32.7\,{\scriptsize$\pm$2.1} & 35.5\,{\scriptsize$\pm$1.5} & 33.5\,{\scriptsize$\pm$3.2} & 36.5\,{\scriptsize$\pm$2.5} & 29.6\,{\scriptsize$\pm$2.3}\\ %
InternVL3.5-2B~\cite{wang2025internvl3} & 32.36\,{\scriptsize$\pm$1.05} & 30.1\,{\scriptsize$\pm$1.4} & 33.4\,{\scriptsize$\pm$2.3} & 35.8\,{\scriptsize$\pm$2.2} & 30.6\,{\scriptsize$\pm$1.4} & 31.0\,{\scriptsize$\pm$3.2} & 37.5\,{\scriptsize$\pm$2.5} & 33.1\,{\scriptsize$\pm$2.4}\\ %
Cambrian-S-3B~\cite{action1} & 31.56\,{\scriptsize$\pm$1.04} & 30.3\,{\scriptsize$\pm$1.4} & 29.6\,{\scriptsize$\pm$2.3} & 36.3\,{\scriptsize$\pm$2.1} & 32.2\,{\scriptsize$\pm$1.5} & 31.5\,{\scriptsize$\pm$3.2} & 35.9\,{\scriptsize$\pm$2.3} & 27.0\,{\scriptsize$\pm$2.2}\\ %
OmniVideo-7B~\cite{omnivideo100k} & 33.97\,{\scriptsize$\pm$1.08} & 28.6\,{\scriptsize$\pm$1.4} & 37.7\,{\scriptsize$\pm$2.4} & 41.0\,{\scriptsize$\pm$2.2} & 34.1\,{\scriptsize$\pm$1.6} & 31.9\,{\scriptsize$\pm$3.2} & 45.5\,{\scriptsize$\pm$2.6} & 26.4\,{\scriptsize$\pm$2.2}\\ %
Qwen3-VL-4B~\cite{bai2025qwen3vltechnicalreport} & 31.44\,{\scriptsize$\pm$1.06} & 27.7\,{\scriptsize$\pm$1.4} & 33.7\,{\scriptsize$\pm$2.3} & 36.6\,{\scriptsize$\pm$2.2} & 31.9\,{\scriptsize$\pm$1.5} & 32.5\,{\scriptsize$\pm$3.2} & 38.5\,{\scriptsize$\pm$2.6} & 24.4\,{\scriptsize$\pm$2.2}\\ %
InternVL3.5-8B~\cite{wang2025internvl3} & 29.81\,{\scriptsize$\pm$1.05} & 22.7\,{\scriptsize$\pm$1.3} & 37.8\,{\scriptsize$\pm$2.4} & 35.9\,{\scriptsize$\pm$2.2} & 26.8\,{\scriptsize$\pm$1.4} & 37.1\,{\scriptsize$\pm$3.3} & 37.7\,{\scriptsize$\pm$2.5} & 25.8\,{\scriptsize$\pm$2.2}\\ %
Qwen3-VL-2B~\cite{bai2025qwen3vltechnicalreport} & 30.44\,{\scriptsize$\pm$1.04} & 30.7\,{\scriptsize$\pm$1.4} & 29.2\,{\scriptsize$\pm$2.3} & 31.3\,{\scriptsize$\pm$2.1} & 33.8\,{\scriptsize$\pm$1.5} & 20.9\,{\scriptsize$\pm$2.8} & 36.4\,{\scriptsize$\pm$2.5} & 24.7\,{\scriptsize$\pm$2.2}\\ %
Cambrian-S-1.5B~\cite{action1} & 29.23\,{\scriptsize$\pm$1.05} & 26.0\,{\scriptsize$\pm$1.4} & 33.6\,{\scriptsize$\pm$2.4} & 31.1\,{\scriptsize$\pm$2.0} & 28.0\,{\scriptsize$\pm$1.5} & 31.5\,{\scriptsize$\pm$3.2} & 30.8\,{\scriptsize$\pm$2.2} & 29.2\,{\scriptsize$\pm$2.3}\\ %
LLaVA-OV-7B~\cite{llavaov} & 28.43\,{\scriptsize$\pm$1.06} & 19.5\,{\scriptsize$\pm$1.2} & 36.5\,{\scriptsize$\pm$2.4} & 38.1\,{\scriptsize$\pm$2.3} & 24.6\,{\scriptsize$\pm$1.4} & 31.8\,{\scriptsize$\pm$3.2} & 39.7\,{\scriptsize$\pm$2.6} & 25.9\,{\scriptsize$\pm$2.3}\\ %
LLaVA-OV-0.5B~\cite{llavaov} & 19.13\,{\scriptsize$\pm$0.97} & 12.3\,{\scriptsize$\pm$1.1} & 31.6\,{\scriptsize$\pm$2.3} & 19.9\,{\scriptsize$\pm$2.0} & 15.4\,{\scriptsize$\pm$1.3} & 27.7\,{\scriptsize$\pm$3.1} & 17.2\,{\scriptsize$\pm$2.2} & 23.1\,{\scriptsize$\pm$2.2}\\ %
\midrule
\rowcolor{grouprow}\multicolumn{9}{l}{\textit{Self-Evolving Methods}}\\
\textit{Qwen3-VL-4B~\cite{bai2025qwen3vltechnicalreport}} & \textit{31.44}\,{\scriptsize$\pm$1.06} & \textit{27.7}\,{\scriptsize$\pm$1.4} & \textit{33.7}\,{\scriptsize$\pm$2.3} & \textit{36.6}\,{\scriptsize$\pm$2.2} & \textit{31.9}\,{\scriptsize$\pm$1.5} & \textit{32.5}\,{\scriptsize$\pm$3.2} & \textit{38.5}\,{\scriptsize$\pm$2.6} & \textit{24.4}\,{\scriptsize$\pm$2.2}\\ %
Video-Zero-4B~\cite{evo3} & 31.62\,{\scriptsize$\pm$1.06} & 27.7\,{\scriptsize$\pm$1.4} & 34.5\,{\scriptsize$\pm$2.4} & 36.5\,{\scriptsize$\pm$2.2} & 32.1\,{\scriptsize$\pm$1.5} & 32.3\,{\scriptsize$\pm$3.2} & 37.8\,{\scriptsize$\pm$2.5} & 25.5\,{\scriptsize$\pm$2.2}\\ %
\noalign{\vskip -4pt}
\dashsep\\
\textit{Qwen3-VL-8B~\cite{bai2025qwen3vltechnicalreport}} & \textit{34.11}\,{\scriptsize$\pm$1.08} & \textit{32.8}\,{\scriptsize$\pm$1.4} & \textit{38.4}\,{\scriptsize$\pm$2.4} & \textit{32.7}\,{\scriptsize$\pm$2.1} & \textit{35.5}\,{\scriptsize$\pm$1.5} & \textit{33.5}\,{\scriptsize$\pm$3.2} & \textit{36.5}\,{\scriptsize$\pm$2.5} & \textit{29.6}\,{\scriptsize$\pm$2.3}\\ %
Video-Zero-8B~\cite{evo3} & 34.17\,{\scriptsize$\pm$1.07} & 32.1\,{\scriptsize$\pm$1.4} & 39.1\,{\scriptsize$\pm$2.4} & 33.3\,{\scriptsize$\pm$2.1} & 35.3\,{\scriptsize$\pm$1.5} & 34.5\,{\scriptsize$\pm$3.2} & 38.1\,{\scriptsize$\pm$2.5} & 28.5\,{\scriptsize$\pm$2.3}\\ %
\noalign{\vskip -4pt}
\dashsep\\
\textit{Qwen2.5-VL-7B~\cite{bai2025qwen25vltechnicalreport}} & \textit{30.77}\,{\scriptsize$\pm$1.07} & \textit{25.2}\,{\scriptsize$\pm$1.4} & \textit{36.9}\,{\scriptsize$\pm$2.4} & \textit{35.8}\,{\scriptsize$\pm$2.2} & \textit{29.6}\,{\scriptsize$\pm$1.5} & \textit{39.5}\,{\scriptsize$\pm$3.3} & \textit{33.4}\,{\scriptsize$\pm$2.4} & \textit{25.6}\,{\scriptsize$\pm$2.2}\\ %
EvoGround~\cite{evo2} & 31.11\,{\scriptsize$\pm$1.06} & 26.3\,{\scriptsize$\pm$1.4} & 34.8\,{\scriptsize$\pm$2.4} & 37.0\,{\scriptsize$\pm$2.2} & 30.1\,{\scriptsize$\pm$1.5} & 39.1\,{\scriptsize$\pm$3.3} & 35.4\,{\scriptsize$\pm$2.4} & 24.9\,{\scriptsize$\pm$2.2}\\ %
\noalign{\vskip -4pt}
\dashsep\\
\textit{InternVL2.5-8B~\cite{chen2024internvl25}} & \textit{31.47}\,{\scriptsize$\pm$1.07} & \textit{27.7}\,{\scriptsize$\pm$1.4} & \textit{34.3}\,{\scriptsize$\pm$2.4} & \textit{36.2}\,{\scriptsize$\pm$2.2} & \textit{29.1}\,{\scriptsize$\pm$1.5} & \textit{33.2}\,{\scriptsize$\pm$3.2} & \textit{39.0}\,{\scriptsize$\pm$2.6} & \textit{29.8}\,{\scriptsize$\pm$2.4}\\ %
VideoSAVi~\cite{kulkarni2025videosaviselfalignedvideolanguage} & 32.73\,{\scriptsize$\pm$1.07} & 28.7\,{\scriptsize$\pm$1.4} & 34.7\,{\scriptsize$\pm$2.4} & 38.8\,{\scriptsize$\pm$2.2} & 30.9\,{\scriptsize$\pm$1.5} & 33.4\,{\scriptsize$\pm$3.2} & 41.6\,{\scriptsize$\pm$2.6} & 29.7\,{\scriptsize$\pm$2.3}\\ %
\midrule
\rowcolor{grouprow}\multicolumn{9}{l}{\textit{Ours}}\\
\textit{LLaVA-OV-2-8B} & \textit{34.83}\,{\scriptsize$\pm$1.09} & \textit{27.7}\,{\scriptsize$\pm$1.3} & \textit{42.7}\,{\scriptsize$\pm$2.5} & \textit{41.1}\,{\scriptsize$\pm$2.3} & \textit{32.7}\,{\scriptsize$\pm$1.5} & \textit{38.4}\,{\scriptsize$\pm$3.3} & \textit{47.3}\,{\scriptsize$\pm$2.7} & \textit{27.6}\,{\scriptsize$\pm$2.3}\\
\rowcolor{ourrow}S$^3$T & 37.44\,{\scriptsize$\pm$1.08} & 32.6\,{\scriptsize$\pm$1.4} & 42.3\,{\scriptsize$\pm$2.4} & 42.1\,{\scriptsize$\pm$2.2} & 37.7\,{\scriptsize$\pm$1.5} & 42.8\,{\scriptsize$\pm$3.3} & 45.5\,{\scriptsize$\pm$2.6} & 27.4\,{\scriptsize$\pm$2.2}\\
\noalign{\vskip -4pt}
\dashsep\\
\textit{Qwen3-VL-8B} & \textit{34.11}\,{\scriptsize$\pm$1.08} & \textit{32.8}\,{\scriptsize$\pm$1.4} & \textit{38.4}\,{\scriptsize$\pm$2.4} & \textit{32.7}\,{\scriptsize$\pm$2.1} & \textit{35.5}\,{\scriptsize$\pm$1.5} & \textit{33.5}\,{\scriptsize$\pm$3.2} & \textit{36.5}\,{\scriptsize$\pm$2.5} & \textit{29.6}\,{\scriptsize$\pm$2.3}\\
\rowcolor{ourrow}S$^3$T & 36.55\,{\scriptsize$\pm$1.08} & 35.5\,{\scriptsize$\pm$1.4} & 38.4\,{\scriptsize$\pm$2.4} & 36.7\,{\scriptsize$\pm$2.2} & 37.2\,{\scriptsize$\pm$1.5} & 35.9\,{\scriptsize$\pm$3.2} & 40.5\,{\scriptsize$\pm$2.5} & 32.5\,{\scriptsize$\pm$2.3}\\
\noalign{\vskip -4pt}
\dashsep\\
\textit{InternVL3.5-8B} & \textit{29.81}\,{\scriptsize$\pm$1.05} & \textit{22.7}\,{\scriptsize$\pm$1.3} & \textit{37.8}\,{\scriptsize$\pm$2.4} & \textit{35.9}\,{\scriptsize$\pm$2.2} & \textit{26.8}\,{\scriptsize$\pm$1.4} & \textit{37.1}\,{\scriptsize$\pm$3.3} & \textit{37.7}\,{\scriptsize$\pm$2.5} & \textit{25.8}\,{\scriptsize$\pm$2.2}\\
\rowcolor{ourrow}S$^3$T & 30.74\,{\scriptsize$\pm$1.06} & 23.9\,{\scriptsize$\pm$1.3} & 39.2\,{\scriptsize$\pm$2.4} & 35.9\,{\scriptsize$\pm$2.2} & 27.9\,{\scriptsize$\pm$1.4} & 35.5\,{\scriptsize$\pm$3.2} & 38.8\,{\scriptsize$\pm$2.5} & 27.6\,{\scriptsize$\pm$2.3}\\
\bottomrule
\end{tabular}}

\newsavebox{\vstatleft}\newsavebox{\vstatradar}\newsavebox{\vstatsoup}\newsavebox{\vstatpre}
\newlength{\vstatleftwd}\newlength{\vstatrightwd}\newlength{\vstatH}
\newlength{\vstatfigH}\newlength{\vstatfigW}

\newsavebox{\vstatpretab}\newlength{\vstatprecapwd}

\begin{table*}[!t]\centering
\caption{\textbf{S$^3$T on VSTAT.} \textbf{(a)} VSTAT accuracy (\%) under the official protocol~\cite{vstat}, reported as accuracy $\pm$ standard error over evaluation questions. Each self-evolving method is listed below its respective base model, and S$^3$T rows are shaded. \textbf{(b)} Per-axis scores of each base model, its S$^3$T-trained counterpart, and competitive open models. \textbf{(c)} LLaVA-OV-2-8B trained with the same S$^3$T recipe and data, varying only the training objective. \textbf{(d)} VSTAT accuracy grouped by video duration. S$^3$T is trained only on $20$-second clips, making the last two groups longer than any training clip. $\Delta$ denotes the change relative to each base model.}
\setlength{\tabcolsep}{3pt}%
\renewcommand{\arraystretch}{1.14}%
\setlength{\vstatleftwd}{0.57\textwidth}%
\setlength{\vstatrightwd}{\dimexpr\textwidth-\vstatleftwd-0.015\textwidth\relax}%
\setlength{\vstatwd}{0pt}%
\sbox{\vstatbox}{\vstattable}%
\setlength{\vstatwd}{\wd\vstatbox}%
\sbox{\vstatradar}{%
\begin{tikzpicture}[x=1pt,y=1pt,line join=round,line cap=round]
\definecolor{radbase}{HTML}{1F6FB2}
\definecolor{radours}{HTML}{2E7D32}
\definecolor{radvz}{HTML}{882255}    %
\definecolor{radcam}{HTML}{E69F00}   %
\definecolor{radivl}{HTML}{56B4E9}   %
\definecolor{radov}{HTML}{CC79A7}    %
\definecolor{radbg}{HTML}{F7F2F7}
\providecommand{\radR}{38pt}
\providecommand{\radr}[1]{\radR*((#1)-\radlo)/(\radhi-\radlo)}
\providecommand{\radarpoly}[2]{%
  \edef\radlist{#2}%
  \foreach \v [count=\i from 0] in \radlist{\coordinate (rp\i) at ({90-\i*360/7}:{\radr{\v}});}
  \draw[#1] (rp0) -- (rp1) -- (rp2) -- (rp3) -- (rp4) -- (rp5) -- (rp6) -- cycle;}
\providecommand{\radvz}{32.1,39.1,33.3,35.3,34.5,38.1,28.5}
\providecommand{\radcam}{34.2,34.0,36.4,35.1,31.0,39.8,32.3}
\providecommand{\radivl}{22.7,37.8,35.9,26.8,37.1,37.7,25.8}
\providecommand{\radov}{19.5,36.5,38.1,24.6,31.8,39.7,25.9}
\tikzset{radref/.style={line width=0.6pt}}
\providecommand{\backboneradar}[7]{%
  \begin{scope}[shift={#1}]
    \def\radlo{#3}\def\radhi{#4}%
    \fill[radbg] (90:\radR) \foreach \i in {1,...,6}{ -- ({90-\i*360/7}:\radR)} -- cycle;
    \foreach \s in {#5}{%
      \draw[white,line width=0.6pt] (90:{\radr{\s}}) \foreach \i in {1,...,6}{ -- ({90-\i*360/7}:{\radr{\s}})} -- cycle;
      \node[anchor=west,text=black!45,font=\tiny,inner sep=1pt] at ($(90:{\radr{\s}})+(1pt,-2pt)$) {\s};}
    \foreach \i in {0,...,6}{\draw[white,line width=0.6pt] (0,0) -- ({90-\i*360/7}:\radR);}
    \radarpoly{radov,radref}{\radov}
    \radarpoly{radivl,radref}{\radivl}
    \radarpoly{radcam,radref}{\radcam}
    \radarpoly{radvz,radref}{\radvz}
    \radarpoly{radbase,line width=0.9pt,densely dashed,fill=radbase,fill opacity=0.06}{#6}
    \radarpoly{radours,line width=1.3pt,fill=radours,fill opacity=0.10}{#7}
    \foreach \lab [count=\i from 0] in {Count,Loc.,Attr.,Atom.,Seq.,Set,Dict}{%
      \node[font=\scriptsize,inner sep=1pt] at ({90-\i*360/7}:{\radR+8pt}) {\lab};}
    \node[font=\footnotesize,anchor=south,inner sep=1pt] at (0,{\radR+15pt}) {#2};
  \end{scope}}
\providecommand{\radkey}[4]{%
  \draw[#2] ($(#1)+(5pt,0)$) -- ++(8pt,0) coordinate (radk);
  \node[anchor=west,font=\tiny,inner sep=1pt] (#4) at (radk) {#3};}
  \backboneradar{(0,0)}{LLaVA-OV-2-8B}{15}{50}{20,30,40}%
    {27.7,42.7,41.1,32.7,38.4,47.3,27.6}{32.6,42.3,42.1,37.7,42.8,45.5,27.4}
  \backboneradar{(128,0)}{Qwen3-VL-8B}{15}{45}{20,30,40}%
    {32.8,38.4,32.7,35.5,33.5,36.5,29.6}{35.5,38.4,36.7,37.2,35.9,40.5,32.5}
  \coordinate (k0) at (-62,-52);
  \path (0,-60);   %
  \radkey{k0}{radbase,line width=0.9pt,densely dashed}{Base}{k1}
  \radkey{k1.east}{radours,line width=1.3pt}{S$^3$T}{k2}
  \radkey{k2.east}{radvz,radref}{Video-Zero-8B}{k3}
  \radkey{k3.east}{radcam,radref}{Cambrian-S-7B}{k4}
  \radkey{k4.east}{radivl,radref}{InternVL3.5-8B}{k5}
  \radkey{k5.east}{radov,radref}{LLaVA-OV-7B}{k6}
\end{tikzpicture}
}%
\sbox{\vstatsoup}{\resizebox{\vstatrightwd}{!}{%
{\scriptsize
\setlength{\tabcolsep}{2pt}\renewcommand{\arraystretch}{1.4}%
\begin{tabular}{l c ccc ccc}
\toprule
 & & \multicolumn{3}{c}{LLaVA-OV-2-8B} & \multicolumn{3}{c}{Qwen3-VL-8B}\\
\cmidrule(lr){3-5}\cmidrule(lr){6-8}
Length & $n$ & Base & S$^3$T & $\Delta$ & Base & S$^3$T & $\Delta$\\
\midrule
$0$--$20$\,s   & $367$ & $42.46$ & $42.61$ & $+0.15$ & $38.34$ & $38.72$ & $+0.38$\\
$20$--$60$\,s  & $919$ & $32.69$ & $36.24$ & \rk{$+3.55$} & $32.63$ & $35.53$ & \rs{$+2.90$}\\
$60$--$100$\,s & $214$ & $30.93$ & $33.72$ & \rs{$+2.79$} & $33.18$ & $37.20$ & \rk{$+4.02$}\\
\bottomrule
\end{tabular}}%
}}%
\sbox{\vstatpretab}{%
{\footnotesize\setlength{\tabcolsep}{2.5pt}\renewcommand{\arraystretch}{1.1}%
\begin{tabular}{l c c c}
\toprule
Objective & Avg & Count & Dict\\
\midrule
Base model & 34.83 & 27.7 & 27.6\\
\midrule
Frame-order~\cite{frameorder,frame2} & 34.98 & 27.7 & \rk{30.5}\\
Arrow-of-time~\cite{arrow,arrow2} & 34.75 & 27.4 & 26.5\\
Pace~\cite{pace,pace2} & \rs{35.45} & \rs{29.0} & 27.1\\
Mask-infill~\cite{videomae} & 34.25 & 25.2 & 26.3\\
\midrule
S$^3$T (ours) & \rk{37.44} & \rk{32.6} & \rs{27.4}\\
\bottomrule
\end{tabular}}%
}%
\setlength{\vstatprecapwd}{\dimexpr\vstatrightwd-\wd\vstatpretab-6pt\relax}%
\sbox{\vstatleft}{\begin{minipage}[t]{\vstatleftwd}\centering
\vspace{0pt}                          %
\resizebox{\linewidth}{!}{\vstattable}
\subcaption{VSTAT leaderboard}\label{tab:main_results}
\end{minipage}}%
\setlength{\vstatH}{\dimexpr\ht\vstatleft+\dp\vstatleft\relax}%
\setlength{\vstatfigH}{\dimexpr\vstatH-\ht\vstatsoup-\dp\vstatsoup-\ht\vstatpretab-\dp\vstatpretab-44pt\relax}%
\edef\vstatscale{\fpeval{min(\vstatfigH/(\ht\vstatradar+\dp\vstatradar), \vstatrightwd/\wd\vstatradar)}}%
\usebox{\vstatleft}\hfill
\begin{subfigure}[t][\vstatH][t]{\vstatrightwd}\centering
\vspace{0pt}
\scalebox{\vstatscale}{\usebox{\vstatradar}}
{\captionsetup{justification=centering,font=small,skip=2pt}\caption{Per-axis profile, base vs.\ S$^3$T}\label{fig:radar}}
\vfill
\begin{minipage}[c]{\wd\vstatpretab}\usebox{\vstatpretab}\end{minipage}\hfill
\begin{minipage}[c]{\vstatprecapwd}
{\captionsetup{justification=raggedright,singlelinecheck=false,font=small,skip=0pt}\caption{Comparison with established temporal self-supervised pretexts}\label{tab:pretext}}
\end{minipage}
\vfill
\usebox{\vstatsoup}
{\captionsetup{justification=centering,font=small,skip=2pt}\caption{VSTAT accuracy by video length}\label{tab:horizon}}
\end{subfigure}
\end{table*}

\begin{algorithm}[t]
\caption{One S$^3$T training step}
\label{alg:sst}
\small
\begin{algorithmic}[1]
\Require frozen base $f_\theta$, single adapter $\phi_i$, probe $q$, frame counts
$k_s{=}12$, $k_t{=}24$, anchor set point $\kappa$, rate $\eta$, bounds
$[\beta_{\min}, \beta_{\max}]$, anchor weight $\beta$
\State sample clip $x$; decode to $T$ frames
\State $V_s \gets k_s$ uniform frames of $x$ \Comment{student view}
\State $V_t \gets k_t$ uniform frames of $x$ \Comment{same span, denser}
\State $a \gets$ greedy decode $f_{\theta+\phi_i}(\cdot \mid V_s, q)$ \Comment{self-generated target}
\State $P^{(k_t)} \gets f_{\theta+\phi_i}(\cdot \mid V_t, q, a)$, no grad \Comment{same adapter; teacher}
\State $P^{\mathrm{ref}} \gets f_{\theta}(\cdot \mid V_s, q, a)$, no grad \Comment{adapter off}
\State $P^{(k_s)} \gets f_{\theta+\phi_i}(\cdot \mid V_s, q, a)$ \Comment{same adapter; student}
\State $d_{\mathrm{pos}} \gets D_{\mathrm{JSD}}\big(\operatorname{sg}[P^{(k_t)}] \,\|\, P^{(k_s)}\big)$
\State $d_{\mathrm{ref}} \gets D_{\mathrm{JSD}}\big(\operatorname{sg}[P^{\mathrm{ref}}] \,\|\, P^{(k_s)}\big)$
\State $\mathcal{L} \gets d_{\mathrm{pos}} + \beta\, d_{\mathrm{ref}}$
\State update $\phi_i \rightarrow \phi_{i+1}$ by back-propagating $\mathcal{L}$ through the student pass only; clip; step
\State $\delta \gets (d_{\mathrm{ref}} - \kappa)/\kappa$
\State $\beta \gets \operatorname{clip}\big(\beta\, e^{\,\eta\, \operatorname{clip}(\delta,\, -3,\, 3)},\; \beta_{\min},\, \beta_{\max}\big)$ \Comment{anchor}
\end{algorithmic}
\end{algorithm}

\section{Experiments}
\label{sec:experiments}

\paragraph{Implementation details.}
We fine-tune LLaVA-OV-2-8B~\cite{llavaov2} using rank-$32$ LoRA~\cite{hu2022lora} on the attention and feed-forward projections of both the vision encoder and language model. The base weights, patch embedding, and multimodal projector remain frozen, with only the adapter updated. Both views use the same encoder and adapter, differing only in frame count ($12$ for the student and $24$ for the teacher). We train for $3000$ steps with one clip per step, using AdamW~\cite{adamw} in bfloat16. Training takes about $2.5$ hours on a single A100-80GB GPU. For Probe Merging, we scale the averaged weight update by $\alpha{=}1.3$, selected on held-out \emph{StateGen} clips (\cref{app:heldout}). Training uses a fixed pool of $300$ unlabeled \emph{StateGen} clips (\cref{sec:method:data}), with only RGB frames provided to the model, without state annotations or generator metadata. Remaining hyperparameters, protocol selection, and generation and sampling details are in \cref{app:data}.

\paragraph{Baselines and evaluation.}
We evaluate on VSTAT~\cite{vstat} using lmms-eval~\cite{zhang2024lmmsevalrealitycheckevaluation}, which contains $1{,}500$ multiple-choice and numerical questions across $834$ simulated and real-world videos. These questions require temporal evidence and cannot be answered from a single keyframe or the final frame alone. The official metric averages multiple-choice accuracy and mean relative accuracy on numerical questions. Each question has a state-element label (Count, Location, or Attribute) and a state-structure label (Atomic, Sequence, Set, or Dictionary). We compare against open-source Video-LLMs, self-evolving methods, and temporal self-supervised objectives. All models follow frame budgets and evaluation settings reported in VSTAT~\cite{vstat}, with each S$^3$T model using the same settings as its base model. The gain grows with the number of evaluation frames, suggesting that S$^3$T helps the model combine evidence across frames (\cref{app:framebudget}).

\subsection{Main results}
\label{sec:exp:main}
Visual state tracking improves without external supervision, with overall VSTAT accuracy increasing from $34.83$ to $37.44$ ($+2.61$). The two probes capture different aspects of the state, and their merged model outperforms either probe alone on Count, Atomic, and Sequence, which require accumulating evidence across the clip. We ablate each component of S$^3$T, including probe merging, in \cref{sec:exp:ablation}. We also compare with prior self-evolving methods, which similarly learn from model-generated supervision but use different training objectives and data. We therefore compare each method against its own base model, while our controlled SFT and temporal-pretext experiments (\cref{tab:ablation_table,tab:pretext}) isolate the effect of the training objective under the same data and model setup. These methods improve accuracy by at most $+1.26$, compared with $+2.61$ for S$^3$T. The gains extend across backbones: on Qwen3-VL-8B, S$^3$T improves accuracy from $34.11$ to $36.55$ ($+2.44$), compared to just $+0.06$ for Video-Zero on the same base model. On InternVL3.5-8B, accuracy improves from $29.81$ to $30.74$ ($+0.93$). Further results are in \cref{app:morebackbones}. These gains are also consistent across five independent runs, with S$^3$T achieving $37.36\pm0.35$ on LLaVA-OV-2-8B, $36.47\pm0.42$ on Qwen3-VL-8B, and $30.69\pm0.23$ on InternVL3.5-8B. The gain is relatively smaller on InternVL3.5-8B, which has the lowest base accuracy of the three backbones and likely provides a less reliable teacher for the student to learn from. Every run improves over its respective base model (\cref{tab:seed-reliability}).

\begin{table}[t]\centering
\footnotesize
\caption{\textbf{Component ablations of S$^3$T.}
We report VSTAT accuracy (\%) overall and on the Count and Dictionary axes. Rows marked (ours) indicate the settings used in our method. Blocks are separate comparisons and are not ranked against each other. Definitions of every setting, more teacher-view alternatives, frame-ratio ablations, and the merging strength are in \cref{app:ablation-defs,app:framebudget,app:souping}.}
\label{tab:ablation_table}
\setlength{\tabcolsep}{4pt}%
\begin{tabular}{l c c c}
\toprule
Setting (on LLaVA-OV-2-8B) & Overall & Count & Dict\\
\midrule
\rowcolor{grouprow}\multicolumn{4}{l}{\textit{Teacher view}}\\
Dense full-clip view $12\!\to\!24$ (ours) & \rk{37.44} & \rk{32.6} & \rs{27.4}\\
Temporal zoom (windowed) & 33.91 & 26.1 & 25.2\\
Identical $12\!\to\!12$ (no privileged view) & 34.83 & 27.7 & \rk{27.6}\\
Shifted equal-density teacher ($12\!\to\!12$) & \rs{35.14} & \rs{28.2} & 27.2\\
\midrule
\rowcolor{grouprow}\multicolumn{4}{l}{\textit{Objective and anchor}}\\
JSD distillation with reference anchor (ours) & \rk{37.44} & \rk{32.6} & 27.4\\
SFT (cross-entropy on teacher text) & \rs{34.45} & 26.0 & 27.5\\
CE on generator's true state & 34.05 & 25.8 & 26.7\\
JSD to generator's true state & 33.90 & \rs{27.5} & \rk{29.2}\\
No anchor ($\beta{=}0$) & 33.68 & 27.3 & \rs{27.7}\\
\midrule
\rowcolor{grouprow}\multicolumn{4}{l}{\textit{Probes and merging}}\\
S$^3$T, $q_1$ and $q_2$ merged (ours) & \rk{37.44} & \rk{32.6} & 27.4\\
No merging, $q_1$ only & \rs{36.88} & \rs{31.9} & 26.8\\
No merging, $q_2$ only & 36.11 & 31.3 & 26.2\\
Five $q_1$ seeds merged & 36.08 & 29.7 & \rs{27.8}\\
Five $q_2$ seeds merged & 36.14 & 30.3 & 27.1\\
$q_1$ alternated with entity-ID probe & 35.42 & 28.3 & 26.1\\
Colour-set probe (no numbers) & 33.32 & 26.4 & 26.7\\
Change questions only (no $q_1$) & 29.35 & 23.3 & \rk{29.4}\\
\midrule
\rowcolor{grouprow}\multicolumn{4}{l}{\textit{Adapted modules}}\\
Vision encoder and language model (ours) & \rk{37.44} & \rk{32.6} & \rs{27.4}\\
Language model only (frozen vision encoder) & \rs{37.12} & \rs{32.4} & \rk{28.4}\\
\bottomrule
\end{tabular}
\vspace{-10pt}
\end{table}
 
The per-axis breakdown shows that improvements are concentrated on the cumulative-state axes, where answering requires integrating evidence across the entire clip. With S$^3$T, LLaVA-OV-2-8B achieves the highest Sequence score and Qwen3-VL-8B the highest Count score among the models compared. The radar plots likewise show improvements in Count, Atomic, and Sequence across both backbones (\cref{fig:radar}). Supervised finetuning (SFT) helps isolate the effect of the distillation objective. It uses the same dense teacher view but fine-tunes on the teacher's generated text rather than its predictive distribution. SFT remains below the base model overall (\cref{tab:ablation_table}), suggesting that matching the teacher's predictive distribution, rather than its generated text, is important to S$^3$T's improvements (see \cref{sec:exp:ablation}). Finally, we examine whether these improvements could instead be explained by answer bias, where a model favors the most common answer for each task without using the video itself. Our analysis in \cref{app:prior} shows that answer bias cannot fully explain the improvement, as S$^3$T also improves on questions where the most common answer is incorrect.

\subsection{Long-horizon tracking}
\label{sec:exp:horizon}
S$^3$T is trained only on $20$-second clips, while VSTAT videos last up to $100$ seconds. To evaluate whether its improvements extend beyond the training duration, we group all $1{,}500$ VSTAT questions by video length and compare S$^3$T with its base model within each group (\cref{tab:horizon}). If S$^3$T merely adapted to the duration of its training clips, its improvements would be largest on videos around $20$ seconds and diminish as video duration increases. Instead, we observe the opposite trend, with improvements of less than half a point on videos up to $20$ seconds, rising to $+3.55$ and $+2.79$ for LLaVA-OV-2-8B, and $+2.90$ and $+4.02$ for Qwen3-VL-8B on longer videos. Both backbones show this pattern, and their base models also achieve lower accuracy on longer videos, indicating that these videos pose greater challenges for state tracking. These results show that $20$-second training clips are sufficient to learn state tracking that generalizes to videos up to five times longer, without the additional frames per step and compute required by longer training clips (\cref{sec:discussion}).

\subsection{Comparison with temporal pretexts}
\label{sec:exp:pretext}

\cref{tab:pretext} compares S$^3$T with four established temporal self-supervised objectives. \textit{Frame-order prediction}~\cite{frameorder,frame2} recovers the order of shuffled clip segments, \textit{arrow-of-time}~\cite{arrow,arrow2} predicts playback direction, and \textit{pace prediction}~\cite{pace,pace2} identifies playback speed. \textit{Masked temporal infill}~\cite{videomae} predicts the content of masked temporal spans. All four use the same base model (LLaVA-OV-2-8B~\cite{llavaov2}) and S$^3$T configuration, with only the learning objective changed. Full pretext definitions are given in \cref{app:pretext-detail}.
None of these objectives reproduces the improvement from S$^3$T. Pace prediction performs best among the four, improving overall accuracy from $34.83$ to $35.45$ ($+0.62$), but this is less than a quarter of S$^3$T's $+2.61$ improvement. Its $+1.3$ improvement on Count is also well below S$^3$T's $+4.9$. Arrow-of-time and masked temporal infill both fall below the base model overall. Frame-order prediction yields the best Dictionary score ($30.5$), but leaves Count unchanged and improves overall accuracy by just $0.15$. These show that S$^3$T's gains stem not from temporal self-supervision alone, but from distilling privileged temporal density.

\subsection{Ablations}
\label{sec:exp:ablation}

\paragraph{Teacher view.}
The teacher-view ablations in \cref{tab:ablation_table} show that improvements depend on denser sampling across the full clip. The windowed teacher performs worse than the base model ($33.91$ vs.\ $34.83$), showing that denser sampling over a shorter window is not enough. Using identical $12\!\to\!12$ views removes the privileged information and gives exactly the base accuracy of $34.83$. Shifting the teacher's $12$ frames to different timestamps also falls short, at $35.14$. Only the dense full-clip $12\!\to\!24$ view raises accuracy to $37.44$ ($+2.61$). Thus, more frames or full-clip coverage alone are insufficient. Additional teacher variants and analysis of teacher--student disagreement are provided in \cref{app:ablation-defs,app:dpos}.

\paragraph{Objective and anchor.}
The objective ablations show that the teacher's predictive distribution is important to S$^3$T's improvement. Fine-tuning on the teacher's generated text gives $34.45$, while direct cross-entropy (CE) training on the generator's true state and JSD toward that state give $34.05$ and $33.90$, respectively. Neither form of supervision reproduces the improvement from matching the teacher's predictive distribution ($37.44$), even when the target is the correct state. Removing the reference anchor also reduces accuracy to $33.68$, below the base model's $34.83$, showing its importance for stable distillation. Additional training-step comparisons and results for forward and reverse KL divergence are provided in \cref{app:ablation-defs,app:jsd_kl}.

\paragraph{Probes and merging.}
The two probes capture complementary aspects of state tracking. Training with $q_1$ and $q_2$ separately gives $36.88$ and $36.11$ overall, while merging their adapters raises accuracy to $37.44$. This is not simply due to weight averaging, as merging five seeds of $q_1$ or $q_2$ gives only $36.08$ and $36.14$, respectively. Probe choice also matters. A colour-set probe that forbids numbers lowers Count accuracy to $26.4$ and overall accuracy to $33.32$, below the base model. Using only the short change questions of $q_2$ further lowers overall accuracy to $29.35$, while alternating $q_1$ with a probe assigning each ball an ID and location gives $35.42$. These results match the two parts of state tracking; without counts, the state probe no longer captures the full current state, and the change questions alone cannot recover it. Both parts are needed, and the probes work best together. S$^3$T combines both through adapter merging, and these exact probes and merging details are in \cref{app:souping}.

\begin{table}[t]
\centering
\footnotesize
\setlength{\tabcolsep}{4pt}
\caption{\textbf{Transfer to real-video benchmarks.} The VSTAT-YouTube result is computed on a predefined cumulative-state subgroup of the main VSTAT evaluation. $^{\mathrm{ns}}$ denotes a non-significant change. Gains occur on tasks requiring a running tally, while broader video-understanding performance is largely preserved. Confidence intervals are reported in \cref{tab:realvideo-ci}.}
\label{tab:transfer_table}
\begin{tabular*}{\columnwidth}{@{\extracolsep{\fill}}lcc@{}}
\toprule
Real-video benchmark
& \begin{tabular}[c]{@{}c@{}}Base~\cite{llavaov2}\end{tabular}
& S$^3$T\\
\midrule
\rowcolor{grouprow}\multicolumn{3}{@{}l}{\textit{Requires a running tally}}\\
VSTAT-YouTube~\cite{vstat}, cum.-state & $43.41$ & $+7.95$\\
MVBench~\cite{mvbench}, Action Count   & $54.50$ & $+4.50$\\
\midrule
\rowcolor{grouprow}\multicolumn{3}{@{}l}{\textit{General video understanding}}\\
TempCompass~\cite{liu2024tempcompass} & & \\
\quad Multiple-choice  & $74.49$ & $-0.06^{\mathrm{ns}}$\\
\quad Yes/no           & $77.62$ & $+0.04^{\mathrm{ns}}$\\
\quad Caption-matching & $85.70$ & $+0.20^{\mathrm{ns}}$\\
MMVU~\cite{zhao2025mmvu} & $55.20$ & $0.00$\\
\bottomrule
\end{tabular*}
\end{table}
 
\paragraph{Adapted modules.}
Adapting both the vision encoder and language model gives $37.44$ overall accuracy and $32.6$ on Count. Adapting only the language model lowers these to $37.12$ and $32.4$, respectively, but improves Dictionary accuracy from $27.4$ to $28.4$. Thus, updating the vision encoder provides an additional improvement, especially on Count, while language-model adaptation alone retains most of the overall improvement. This agrees with prior work showing that the language model often limits multimodal reasoning, as attention to visual tokens can weaken during decoding or be dominated by language priors~\cite{visualattn1,visualattn2,vise,visualattn3,visualattn4}.

\subsection{Generalization to real video}
\label{sec:exp:transfer}

Although trained only on synthetic clips, S$^3$T improves on real-video tasks requiring cumulative-state tracking, gaining $+7.95$ points on a predefined $185$-question VSTAT-YouTube subgroup and $+4.50$ on MVBench Action Count. Both improvements have $95\%$ paired confidence intervals excluding zero, with the VSTAT-YouTube result also remaining significant under video-level resampling (\cref{tab:realvideo-ci}). Broader video-understanding performance is largely preserved: changes across TempCompass multiple-choice, yes/no, and caption-matching tasks are statistically non-significant, while MMVU performance remains unchanged. Further results, including experiments training on real Kinetics videos, are provided in \cref{app:transfer-appendix}.

\section{Discussion}
\label{sec:discussion}

S$^3$T can be viewed as learning with privileged information~\cite{vapnik2015learning,lopez2015unifying}, where the dense view provides additional temporal evidence unavailable to the sparse-view student. Even before training, this helps the base model estimate the scene state more accurately. On held-out \emph{StateGen} clips with known end states, the $24$-frame view predicts more state tokens correctly and gives the exact count on $20.0\%$ of clips, compared to $17.0\%$ for the $12$-frame view (\cref{app:premise}). However, additional information alone is not enough, as the distillation target also matters. Rather than imitating the teacher's generated text or using ground-truth states, S$^3$T uses the teacher's predictive distribution to evaluate the student's own prediction under the denser view. This provides token-level corrections showing how additional temporal evidence would change the student's prediction, rather than a single answer to imitate. The student thus learns to estimate the scene state and refine it using additional observations. This does not require long training videos, as even $20$-second clips contain multiple state changes that require accumulating evidence over time. Despite training only on these clips, S$^3$T achieves larger improvements on videos between $20$ and $100$ seconds than on shorter videos. This suggests that learning to accumulate state over short sequences can support longer temporal horizons without equally long training videos. Together, our findings establish temporal density as a useful source of privileged information for self-distillation, supporting long-horizon visual state tracking without external supervision.

\section{Conclusion}
\label{sec:conclusion}

We introduced S$^3$T, the first fully self-contained framework for improving continuous visual state tracking in videos without external supervision. S$^3$T uses a denser view of the same clip as privileged information, distilling its predictive distribution into a sparse view of the same model without labels, an external teacher, reward model, or tracker. On VSTAT, it improves performance across cumulative-state tasks, with improvements extending to videos up to five times longer than the $20$-second clips used for training. These improvements also transfer to real videos requiring cumulative-state tracking while preserving general video-understanding performance. Our results show that temporal sampling density can provide self-generated supervision, allowing models to learn long-horizon visual state tracking from short, unlabeled videos.

\section*{Acknowledgement}
\label{sec:acknowledgement}
We acknowledge funding from the Research Council of Finland (projects 362408 and 339730). 
This work was supported by the Research Council of Finland Flagship programme: Finnish Center for Artificial Intelligence FCAI. 
We acknowledge the computational resources provided by the Aalto Science-IT project.

\section*{LLM Usage Statement}

We used generative AI tools from OpenAI and Anthropic selectively to improve grammar, phrasing, and clarity in the manuscript, and for limited coding assistance, primarily debugging and implementation support. The research ideas, methodology, experimental design, and interpretation of results were developed by the authors. All AI-assisted code was reviewed and tested, and the reported results were verified against the underlying experiments. The authors reviewed and edited all AI-assisted text to ensure accuracy and consistency with the work. We take full responsibility for the methods, results, and final manuscript.

{
    \small
    \bibliographystyle{ieeenat_fullname}

}

\clearpage
\appendix
\setcounter{page}{1}
\maketitlesupplementary

\renewcommand{\thetable}{A\arabic{table}}
\setcounter{table}{0}

\renewcommand{\thefigure}{A\arabic{figure}}
\setcounter{figure}{0}

\renewcommand{\thealgorithm}{A\arabic{algorithm}}
\setcounter{algorithm}{0}

\noindent
This supplementary material provides additional implementation details, experimental analyses, and results supporting the main paper. We begin with the training data and hyperparameters (\cref{app:data}), followed by a test showing that, before training, the dense view gives a more accurate estimate of the scene state than the sparse view (\cref{app:premise}). We then examine answer bias (\cref{app:prior}), report results on additional backbones (\cref{app:morebackbones}), and describe the temporal-pretext baselines (\cref{app:pretext-detail}). Further ablation details are provided in \cref{app:ablation-defs}, with additional analyses of training and evaluation frame budgets, teacher--student disagreement, and distillation divergences in \cref{app:framebudget,app:dpos,app:jsd_kl}. We also provide the exact probe questions, explain Probe Merging, and describe how its strength $\alpha$ is selected using held-out data (\cref{app:souping}). Finally, \cref{app:transfer-appendix} presents additional results on real videos.

\section{Training data and hyperparameters}
\label{app:data}

The $300$ training clips in \cref{sec:method:data} are generated by \emph{StateGen} using seeds $0$--$299$, which fixes the pool exactly. Each clip has a resolution of $512\times512$, lasts $20$ seconds at $12$\,fps, and contains $T=240$ source frames. One clip is sampled per step for $3000$ steps, so each clip is seen about ten times.

Each scene starts with $6$--$15$ near-identical entities on a procedurally textured background and evolves through $7$--$11$ events drawn from \{add, remove, move, swap, recolor\}. These events occur in near-equal proportions across the pool ($497$--$574$ occurrences each). Events are extended transitions with separate start and end frames. Consecutive event centres are at least $13$ frames apart (median $22$), so they do not overlap. Each clip also contains one to three occluders and continuous camera motion sampled from zoom, rotation, and pan. These repeatedly hide and reveal entities, making the current state difficult to recover from a single frame.

\emph{StateGen} also records a JSON log of the exact procedural state of each clip. Logs from the $300$ training clips are used only for offline analysis and are never given to the model, used as training targets or rewards, or used to filter the training data. Ground-truth counts from separate held-out \emph{StateGen} clips are used for model selection, but not for training (\cref{app:heldout,tab:qwen-heldout}). S$^3$T is trained only from RGB video and its own generated outputs.

\paragraph{Hyperparameters.}
The LoRA adapters use rank $r{=}32$, scaling factor $\alpha_{\mathrm{LoRA}}{=}64$, and dropout $0.05$. We use AdamW~\cite{adamw} with a learning rate of $1.5\times10^{-5}$, weight decay of $0.01$, and gradient clipping at $1.0$. Training runs in bfloat16 on a single A100-80GB GPU and takes about $2.5$ hours.

The anchor weight $\beta$ is updated at rate $\eta$ to keep $d_{\mathrm{ref}}$ near a target $\kappa$:
\begin{equation}
  \begin{aligned}
    \beta &\;\leftarrow\;
    \operatorname{clip}\!\Big(
      \beta \cdot
      \exp\!\big(
        \eta \cdot \operatorname{clip}(\delta,\,-c,\,c)
      \big),\;
      \beta_{\min},\;\beta_{\max}
    \Big), \\
    \delta &= \frac{d_{\mathrm{ref}}-\kappa}{\kappa}.
  \end{aligned}
  \label{eq:beta}
\end{equation}
Here, $\delta$ measures the relative difference between $d_{\mathrm{ref}}$ and its target. The inner clip limits each update, while the outer clip keeps $\beta$ within the allowed range. We use $\eta{=}0.05$, $c{=}3$, initialize $\beta$ at $0.05$, and set $\beta_{\max}{=}5$.

For LLaVA-OV-2-8B, we use $\kappa{=}0.06$ and $\beta_{\min}{=}10^{-3}$. Applying the same settings to Qwen3-VL-8B resulted in limited learning on held-out \emph{StateGen} clips. Increasing $\kappa$ to $0.15$ allowed larger updates but also increased variation across training seeds. We therefore introduced two stabilizers. Digit suppression reduces the JSD weight on digit positions from $1.0$ to $0.2$, while increasing the minimum anchor weight to $\beta_{\min}{=}0.05$ helps stabilize training. As shown in \cref{tab:qwen-heldout}, the combined configuration achieves the lowest held-out count error ($2.25\pm0.07$), compared with $2.50\pm0.05$ using the original settings. We selected these settings using held-out clips and training diagnostics before evaluating the configurations on VSTAT. The VSTAT results in \cref{tab:qwen-heldout} show why both stabilizers are needed. Without them, the higher anchor target falls below the base model's $34.11$ on average, and one run drops to $24.43$. Each stabilizer alone reduces this variation, and together they give the most accurate and stable configuration.

\begin{table}[h]
\centering\scriptsize
\setlength{\tabcolsep}{3pt}
\caption{\textbf{Qwen3-VL-8B stabilization on held-out \emph{StateGen} clips.} Mean absolute count error for different anchor settings and digit suppression, evaluated independently of VSTAT. Values are reported across independent training runs. VSTAT accuracy was measured after selection and is the mean $\pm$ standard deviation over three runs.}
\label{tab:qwen-heldout}
\begin{tabular}{@{}lccccc@{}}
\toprule
Configuration
& $\kappa$
& $\beta_{\min}$
& \begin{tabular}[c]{@{}c@{}}Digit\\sup.\end{tabular}
& \begin{tabular}[c]{@{}c@{}}Held-out\\count MAE $\downarrow$\end{tabular}
& VSTAT $\uparrow$\\
\midrule
Shared settings          & 0.06 & $10^{-3}$ & No  & $2.50\pm0.05$ & $34.24$\\
Higher anchor target     & 0.15 & $10^{-3}$ & No  & $3.40\pm0.70$ & $31.23\pm4.32$\\
With digit suppression   & 0.15 & $10^{-3}$ & Yes & $2.60\pm0.22$ & $35.36\pm1.15$\\
With higher anchor floor & 0.15 & 0.05      & No  & $2.95\pm0.30$ & $32.98\pm2.75$\\
Final configuration      & 0.15 & 0.05      & Yes & $\mathbf{2.25\pm0.07}$ & $\mathbf{36.47\pm0.42}$\\
\bottomrule
\end{tabular}
\end{table}

\paragraph{Selection protocol.}
We evaluate all ablations on the same $1{,}500$ VSTAT questions, while selecting training configurations independently of VSTAT. The training objective and anchor, $3000$-step schedule, LoRA rank and placement, optimizer and learning rate, training clips and generator settings, and the two probes were fixed independently of the benchmark. The $12\!\to\!24$ temporal views were chosen based on our temporal-density hypothesis, while the merging strength $\alpha$ was selected using held-out \emph{StateGen} clips (\cref{app:heldout}). Qwen-specific stabilization settings were likewise selected using held-out clips and training diagnostics (\cref{tab:qwen-heldout}). The generator's states for these held-out clips are used only for this selection, never during training. Alternative teacher views, frame budgets, divergences, checkpoint steps, and merging strengths are evaluated on VSTAT to examine their effects, rather than to select the final configuration.

To assess the reliability of the main improvements, we also compare the base and S$^3$T models on the same VSTAT questions using paired bootstrap resampling (\cref{tab:paired-vstat}). The $95\%$ confidence intervals exclude zero for both LLaVA-OV-2-8B and Qwen3-VL-8B, supporting improvements beyond variation from the sampled evaluation questions. These intervals measure uncertainty in the accuracy differences and are distinct from the standard errors over evaluation questions reported by \texttt{lmms-eval}~\cite{zhang2024lmmsevalrealitycheckevaluation} in the main results.

\begin{table}[h]
\centering\footnotesize
\caption{\textbf{Paired accuracy improvements on VSTAT.}
Accuracy differences between each base model and its S$^3$T-trained counterpart on the same evaluation questions. Intervals are computed using paired bootstrap resampling over evaluation questions. Both $95\%$ confidence intervals exclude zero.}
\label{tab:paired-vstat}
\begin{tabular}{lcc}
\toprule
Backbone & $\Delta$ accuracy & 95\% CI\\
\midrule
LLaVA-OV-2-8B
& $+2.61$ & [$+0.95$, $+3.77$]\\
Qwen3-VL-8B
& $+2.44$ & [$+1.05$, $+3.85$]\\
\bottomrule
\end{tabular}
\end{table}

We also evaluate the full S$^3$T pipeline across five independent training runs for each of three backbones (\cref{tab:seed-reliability}). All five runs improve on their respective base models. The mean accuracy is $37.36\pm0.35$ for LLaVA-OV-2-8B, $36.47\pm0.42$ for Qwen3-VL-8B, and $30.69\pm0.23$ for InternVL3.5-8B, where the variation is measured across training runs. These results show that the improvements are not limited to a single training seed.

\begin{table}[h]
\centering\footnotesize
\caption{\textbf{Reliability across training seeds.}
VSTAT accuracy (\%) over five independent runs of the full S$^3$T pipeline, including both probe-specific adapters and their merge. Results are mean $\pm$ standard deviation across runs. Improvements are measured relative to each backbone's base model.}
\label{tab:seed-reliability}
\begin{tabular}{lccc}
\toprule
Backbone & VSTAT & $\Delta$ base & Improved runs\\
\midrule
LLaVA-OV-2-8B
& $37.36\pm0.35$ & $+2.53$ & $5/5$\\
Qwen3-VL-8B
& $36.47\pm0.42$ & $+2.36$ & $5/5$\\
InternVL3.5-8B
& $30.69\pm0.23$ & $+0.88$ & $5/5$\\
\bottomrule
\end{tabular}
\end{table}

\section{Is the dense view a more reliable teacher?}
\label{app:premise}

S$^3$T relies on the teacher's dense view giving a better estimate of the scene state than the student's sparse view. We test this directly, before any training, by asking the frozen base model the same probe on $300$ held-out \emph{StateGen} clips, once with $12$ frames and once with $24$. These are exactly the student and teacher views at the start of training. Since the clips are procedurally generated, their true end states are known, so each answer can be scored against the actual scene state. We report the percentage of predicted state tokens that are correct and the percentage of clips for which the predicted count is exactly correct. We use these ground-truth states only for this analysis, never for training.

\begin{table}[h]
\centering\footnotesize
\caption{\textbf{The dense view gives a more accurate estimate of the scene state.} Answers of the frozen base model to the same probe with $12$ frames (student view) and $24$ frames (teacher view), scored against the known end states of $300$ held-out \emph{StateGen} clips.}
\begin{tabular}{l c c c}
\toprule
Measure & $12$ frames & $24$ frames & Difference\\
\midrule
Correct state tokens (\%) $\uparrow$ & $63.38$ & $\mathbf{64.45}$ & $+1.07$\\
Exactly correct count (\%) $\uparrow$ & $17.0$ & $\mathbf{20.0}$ & $+3.0$\\
\bottomrule
\end{tabular}
\label{tab:premise}
\end{table}

With $24$ rather than $12$ frames, the model predicts more state tokens correctly ($64.45\%$ vs.\ $63.38\%$) and gives the exact count on more clips ($20.0\%$ vs.\ $17.0\%$; \cref{tab:premise}). Both answers come from the same model and probe and are scored against the same ground truth, so the difference comes only from the additional frames. The dense view therefore gives a more accurate estimate of the scene state, which is what makes it useful as a teacher. The gap is modest, but S$^3$T does not need a perfect teacher, only one whose estimate is more accurate than the student's. S$^3$T distills this more accurate estimate into the sparse-view student through the teacher's predictive distribution.
\section{Answer-bias analysis}
\label{app:prior}

A model could improve its VSTAT accuracy without using the video by favoring the most common answer for each task. We examine whether this explains S$^3$T's improvement. For each question, we define the answer prior as the most frequent correct answer among questions from the same source task, breaking ties by the answer that appears first in the benchmark. This prior does not use the video and is used only to divide the benchmark into two groups. Of the $1{,}500$ questions, $366$ have correct answers matching the prior, while the remaining $1{,}134$ do not. On the latter group, predicting the prior would give an incorrect answer.

\begin{table*}[t]
\centering\footnotesize
\setlength{\tabcolsep}{4pt}
\caption{\textbf{Extended experiments on other base models.} VSTAT accuracy (\%) with the same objective and training data as in \cref{tab:main_results}. The base and Video-Zero rows are the same runs as in \cref{tab:main_results}. Each block compares a base model with the same model trained with S$^3$T and, where available, with Video-Zero trained on the same base. $\Delta$ is the change in average accuracy relative to the base model in that block.}
\begin{tabular}{l cc ccc cccc}
\toprule
 & & & \multicolumn{3}{c}{State Element} & \multicolumn{4}{c}{State Structure}\\
\cmidrule(lr){4-6}\cmidrule(lr){7-10}
Method & Avg & $\Delta$ & Count & Loc. & Attr. & Atom. & Seq. & Set & Dict\\
\midrule
\textit{Qwen3-VL-4B (base)} & \textit{31.44}\,{\scriptsize$\pm$1.06} & -- & \textit{27.7}\,{\scriptsize$\pm$1.4} & \textit{33.7}\,{\scriptsize$\pm$2.3} & \textit{36.6}\,{\scriptsize$\pm$2.2} & \textit{31.9}\,{\scriptsize$\pm$1.5} & \textit{32.5}\,{\scriptsize$\pm$3.2} & \textit{38.5}\,{\scriptsize$\pm$2.6} & \textit{24.4}\,{\scriptsize$\pm$2.2}\\
Video-Zero-4B~\cite{evo3} & 31.62\,{\scriptsize$\pm$1.06} & $+0.18$ & 27.7\,{\scriptsize$\pm$1.4} & \textbf{34.5}\,{\scriptsize$\pm$2.4} & 36.5\,{\scriptsize$\pm$2.2} & 32.1\,{\scriptsize$\pm$1.5} & 32.3\,{\scriptsize$\pm$3.2} & 37.8\,{\scriptsize$\pm$2.5} & \textbf{25.5}\,{\scriptsize$\pm$2.2}\\
\rowcolor{ourrow}S$^3$T & \textbf{32.60}\,{\scriptsize$\pm$1.06} & $\mathbf{+1.16}$ & \textbf{30.1}\,{\scriptsize$\pm$1.4} & 34.3\,{\scriptsize$\pm$2.4} & \textbf{37.3}\,{\scriptsize$\pm$2.2} & \textbf{33.4}\,{\scriptsize$\pm$1.5} & \textbf{33.6}\,{\scriptsize$\pm$3.2} & \textbf{39.3}\,{\scriptsize$\pm$2.5} & 25.4\,{\scriptsize$\pm$2.2}\\
\midrule
\textit{Qwen3-VL-8B (base)} & \textit{34.11}\,{\scriptsize$\pm$1.08} & -- & \textit{32.8}\,{\scriptsize$\pm$1.4} & \textit{38.4}\,{\scriptsize$\pm$2.4} & \textit{32.7}\,{\scriptsize$\pm$2.1} & \textit{35.5}\,{\scriptsize$\pm$1.5} & \textit{33.5}\,{\scriptsize$\pm$3.2} & \textit{36.5}\,{\scriptsize$\pm$2.5} & \textit{29.6}\,{\scriptsize$\pm$2.3}\\
Video-Zero-8B~\cite{evo3} & 34.17\,{\scriptsize$\pm$1.07} & $+0.06$ & 32.1\,{\scriptsize$\pm$1.4} & \textbf{39.1}\,{\scriptsize$\pm$2.4} & 33.3\,{\scriptsize$\pm$2.1} & 35.3\,{\scriptsize$\pm$1.5} & 34.5\,{\scriptsize$\pm$3.2} & 38.1\,{\scriptsize$\pm$2.5} & 28.5\,{\scriptsize$\pm$2.3}\\
\rowcolor{ourrow}S$^3$T & \textbf{36.55}\,{\scriptsize$\pm$1.08} & $\mathbf{+2.44}$ & \textbf{35.5}\,{\scriptsize$\pm$1.4} & 38.4\,{\scriptsize$\pm$2.4} & \textbf{36.7}\,{\scriptsize$\pm$2.2} & \textbf{37.2}\,{\scriptsize$\pm$1.5} & \textbf{35.9}\,{\scriptsize$\pm$3.2} & \textbf{40.5}\,{\scriptsize$\pm$2.5} & \textbf{32.5}\,{\scriptsize$\pm$2.3}\\
\midrule
\textit{Cambrian-S-7B (base)} & \textit{34.70}\,{\scriptsize$\pm$1.07} & -- & \textit{34.2}\,{\scriptsize$\pm$1.4} & \textit{34.0}\,{\scriptsize$\pm$2.4} & \textit{36.4}\,{\scriptsize$\pm$2.1} & \textit{35.1}\,{\scriptsize$\pm$1.5} & \textit{31.0}\,{\scriptsize$\pm$3.2} & \textit{39.8}\,{\scriptsize$\pm$2.4} & \textit{32.3}\,{\scriptsize$\pm$2.3}\\
\rowcolor{ourrow}S$^3$T & \textbf{35.60}\,{\scriptsize$\pm$1.07} & $\mathbf{+0.90}$ & \textbf{34.7}\,{\scriptsize$\pm$1.4} & \textbf{34.7}\,{\scriptsize$\pm$2.1} & \textbf{38.3}\,{\scriptsize$\pm$2.2} & \textbf{35.6}\,{\scriptsize$\pm$1.5} & \textbf{32.3}\,{\scriptsize$\pm$3.2} & \textbf{41.3}\,{\scriptsize$\pm$2.9} & \textbf{33.4}\,{\scriptsize$\pm$2.5}\\
\midrule
\textit{InternVL3.5-8B (base)} & \textit{29.81}\,{\scriptsize$\pm$1.05} & -- & \textit{22.7}\,{\scriptsize$\pm$1.3} & \textit{37.8}\,{\scriptsize$\pm$2.4} & \textbf{\textit{35.9}}\,{\scriptsize$\pm$2.2} & \textit{26.8}\,{\scriptsize$\pm$1.4} & \textbf{\textit{37.1}}\,{\scriptsize$\pm$3.3} & \textit{37.7}\,{\scriptsize$\pm$2.5} & \textit{25.8}\,{\scriptsize$\pm$2.2}\\
\rowcolor{ourrow}S$^3$T & \textbf{30.74}\,{\scriptsize$\pm$1.06} & $\mathbf{+0.93}$ & \textbf{23.9}\,{\scriptsize$\pm$1.3} & \textbf{39.2}\,{\scriptsize$\pm$2.4} & \textbf{35.9}\,{\scriptsize$\pm$2.2} & \textbf{27.9}\,{\scriptsize$\pm$1.4} & 35.5\,{\scriptsize$\pm$3.2} & \textbf{38.8}\,{\scriptsize$\pm$2.5} & \textbf{27.6}\,{\scriptsize$\pm$2.3}\\
\bottomrule
\end{tabular}
\label{tab:morebackbones}
\end{table*}
 
\begin{table}[h]
\centering\footnotesize
\caption{\textbf{VSTAT accuracy grouped by whether the correct answer matches the most frequent answer for its task.} Comparison of the base model LLaVA-OV-2-8B and S$^3$T.}
\label{tab:prior}
\begin{tabular}{l c c c}
\toprule
Question group & $n$ & Base & S$^3$T \\
\midrule
Correct answer matches the prior & $366$ & $38.03$ & $\mathbf{43.69}$ \\
Correct answer differs & $1134$ & $33.79$ & $\mathbf{35.42}$ \\
\midrule
All questions & $1500$ & $34.83$ & $\mathbf{37.44}$ \\
\bottomrule
\end{tabular}
\end{table}

S$^3$T improves accuracy in both groups (\cref{tab:prior}). On the $1{,}134$ questions where the prior is incorrect, accuracy improves by $+1.63$ points, with a $95\%$ bootstrap confidence interval of $[+0.10,+3.21]$ that excludes zero. These questions account for nearly half of the overall improvement. The improvement is also distributed across both groups, with $24.5\%$ of the questions showing improved predictions belonging to the prior-matching group, compared with its $24.4\%$ share of the benchmark. However, the improvement is larger when the correct answer matches the prior ($+5.66$ vs.\ $+1.63$), and the proportion of predictions matching the prior increases from $19.7\%$ to $23.0\%$ after training. Thus, S$^3$T does show some shift toward common answers, but this cannot fully explain its improvement, since accuracy also increases on questions where the prior is incorrect. The result is consistent across four tie-breaking rules (first or last appearance, and lexicographically lowest or highest answer). The improvement on questions where the prior is incorrect ranges from $+1.63$ to $+1.88$, with the $95\%$ bootstrap confidence interval excluding zero in every case.

\section{Extended experiments on other base models}
\label{app:morebackbones}

\cref{tab:morebackbones} gives additional comparisons on more base models, using the same objective and training data. On Qwen3-VL-8B, S$^3$T improves average accuracy by $+2.44$, while Video-Zero trained on the same base improves it by only $+0.06$. On Qwen3-VL-4B, S$^3$T improves by $+1.16$, compared with $+0.18$ for Video-Zero. S$^3$T also improves Cambrian-S-7B and InternVL3.5-8B.

\section{Temporal-pretext baselines}
\label{app:pretext-detail}

\cref{tab:pretext} compares S$^3$T with four established temporal self-supervised tasks. Each uses unlabeled clips to construct its own training targets, without relying on annotations or generator metadata. All baselines use the same LLaVA-OV-2-8B backbone, $300$ training clips, LoRA rank $r{=}32$, $3000$ training steps, and reference anchor as S$^3$T.

For the first three tasks, a transformation $T$ is applied to clip $x$, and its identity determines a short text label $\ell(T)$. The model learns to predict this label using cross-entropy, with the same reference anchor and adaptive $\beta$ schedule as S$^3$T (\cref{eq:loss}):
\begin{equation}
  \mathcal{L}_{\text{pretext}}
  =
  \operatorname{CE}\!\big(f_{\theta+\phi}(T(x)),\ell(T)\big)
  \;+\;\beta\, d_{\mathrm{ref}} .
  \label{eq:pretext}
\end{equation}
The three tasks differ in how the frames are transformed and what the model predicts:
\begin{itemize}
\item \textbf{Frame order}~\cite{frameorder,frame2} divides the student's $12$ frames into four consecutive segments and shuffles them using a random permutation $\pi$. The model predicts the original temporal position of each displayed segment as a four-digit sequence (e.g.\ \emph{3 1 4 2}).
\item \textbf{Arrow of time}~\cite{arrow,arrow2} reverses the frame order with probability $0.5$. The model predicts whether the clip is played \emph{forward} or \emph{backward}.
\item \textbf{Pace}~\cite{pace,pace2} samples $12$ frames at a stride of $1$, $2$, or $4$ from a randomly selected portion of the clip. The model predicts the playback speed as \emph{normal}, \emph{2x}, or \emph{4x}.
\end{itemize}

\textbf{Masked temporal infill}~\cite{videomae} instead uses the same distillation objective as S$^3$T. A contiguous window of nine source frames, centered at a random interior position, is removed from the student's view, while the teacher receives the full clip at the same $12$-frame budget. The teacher evaluates the student's answer to the same probe, using the original loss in \cref{eq:loss}. Unlike S$^3$T, which gives the teacher a denser view, this baseline gives the student an incomplete view while keeping the teacher's frame budget unchanged. None of the four baselines uses ground-truth states or labels from the clip generator.

None of these baselines reproduces S$^3$T's improvement in state tracking (\cref{tab:pretext}). Pace performs best overall at $35.45$, but achieves less than a quarter of S$^3$T's improvement. Frame order performs best on Dictionary ($30.5$), although its overall accuracy improves only slightly. These tasks can often be solved using local appearance or motion cues, without tracking the scene state throughout the clip.

Masked infill provides a particularly useful comparison because it retains S$^3$T's distillation objective, reference anchor, and self-generated supervision, but changes how the teacher and student views differ. Its accuracy falls to $34.25$, below the $34.83$ base model. This suggests that learning from an incomplete view does not provide the same benefit as distilling predictions from a temporally denser view. Together, these comparisons show that temporal self-supervision alone is insufficient to reproduce S$^3$T's improvement. What matters is using additional temporal observations to provide a more informative teacher distribution for the student.

\section{Ablation definitions}
\label{app:ablation-defs}

This section provides the definitions and implementation details of the ablations in \cref{tab:ablation_table}, along with additional teacher-view comparisons. Unless stated otherwise, the student uses $12$ frames, the teacher uses $24$ frames, and all other training settings remain unchanged. Our default teacher samples $24$ frames uniformly across the full clip.

\paragraph{Identical $12\!\to\!12$ (no privileged view).}
The teacher receives the same $12$ frames as the student, making their inputs identical. Everything else remains unchanged, and the teacher--student divergence is zero by construction. This tests what happens when the privileged information is removed from self-distillation.

\paragraph{Shifted equal-density teacher ($12\!\to\!12$).}
Both views contain $12$ evenly spaced frames, but the teacher's frames are shifted by half a sampling interval, placing them between the student's frames. The two views have the same frame count and density but contain frames from different timestamps. This separates the benefit of additional frames from that of observing different frames.

\paragraph{Temporal zoom (windowed).}
Instead of sampling across the full clip, the teacher focuses on a short window around a detected scene change. Given a boundary at frame $b$, it samples uniformly from $[b-8,\,b+8]$, covering $17$ source frames or approximately $1.4$ seconds of the $20$-second clip. In this experiment, the teacher uses the same $12$-frame budget as the student. This allows us to compare frame placement without changing the number of frames. Concentrating the frames within a shorter window increases their local density but reduces temporal coverage, so the experiment tests whether a more detailed view of a short interval is sufficient.

\paragraph{Event-aware teacher.}
This variant allocates more teacher frames around scene changes detected by the frozen model, without using labels or an external detector. At each candidate time, the model examines two short, non-overlapping windows on either side of that point using the same state probe. We measure the token-averaged difference between the resulting next-token distributions, with larger differences indicating a greater change in the model's reading of the scene.

To check the quality of these detected boundaries, we compare them against the known event times in \emph{StateGen}. The detector achieves an AUROC of $0.851$~[$0.833$,$0.869$], compared with $0.501\pm0.014$ under shuffled labels, approximately $25$ standard deviations above chance. Its performance is also consistent when candidates are ranked within time-matched groups (AUROC $0.81$--$0.85$), suggesting that detection does not depend on fixed event positions. Across $253$ clips, it produces $211$ distinct boundary sets, and $75.5\%$ of its proposed boundaries match scheduled events. We further compare it with shot-cut detection methods on $253$ synthetic clips (\cref{tab:scenecut}). These methods achieve relatively high recall ($0.78$ and $0.73$) but lower precision ($0.61$ and $0.56$), meaning that many detected boundaries do not correspond to actual state changes. In contrast, our detector achieves $0.82$ precision and $0.74$ recall, allowing it to identify state changes more reliably. These results suggest that poor event detection is unlikely to explain the event-aware teacher's lower accuracy.

\begin{table}[h]
\centering\footnotesize
\caption{\textbf{Detection of state changes compared with shot-cut detection.} Results on $253$ synthetic clips. AUROC measures how often a true event boundary is ranked above a non-boundary, with $0.5$ corresponding to chance.}
\begin{tabular}{lcc}
\toprule
Detector & AUROC & F1\\
\midrule
Model before/after JSD (ours) & $\mathbf{0.851}$ & $\mathbf{0.776}$\\
ffmpeg scene-score            & $0.645$ & $0.63$--$0.69$\\
PySceneDetect                 & $0.576$ & $0.63$--$0.69$\\
Random                        & $0.503$ & --\\
\bottomrule
\end{tabular}
\label{tab:scenecut}
\end{table}

\paragraph{Multi-scale teacher.}
Instead of using a single $24$-frame view, the teacher combines predictions from several frame budgets, either $\{16,32,48\}$ or $\{12,24,48\}$. The student is then trained to match the combined teacher prediction.

\paragraph{Adaptive teachers.}
These variants choose the teacher's frame budget separately for each clip. The \emph{router} uses a lightweight rule to select the number of frames, while the \emph{budget-controlled} variant adjusts the teacher density to keep $d_{\text{pos}}$ within a target range. Both aim to provide additional temporal information while keeping the teacher's prediction learnable from the sparse student view.

\paragraph{Frame ratio.}
These experiments vary the student and teacher frame budgets (\cref{app:framebudget}). The $12\!\to\!36$ and $12\!\to\!48$ settings increase the teacher budget while keeping the student at $12$ frames. The $8\!\to\!32$ and $24\!\to\!48$ settings change both budgets. Together, these comparisons examine how performance depends on the frame ratio and the number of frames available to the student.

\paragraph{SFT (cross-entropy on teacher text).}
The teacher generates an answer greedily from its dense $24$-frame view, and the student learns to reproduce that text using cross-entropy on its $12$-frame view. The two views and reference anchor remain unchanged. The only difference is that the student learns from the teacher's generated token sequence rather than its full next-token predictive distribution. This directly compares supervision from generated text with distribution-based distillation.

\paragraph{CE on the generator's true state.}
This supervised reference trains the student directly on the ground-truth state string recorded by the clip generator. The student receives the same $12$-frame view and uses the same LoRA configuration, training schedule, and reference anchor as S$^3$T. The distillation term is replaced with cross-entropy on the correct state, allowing us to compare self-generated distribution-based supervision with direct ground-truth supervision under otherwise matched training conditions. Unlike S$^3$T, this experiment uses the generator's state labels during training.

\paragraph{JSD to the generator's true state.}
This is a second supervised reference experiment rather than a competing method. The objective, both views, and reference anchor remain unchanged, but the text being scored is the ground-truth state string recorded by the clip generator rather than the model's own generated answer. This tests whether using the correct state as the scored answer is sufficient to improve distillation. Unlike direct CE training on the true state, this variant retains the distribution-based objective. These two supervised references are the only experiments in which generator-provided state labels are used during training.

\paragraph{Ground-truth supervision and training steps.}
Direct CE training on the generator's true state reaches $34.05$ overall accuracy, while JSD toward the same state gives $33.90$ and SFT on the teacher's generated text gives $34.45$. All three remain below the base-model accuracy of $34.83$, with changes of $-0.78$, $-0.93$, and $-0.38$, respectively, whereas S$^3$T reaches $37.44$ (\cref{tab:ablation_table}).

We also compare supervised fine-tuning and S$^3$T at different training steps (\cref{tab:sft-steps}). SFT remains below the base model at every evaluated checkpoint, while S$^3$T improves beyond it from $600$ steps onward. This suggests that the difference between the objectives is not simply due to the supervised run receiving too few training steps. One possible explanation is that the correct state may not always be recoverable from the student's sparse view, making a fixed ground-truth answer less useful than the dense teacher's distribution over the student's own prediction.

\begin{table}[h]
\centering\footnotesize
\caption{\textbf{Training-step comparison.} VSTAT accuracy (\%) for supervised fine-tuning on the teacher's generated text and S$^3$T at different training checkpoints. Both use the same training data and backbone.}
\label{tab:sft-steps}
\begin{tabular}{lccccc}
\toprule
Steps & $300$ & $600$ & $1200$ & $1800$ & $2400$\\
\midrule
SFT & $34.25$ & $34.39$ & $34.45$ & $33.95$ & $34.09$\\
S$^3$T & $\mathbf{34.62}$ & $\mathbf{35.61}$ & $\mathbf{35.75}$ & $\mathbf{35.18}$ & $\mathbf{35.70}$\\
\bottomrule
\end{tabular}
\end{table}

\paragraph{Reference anchor.}
By default, the reference anchor measures the JSD between the adapted student and the frozen base model on the same sparse view. Its contribution is weighted by $\beta$, which follows the adaptive rule in \cref{sec:method:tsd}. The ``no anchor'' variant sets $\beta{=}0$, leaving the student trained only with the teacher distillation term.

\paragraph{Frozen vision encoder.}
LoRA is applied only to the language model, while the vision encoder remains frozen at its base weights. All other settings, including the two probes and Probe Merging, remain unchanged, except that the merging strength is set to $\alpha{=}1.5$ (\cref{app:souping}).

\paragraph{Results for alternative teachers.}
\cref{tab:ablation-extra} compares the additional teacher-view variants. None outperforms the fixed $12\!\to\!24$ dense full-clip teacher. The strongest alternative is the budget-controlled adaptive teacher, which achieves $36.49$ overall accuracy compared with $37.44$ for the fixed view. It also requires a separate decision rule to select the frame budget for each clip. We therefore use the simpler fixed teacher, which needs no per-clip frame-budget selection.

\begin{table}[h]
\centering\footnotesize
\caption{\textbf{Alternative teacher views.} VSTAT accuracy (\%) with the $12$-frame student and training objective fixed, while only the teacher view changes. The first row repeats our default setting from \cref{tab:ablation_table}.}
\begin{tabular}{l c c c}
\toprule
Teacher view & Overall & Count & Dict\\
\midrule
Dense full-clip view $12\!\to\!24$ (ours) & $\mathbf{37.44}$ & $\mathbf{32.6}$ & $27.4$\\
Event-aware                           & $35.11$ & $28.5$ & $27.5$\\
Multi-scale $\{16,32,48\}$            & $36.23$ & $29.9$ & $\mathbf{28.1}$\\
Multi-scale $\{12,24,48\}$            & $36.04$ & $29.3$ & $28.0$\\
Adaptive (router)                     & $36.26$ & $30.9$ & $27.9$\\
Adaptive (budget-controlled)          & $36.49$ & $30.4$ & $27.7$\\
\bottomrule
\end{tabular}
\label{tab:ablation-extra}
\end{table}

\section{Frame budgets}
\label{app:framebudget}

\cref{fig:teacherstudent} shows how the improvement varies with the student and teacher frame budgets during training for LLaVA-OV-2-8B. The largest improvements occur around a $12$-frame student and a $24$-frame teacher. The student budget is more sensitive than the teacher budget. Above approximately $20$ student frames, training reduces accuracy across all tested teacher budgets, while a wider range of teacher budgets still gives improvements. This is consistent with learning from privileged information. If the teacher sees too few additional frames, it provides little new evidence, while a much denser teacher view may produce predictions that are harder for the student to learn from.

\begin{figure}[h]
  \centering
  \newsavebox{\vstatgainbox}%
  \begin{lrbox}{\vstatgainbox}%
\pgfplotsset{colormap={rdbucmap}{rgb255=(5,48,97) rgb255=(23,82,144) rgb255=(42,113,178) rgb255=(63,142,192) rgb255=(107,172,209) rgb255=(155,201,224) rgb255=(194,221,236) rgb255=(224,236,243) rgb255=(247,246,246) rgb255=(251,229,216) rgb255=(251,204,180) rgb255=(245,170,137) rgb255=(228,128,102) rgb255=(208,85,72) rgb255=(186,40,50) rgb255=(147,14,38) rgb255=(103,0,31)}}

\end{lrbox}%
  \resizebox{\columnwidth}{!}{\usebox{\vstatgainbox}}
  \caption{\textbf{The improvement depends on both frame budgets.} Runs use the same base model, data, and training schedule, varying only the student and teacher frame counts. The dotted line marks $\text{teacher}=\text{student}$. Warm colors indicate improvements, cool colors indicate decreases, and the dashed line marks zero change.}
  \label{fig:teacherstudent}
\end{figure}

\cref{tab:frameratio} compares different frame budgets with our default $12\!\to\!24$ setting, keeping all other settings fixed. With the student fixed at $12$ frames, increasing the teacher budget from $24$ to $36$ and $48$ progressively reduces accuracy. Changing both budgets also performs worse. In particular, $24\!\to\!48$ retains the same $2\times$ ratio as our default but scores $2.20$ points below the base model. This shows that the frame ratio alone does not determine the improvement, and the number of frames available to the student also matters.

\begin{table}[h]
\centering\footnotesize
\caption{\textbf{Training frame ratio for LLaVA-OV-2-8B.} VSTAT accuracy (\%) when only the student and teacher frame counts are changed.}
\begin{tabular}{l c c c}
\toprule
Student $\to$ teacher & Overall & Count & Dict\\
\midrule
$12\!\to\!24$ (ours) & $\mathbf{37.44}$ & $\mathbf{32.6}$ & $27.4$\\
$12\!\to\!36$        & $35.56$ & $27.6$ & $\mathbf{27.8}$\\
$12\!\to\!48$        & $34.67$ & $27.1$ & $27.5$\\
$8\!\to\!32$         & $33.81$ & $28.6$ & $26.7$\\
$24\!\to\!48$        & $32.63$ & $26.5$ & $27.5$\\
\bottomrule
\end{tabular}
\label{tab:frameratio}
\end{table}

\cref{tab:framebudget} examines how performance changes with the number of frames used at evaluation. The base model improves from $30.67$ at $8$ frames to $34.51$ at $24$ frames, but gains little from additional frames, reaching $34.83$ at $64$ frames. S$^3$T performs similarly to the base model at $12$ and $24$ frames, but improves more at larger budgets, reaching $37.44$ at the standard $64$-frame setting. These results show that the improvement is not fully realized at the student's training frame budget and becomes more pronounced when additional frames are available at evaluation. While this is consistent with learning to use temporal evidence more effectively, the results do not isolate the mechanism responsible for this behavior.

\begin{table}[t]
\centering
\caption{\textbf{VSTAT accuracy at different evaluation frame budgets on LLaVA-OV-2-8B.} The same two checkpoints are evaluated at every budget, changing only the number of frames provided.}
\label{tab:framebudget}
\begin{tabular}{lccccccc}
\toprule
Frames & $8$ & $12$ & $24$ & $32$ & $64$ \\
\midrule
Base & $30.67$ & $31.60$ & $34.51$ & $33.95$ & $\mathbf{34.83}$ \\
S$^3$T & $31.16$ & $31.89$ & $34.23$ & $35.02$ & $\mathbf{37.44}$ \\
\bottomrule
\end{tabular}
\end{table}

\section{Teacher--student disagreement}
\label{app:dpos}

Distilling toward the generator's true state gives a correct target yet reduces accuracy, so a more accurate target does not necessarily lead to better distillation. To better understand this result, we examine teacher--student disagreement during training. \cref{tab:dpos} reports the mean $d_{\mathrm{pos}}$ across all $3000$ training steps alongside the final VSTAT accuracy.

\begin{table}[h]
\centering\footnotesize
\caption{\textbf{Teacher--student disagreement during LLaVA-OV-2-8B training and final accuracy.} Mean $d_{\mathrm{pos}}$ over $3000$ steps and VSTAT accuracy (\%) of the resulting model.}
\begin{tabular}{lcc}
\toprule
Setting & Mean $d_{\mathrm{pos}}$ & VSTAT\\
\midrule
JSD to the generator's true state & $0.0013$ & $33.90$\\
Shifted $12\!\to\!12$ & $0.0051$ & $35.14$\\
Dense full-clip view $12\!\to\!24$ (ours) & $\mathbf{0.0054}$ & $\mathbf{37.44}$\\
\bottomrule
\end{tabular}
\label{tab:dpos}
\end{table}

The shifted and dense full-clip teachers have similar levels of disagreement, yet their final accuracies differ considerably. This indicates that the amount of teacher--student disagreement alone does not explain the improvement. The teacher must provide additional information that the student can learn from its sparse view. A teacher that is too similar to the student provides little new supervision, while one whose predictions depend on information unavailable to the student may be difficult to learn from. The results therefore suggest a balance between providing new temporal evidence and keeping the teacher's predictions learnable from the student view. The true-state experiment uses a different answer string, so its $d_{\mathrm{pos}}$ is not directly comparable with the other settings. Although this analysis covers only three configurations, it is consistent with the frame-budget results in \cref{app:framebudget}, where increasing the teacher's frame budget does not always improve performance. A more precise understanding of this relationship remains an open question.

\section{Choice of distillation divergence}
\label{app:jsd_kl}

We compare JSD with forward and reverse KLD while keeping the training setup fixed, including the JSD reference anchor. All runs are on LLaVA-OV-2-8B, and use $3000$ training steps and three seeds. Although the optimized divergence differs across objectives, a separate JSD monitor remains near $0.005$ in all three settings, indicating similar levels of teacher--student disagreement. Both KLD variants remain trainable but produce larger gradient norms and spikes (\cref{fig:kl_jsd}), with maximum norms of $15.96$ and $29.32$, compared with $5.53$ for JSD. JSD also achieves higher VSTAT accuracy ($37.44$ versus $35.19$ and $35.14$; \cref{tab:jsd_kl}), making it the more stable and effective objective in our experiments.

\begin{figure}[h]
  \setlength{\abovecaptionskip}{0pt}
  \centering

   \vspace*{-4pt}
  \caption{\textbf{JSD provides more stable training than forward or reverse KLD.} Training divergence \textit{(left)} and pre-clipping gradient norm \textit{(right)} over $3000$ steps, averaged across three seeds. Only the distillation divergence changes between runs.}
  \label{fig:kl_jsd}
\end{figure}

\begin{table}[h]
  \centering\footnotesize
  \caption{\textbf{Comparison of distillation divergences.} Mean and maximum training divergence, pre-clipping gradient norm, and downstream accuracy over $3000$ training steps. Training statistics are averaged across three seeds. The reference anchor remains JSD in all runs.}
  \resizebox{\columnwidth}{!}{
  \begin{tabular}{lccccccc}
    \toprule
    & \multicolumn{2}{c}{Training divergence}
    & \multicolumn{2}{c}{Pre-clip grad.\ norm}
    & \multicolumn{3}{c}{Accuracy (\%)} \\
    \cmidrule(lr){2-3}
    \cmidrule(lr){4-5}
    \cmidrule(lr){6-8}
    Divergence & mean & max & mean & max & VSTAT & Count & Dictionary \\
    \midrule
    JSD (ours)    & \textbf{0.0054} & \textbf{0.133} & \textbf{0.237} & \textbf{5.53}
                  & \textbf{37.44} & \textbf{32.6} & \textbf{27.4} \\
    Forward KLD   & 0.0208 & 1.175 & 0.873 & 15.96
                  & 35.19 & 27.1 & 27.0 \\
    Reverse KLD   & 0.0204 & 0.448 & 0.918 & 29.32
                  & 35.14 & 27.9 & 26.0 \\
    \bottomrule
  \end{tabular}}
  \label{tab:jsd_kl}
\end{table}

\section{\texorpdfstring{Probes and Probe Merging}{Probes and Probe Merging}}
\label{app:souping}

S$^3$T trains two runs that differ only in the probe $q$ (\cref{sec:method:tsd}) and combines their weight updates into a single model, following model soups~\cite{souping}. We provide the exact probes, examine how their design affects learning, and describe the merging procedure and selection of its strength $\alpha$.

\subsection{Probes}

Every probe begins with the same two sentences: \emph{``This is a short video of colored balls moving on a background. Some balls may be briefly hidden behind moving bars, and the camera may pan or zoom.''} The two probes used in S$^3$T are defined as follows.

\begin{itemize}
\item \textbf{State probe $q_1$:} Used at every step of the first run. It asks \emph{``List the colors of the balls you can see and how many there are. Be concise (one short line).''}

\item \textbf{Change probe $q_2$:} Used in the second run, alternating $q_1$ on even steps with one of six questions on odd steps: \emph{``Does any ball change color during the clip? Answer only `yes' or `no'.''}, \emph{``How many DIFFERENT ball colors appear in total? Answer with one number only.''}, \emph{``Is any ball added (appears) during the clip?''}, \emph{``Is any ball removed (disappears)?''}, \emph{``Does any ball move to a different region?''}, and \emph{``Which ball color is the most common? Answer with one color word only.''} The question assigned to each clip is determined by a hash of its identifier, so the same clip always receives the same question, shared by both views.
\end{itemize}

We also evaluate three alternative probe designs in \cref{tab:ablation_table}.

\begin{itemize}
\item \textbf{Change questions only:} Uses the clip's assigned change question at every step, without $q_1$.

\item \textbf{$q_1$ alternated with an entity-ID probe:} Uses $q_1$ on even steps and the following question on odd steps: \emph{``Assign each ball a stable ID, numbered left-to-right by where it first appears. List every visible ball, one per line, as `id$\langle k\rangle$: $\langle$color$\rangle$ @ $\langle$region$\rangle$', then a final line `COUNT: $\langle n\rangle$'. Colors: red, green, blue, yellow, purple, orange, cyan, pink. Regions: top-left, top-right, bottom-left, bottom-right, center.''}

\item \textbf{Colour-set probe:} Used at every step. It asks \emph{``Which ball colors are present? List each color once, in one short line. Do not give any numbers.''}
\end{itemize}

All probes are specified independently of the clip contents and generator.

\subsection{Probe design}

The probe determines which aspects of the scene state receive supervision, since the student learns from the teacher's distribution over its answer to $q$. Our probes name the objects and ask about their colors, counts, and changes over time, which are central to tracking state in \emph{StateGen}. The probe specifies what the model should describe, but never provides the answer. No clip is paired with an answer during training, so the probes provide task instructions without ground-truth supervision.

The ablations in \cref{tab:ablation_table} show the importance of explicitly asking for the current state, including counts. When the colour-set probe forbids numbers, Count accuracy falls to $26.4$, below the base model's $27.7$, and overall accuracy drops to $33.32$. Using only the short change questions performs even worse, reducing overall accuracy to $29.35$. These restricted answers provide less information about the full scene state for the student to learn from. Asking for additional entity IDs and locations also does not improve performance, reaching $35.42$ when alternated with $q_1$.

The two probes focus on different aspects of state tracking. The state probe $q_1$ asks the model to describe the current state, while $q_2$ also asks about changes across the clip. We train a separate run with each probe and combine their learned updates through Probe Merging. On LLaVA-OV-2-8B, the individual runs achieve $36.88$ and $36.11$ overall, compared with $37.44$ after merging. The same pattern holds on Qwen3-VL-8B, where the individual runs score $36.27$ and $36.23$, and the merged model reaches $36.55$. The two runs perform better on different axes (\cref{fig:soupradar}), while the selected merged configuration improves overall accuracy over either individual run on both backbones.

\begin{figure}[t]
  \setlength{\abovecaptionskip}{4pt}
  \centering
\definecolor{armstate}{HTML}{6B4E9B}%
\definecolor{armmixed}{HTML}{B8863B}%
\definecolor{soupgreen}{HTML}{2E7D32}%
\definecolor{soupbase}{HTML}{1F6FB2}%
\definecolor{ctrlred}{HTML}{A63A2E}%
\definecolor{radbg}{HTML}{F7F2F7}%
\begin{tikzpicture}[x=1pt,y=1pt,line join=round,line cap=round]
\footnotesize
\fill[radbg] (60.00,96.00) circle (30.00pt);
\draw[white,line width=0.6pt] (60.00,96.00) circle (7.50pt);
\draw[white,line width=0.6pt] (60.00,96.00) circle (22.50pt);
\draw[white,line width=0.6pt] (60.00,96.00) circle (30.00pt);
\draw[white,line width=0.6pt] (60.00,96.00) -- (60.00,126.00);
\draw[white,line width=0.6pt] (60.00,96.00) -- (83.45,114.70);
\draw[white,line width=0.6pt] (60.00,96.00) -- (89.25,89.32);
\draw[white,line width=0.6pt] (60.00,96.00) -- (73.02,68.97);
\draw[white,line width=0.6pt] (60.00,96.00) -- (46.98,68.97);
\draw[white,line width=0.6pt] (60.00,96.00) -- (30.75,89.32);
\draw[white,line width=0.6pt] (60.00,96.00) -- (36.55,114.70);
\draw[soupbase,line width=0.6pt,dash pattern=on 1.6pt off 1.4pt] (60.00,96.00) circle (15.00pt);
\draw[armstate,line width=0.75pt] (60.00,122.98) -- (71.52,105.19) -- (84.54,90.40) -- (72.17,70.72) -- (48.81,72.77) -- (38.92,91.19) -- (49.05,104.73) -- cycle;
\draw[armmixed,line width=0.75pt] (60.00,125.37) -- (68.74,102.97) -- (74.59,92.67) -- (73.10,68.80) -- (51.03,77.37) -- (53.89,94.61) -- (48.99,104.78) -- cycle;
\fill[soupgreen,opacity=0.11] (60.00,127.79) -- (72.77,106.18) -- (79.90,91.46) -- (73.92,67.09) -- (46.84,68.67) -- (48.06,93.27) -- (46.77,106.55) -- cycle;
\draw[soupgreen,line width=1.35pt] (60.00,127.79) -- (72.77,106.18) -- (79.90,91.46) -- (73.92,67.09) -- (46.84,68.67) -- (48.06,93.27) -- (46.77,106.55) -- cycle;
\filldraw[fill=soupgreen,draw=white,line width=0.4pt] (60.00,127.79) circle (1.5pt);
\filldraw[fill=soupgreen,draw=white,line width=0.4pt] (72.77,106.18) circle (1.5pt);
\filldraw[fill=soupgreen,draw=white,line width=0.4pt] (73.92,67.09) circle (1.5pt);
\filldraw[fill=soupgreen,draw=white,line width=0.4pt] (46.84,68.67) circle (1.5pt);
\filldraw[fill=soupgreen,draw=white,line width=0.4pt] (46.77,106.55) circle (1.5pt);
\node[font=\scriptsize,inner sep=0.6pt] at (60.00,135.50) {Count};
\node[font=\scriptsize,inner sep=0.6pt] at (90.88,120.63) {Loc.};
\node[font=\scriptsize,inner sep=0.6pt] at (103.51,85.21) {Attr.};
\node[font=\scriptsize,inner sep=0.6pt] at (77.14,57.41) {Atom.};
\node[font=\scriptsize,inner sep=0.6pt] at (42.86,57.41) {Seq.};
\node[font=\scriptsize,inner sep=0.6pt] at (16.49,85.21) {Set};
\node[font=\scriptsize,inner sep=0.6pt] at (29.12,120.63) {Dict};
\node[font=\scriptsize,inner sep=0pt] at (60.00,148.00) {(a) LLaVA-OV-2-8B};
\draw[black!30,line width=0.4pt] (30.00,32.00) -- (90.00,32.00);
\draw[soupbase,line width=0.6pt,dash pattern=on 1.6pt off 1.4pt] (60.00,28.80) -- (60.00,35.20);
\node[anchor=north,inner sep=1pt,font=\scriptsize,text=black!55] at (30.00,30.00) {$-6$};
\node[anchor=north,inner sep=1pt,font=\scriptsize,text=black!55] at (90.00,30.00) {$+6$};
\fill[armstate] (70.25,32.00) circle (1.7pt);
\fill[armmixed] (64.90,30.50) rectangle (67.90,33.50);
\filldraw[fill=soupgreen,draw=white,line width=0.5pt] (73.05,34.70) -- (75.75,32.00) -- (73.05,29.30) -- (70.35,32.00) -- cycle;
\node[anchor=south,inner sep=1pt,font=\scriptsize] at (60.00,36.00) {\textbf{Overall} 36.88\,/\,36.11\,$\rightarrow$\,\textbf{37.44}};
\fill[radbg] (176.00,96.00) circle (30.00pt);
\draw[white,line width=0.6pt] (176.00,96.00) circle (7.50pt);
\draw[white,line width=0.6pt] (176.00,96.00) circle (22.50pt);
\draw[white,line width=0.6pt] (176.00,96.00) circle (30.00pt);
\draw[white,line width=0.6pt] (176.00,96.00) -- (176.00,126.00);
\draw[white,line width=0.6pt] (176.00,96.00) -- (199.45,114.70);
\draw[white,line width=0.6pt] (176.00,96.00) -- (205.25,89.32);
\draw[white,line width=0.6pt] (176.00,96.00) -- (189.02,68.97);
\draw[white,line width=0.6pt] (176.00,96.00) -- (162.98,68.97);
\draw[white,line width=0.6pt] (176.00,96.00) -- (146.75,89.32);
\draw[white,line width=0.6pt] (176.00,96.00) -- (152.55,114.70);
\draw[soupbase,line width=0.6pt,dash pattern=on 1.6pt off 1.4pt] (176.00,96.00) circle (15.00pt);
\draw[armstate,line width=0.75pt] (176.00,120.18) -- (191.78,108.58) -- (200.94,90.31) -- (186.33,74.56) -- (166.36,75.98) -- (153.57,90.88) -- (156.05,111.91) -- cycle;
\draw[armmixed,line width=0.75pt] (176.00,119.30) -- (189.23,106.55) -- (205.49,89.27) -- (185.31,76.66) -- (162.35,67.65) -- (150.81,90.25) -- (158.81,109.71) -- cycle;
\fill[soupgreen,opacity=0.11] (176.00,121.35) -- (189.71,106.93) -- (204.58,89.48) -- (185.78,75.69) -- (165.40,73.98) -- (147.62,89.52) -- (155.69,112.20) -- cycle;
\draw[soupgreen,line width=1.35pt] (176.00,121.35) -- (189.71,106.93) -- (204.58,89.48) -- (185.78,75.69) -- (165.40,73.98) -- (147.62,89.52) -- (155.69,112.20) -- cycle;
\filldraw[fill=soupgreen,draw=white,line width=0.4pt] (176.00,121.35) circle (1.5pt);
\filldraw[fill=soupgreen,draw=white,line width=0.4pt] (147.62,89.52) circle (1.5pt);
\filldraw[fill=soupgreen,draw=white,line width=0.4pt] (155.69,112.20) circle (1.5pt);
\node[font=\scriptsize,inner sep=0.6pt] at (176.00,135.50) {Count};
\node[font=\scriptsize,inner sep=0.6pt] at (206.88,120.63) {Loc.};
\node[font=\scriptsize,inner sep=0.6pt] at (219.51,85.21) {Attr.};
\node[font=\scriptsize,inner sep=0.6pt] at (193.14,57.41) {Atom.};
\node[font=\scriptsize,inner sep=0.6pt] at (158.86,57.41) {Seq.};
\node[font=\scriptsize,inner sep=0.6pt] at (132.49,85.21) {Set};
\node[font=\scriptsize,inner sep=0.6pt] at (145.12,120.63) {Dict};
\node[font=\scriptsize,inner sep=0pt] at (176.00,148.00) {(b) Qwen3-VL-8B};
\draw[black!30,line width=0.4pt] (146.00,32.00) -- (206.00,32.00);
\draw[soupbase,line width=0.6pt,dash pattern=on 1.6pt off 1.4pt] (176.00,28.80) -- (176.00,35.20);
\node[anchor=north,inner sep=1pt,font=\scriptsize,text=black!55] at (146.00,30.00) {$-6$};
\node[anchor=north,inner sep=1pt,font=\scriptsize,text=black!55] at (206.00,30.00) {$+6$};
\fill[armstate] (186.80,32.00) circle (1.7pt);
\fill[armmixed] (185.10,30.50) rectangle (188.10,33.50);
\filldraw[fill=soupgreen,draw=white,line width=0.5pt] (188.20,34.70) -- (190.90,32.00) -- (188.20,29.30) -- (185.50,32.00) -- cycle;
\node[anchor=south,inner sep=1pt,font=\scriptsize] at (176.00,36.00) {\textbf{Overall} 36.27\,/\,36.23\,$\rightarrow$\,\textbf{36.55}};
\node[anchor=east,inner sep=0pt,font=\scriptsize,text=soupbase] at (43.50,96.00) {base};
\node[anchor=west,inner sep=0pt,font=\scriptsize,text=black!45] at (83.00,96.00) {$+3$};
\draw[armstate,line width=0.90pt] (30.00,10.00) -- (39.00,10.00);
\node[anchor=west,inner sep=0pt,font=\scriptsize] at (41.00,10.00) {$q_1$ run};
\draw[armmixed,line width=0.90pt] (70.00,10.00) -- (79.00,10.00);
\node[anchor=west,inner sep=0pt,font=\scriptsize] at (81.00,10.00) {$q_2$ run};
\draw[soupgreen,line width=1.40pt] (110.00,10.00) -- (119.00,10.00);
\node[anchor=west,inner sep=0pt,font=\scriptsize] at (121.00,10.00) {\textbf{merged} (ours)};
\draw[soupbase,line width=0.70pt,dash pattern=on 1.6pt off 1.4pt] (180.00,10.00) -- (189.00,10.00);
\node[anchor=west,inner sep=0pt,font=\scriptsize] at (191.00,10.00) {base};
\node[anchor=center,inner sep=0pt,font=\scriptsize] at (118.00,1.00) {radius $=$ score $-$ base ($\pm6$ points full scale)};
\node[anchor=center,inner sep=0pt,font=\scriptsize] at (118.00,-7.00) {\textcolor{soupgreen}{$\bullet$}\,$=$\,merged above \emph{both} runs};
\end{tikzpicture}%
   \caption{\textbf{The two probe runs perform better on different axes, and merging improves overall accuracy.} Per-axis change from the base model for the $q_1$ run, $q_2$ run, and merged model, with the base shown as the dashed circle. Both panels show S$^3$T, on \emph{(a)} LLaVA-OV-2-8B and \emph{(b)} Qwen3-VL-8B.}
  \label{fig:soupradar}
\end{figure}

\subsection{Merging probes versus merging seeds}

\begin{table}[h]
\centering\scriptsize
\setlength{\tabcolsep}{2pt}
\caption{\textbf{Merging seeds versus merging probes.} VSTAT accuracy (\%) for different combinations of trained runs on LLaVA-OV-2-8B. The last column compares each setting with its best single run ($36.48$ for the frozen-encoder ablation and $36.88$ for S$^3$T and the five-seed controls).}
\label{tab:soupcompl}
\begin{tabular}{llcc}
\toprule
Runs merged & Merge & VSTAT & vs.\ best single run\\
\midrule
$5$ seeds of $q_1$ & plain average & $34.65$ & $-2.23$\\
$5$ seeds of $q_1$ & Probe Merging, $\alpha{=}1.0$ & $35.66$ & $-1.22$\\
$5$ seeds of $q_1$ & Probe Merging, $\alpha{=}1.5$ & $36.08$ & $-0.80$\\
$5$ seeds of $q_1$ & Probe Merging, $\alpha{=}2.0$ & $36.03$ & $-0.85$\\
\midrule
S$^3$T ($q_1$ and $q_2$) & Probe Merging, $\alpha{=}1.3$ & $\mathbf{37.44}$ & $\mathbf{+0.56}$\\
\bottomrule
\end{tabular}
\end{table}

To determine whether the improvement comes from combining different probes or simply averaging weights across runs, we compare the two approaches in \cref{tab:soupcompl}. Merging five seeds trained with $q_1$ does not improve over the best single $q_1$ run ($36.88$). A plain weight average gives $34.65$, below the base model, while scaled merging reaches at most $36.08$ among the tested strengths. In contrast, merging $q_1$ and $q_2$ improves over the better individual run for both S$^3$T and the frozen-vision-encoder ablation. These comparisons show that combining runs trained with different probes is more useful than averaging runs trained with the same probe, supporting our choice to merge the two probe-specific updates.

\begin{table*}[t]
\centering
\caption{\textbf{Additional real-video results.}
Changes in accuracy are relative to the LLaVA-OV-2-8B base model unless absolute scores are shown. Brackets indicate $95\%$ bootstrap confidence intervals over evaluation items, and [n.s.] denotes a statistically non-significant change.}
\label{tab:transfer-appendix}
{
\begin{tabular}{l l l}
\toprule
Benchmark & Measure & Result\\
\midrule
Kinetics real-video training ($3$ seeds)
& $\Delta$ overall
& $+0.66 \pm 0.28$~[n.s.]\\

VET-Bench, shell game, $64$f
& 3-way acc
& base $0.31$, S$^3$T $0.33$ ($+0.02$~[n.s.])\\

VET-Bench, shell game, $128$f
& 3-way acc
& base $0.28$, S$^3$T $0.30$ ($+0.02$~[n.s.])\\
\bottomrule
\end{tabular}}
\end{table*}

\begin{table*}[t]
\centering\footnotesize
\caption{\textbf{Uncertainty in real-video cumulative-state results.}
Accuracy improvements over LLaVA-OV-2-8B in percentage points. Confidence intervals use $10{,}000$ paired bootstrap resamples. Video-clustered intervals account for questions sharing a video.}
\label{tab:realvideo-ci}
{
\begin{tabular}{lcccc}
\toprule
Benchmark & Questions & $\Delta$ & 95\% paired CI & 95\% video CI\\
\midrule
VSTAT-YouTube, cumulative-state
& $185$ & $+7.95$ & $[+4.16,+12.00]$ & $[+3.92,+12.10]$\\
MVBench, Action Count
& $200$ & $+4.50$ & $[+1.00,+8.50]$ & --\\
\bottomrule
\end{tabular}}
\end{table*}

\subsection{Merging procedure and strength}
\label{app:heldout}

We merge the trained weight updates rather than model outputs. A LoRA adapter adds $\Delta W=s(BA)$ to each adapted module, where $A$ and $B$ are low-rank factors and $s$ is a fixed scale. For each module, we average the updates from the two runs, scale them by a global strength $\alpha$, and add them to the frozen base weights. This applies to both the language model and vision encoder.

We select $\alpha$ using $100$ held-out \emph{StateGen} clips, independently of VSTAT. For each backbone, we test different strengths on clips from seeds separate from training, using the state probe $q_1$ and the student's $12$-frame view. We compare predicted counts with the generator's ground truth and choose the strength with the lowest mean absolute count error. These clips are used only for model selection, never training. \Cref{tab:heldout} reports the results for both backbones, including strengths around the selected value.

\begin{table}[h]
\centering\footnotesize
\caption{\textbf{Mean absolute count error on held-out \emph{StateGen} clips.} Results for LLaVA-OV-2-8B and Qwen3-VL-8B at different merging strengths $\alpha$, selected independently of VSTAT. Lower is better.}
\label{tab:heldout}
{
\begin{tabular}{lcc}
\toprule
$\alpha$ & LLaVA-OV-2-8B & Qwen3-VL-8B\\
\midrule
1.00 & 3.02 & 3.05\\
1.10 & 3.16 & 2.86\\
1.20 & 3.20 & 2.78\\
1.25 & 3.18 & 2.74\\
1.30 & \textbf{2.76} & \textbf{2.41}\\
1.35 & 3.14 & 2.82\\
1.40 & 3.12 & 2.94\\
1.50 & 3.26 & 3.12\\
1.60 & 3.36 & 3.28\\
1.70 & 3.53 & 3.36\\
\bottomrule
\end{tabular}}
\end{table}

Both backbones achieve their lowest held-out error at $\alpha=1.3$. This supports selecting the merging strength without relying on VSTAT. The improvement does not depend on this tuning. With $\alpha=1.0$ and all other settings unchanged, S$^3$T still reaches $36.62$ on LLaVA-OV-2-8B ($+1.79$) and $35.34$ on Qwen3-VL-8B ($+1.23$).

\section{Additional real-video results}
\label{app:transfer-appendix}

\cref{tab:transfer-appendix} reports additional real-video experiments beyond \cref{tab:transfer_table}. Training on real Kinetics videos yields a small, non-significant overall gain of $+0.66$, but improves cumulative-state accuracy by $+4.09$. On the shell-game task of VET-Bench~\cite{vetbench}, which requires tracking a hidden object through swaps rather than maintaining a running count, both the base model and S$^3$T perform near the three-way chance level at $64$ and $128$ frames, with no significant improvement. These results suggest that S$^3$T can also learn from real videos, with gains concentrated on cumulative-state tracking, while tracking individual objects through occlusion and swaps remains challenging.

\paragraph{Real-video evaluation and uncertainty.}
The VSTAT-YouTube cumulative-state subgroup contains $185$ questions from $98$ videos within the $850$-question YouTube split of the main $1{,}500$-question VSTAT evaluation. It is defined by task semantics, using the latte art, LEGO, matryoshka, graffiti, cooking, coffee, order packing, horse racing, and carousel task families, rather than by model performance. This is a subgroup analysis of VSTAT, not an independent benchmark.

We compute paired bootstrap confidence intervals by resampling matched evaluation questions. For VSTAT-YouTube, we also resample at the video level to account for multiple questions from the same video. The $+7.95$-point improvement remains significant under both procedures (\cref{tab:realvideo-ci}). On the independent MVBench Action Count benchmark, S$^3$T improves by $+4.50$ points over $200$ questions, with a paired confidence interval that also excludes zero.

\end{document}